\documentclass[accepted]{uai2026} 

\usepackage[british]{babel}

\usepackage[sort, authoryear, round]{natbib}
\usepackage{listings}
\usepackage{xcolor}
\usepackage{booktabs} 
\usepackage{amsmath}  
\usepackage{graphicx}
\usepackage{algorithm}
\usepackage{algpseudocode}

\definecolor{promptbg}{RGB}{248,248,248}
\definecolor{promptborder}{RGB}{200,200,200}
\definecolor{promptkeyword}{RGB}{0,102,204}
\definecolor{promptcomment}{RGB}{120,120,120}
\definecolor{promptstring}{RGB}{163,21,21}

\lstdefinestyle{llmprompt}{
  backgroundcolor=\color{promptbg},
  basicstyle=\ttfamily\small,
  frame=single,
  rulecolor=\color{promptborder},
  frameround=tttt,
  breaklines=true,
  showstringspaces=false,
  tabsize=2,
  keywordstyle=\color{promptkeyword}\bfseries,
  commentstyle=\color{promptcomment}\itshape,
  stringstyle=\color{promptstring},
  captionpos=b,
  numbers=none,
  aboveskip=1em,
  belowskip=1em,
  literate={→}{{$\rightarrow$}}1 {✓}{{$\checkmark$}}1 {✗}{{$\times$}}1
}

\newcommand{\contraryempty}{\overline{\phantom{a}}}
\newcommand{\contrary}[1]{\overline{#1}}
\newcommand{\gpath}{\mathit{p}}

\newtheorem{theorem}{Theorem}
\numberwithin{theorem}{section} 

\newtheorem{example}[theorem]{Example}

\usepackage{tikz}
\usetikzlibrary{decorations.pathreplacing,calligraphy,shapes.geometric,fit}
\tikzset{
    cnode/.style={
        circle,
        draw,
        minimum size=.5cm,
        align=center,
        text centered,        
        anchor=center,          
        inner sep=1pt,
        font=\small
    }
}

\usepackage{amsfonts}
\newcommand{\independent}{\ensuremath{\!\perp \!\!\! \perp\!}}

\newcommand{\given}{\ensuremath{\mid}}

\def\DGP{\ensuremath{\mathcal{P}}}

\def\DAG{G}
\def\SKEL{\ensuremath{\mathcal{C}}}

\def\edgSet{\ensuremath{E}}

\def\nodeSet{\ensuremath{\mathbf{V}}}

\def\Xvar{\ensuremath{\mathbf{X}}}
\def\Yvar{\ensuremath{\mathbf{Y}}}
\def\Zvar{\ensuremath{\mathbf{Z}}}

\usepackage{multicol}
\usepackage{multirow}
\usepackage{subcaption}

\usepackage{colortbl}
\usepackage{siunitx}
\definecolor{bestcell}{RGB}{217,239,207} 
\newcommand{\best}[1]{\cellcolor{bestcell}{\bfseries #1}}
\newcommand{\up}{\(\uparrow\)}
\newcommand{\down}{\(\downarrow\)}

\title{Leveraging Large Language Models for Causal Discovery: a Constraint-based, Argumentation-driven Approach}

\author[1]{Zihao Li}
\author[1]{Fabrizio Russo}

\affil[1]{%
    Department of Computing\\

    Imperial College London\\
    UK\\

    \href{mailto:zihao.li24@imperial.ac.uk}{zihao.li24@imperial.ac.uk}\\
    \href{mailto:fabrizio@imperial.ac.uk}{fabrizio@imperial.ac.uk}
}

\begin{document}
\maketitle

\begin{abstract}
Causal discovery seeks to uncover causal relations from data, typically represented as causal graphs, and is essential for predicting the effects of interventions. While expert knowledge is required to construct principled causal graphs, many statistical methods have been proposed to leverage observational data with varying formal guarantees. Causal Assumption-based Argumentation (ABA) is a framework that uses symbolic reasoning to ensure correspondence between input constraints and output graphs, while offering a principled way to combine data and expertise. We explore the use of large language models (LLMs) as imperfect experts, eliciting semantic structural constraints from variable names and descriptions and integrating them with statistical evidence through Causal ABA. We propose ABAPC-LLM as a principled and conservative approach to hybrid data- and LLM-driven causal discovery. Additionally, we introduce an evaluation protocol to mitigate memorisation bias when assessing LLMs for causal discovery and show competitive performance on novel semantically grounded random benchmarks, as well as on standard small- and medium-sized benchmarks.
\end{abstract}
\section{Introduction}\label{sec:intro}
Understanding the causal structure that governs observational data is central to explanation, prediction, and decision-making and underpins reasoning about interventions~\citep{spirtes2000causation,pearl2009causality,peters2017elements}. Although interventional/experimental data can help resolve causal directions, it is costly and often unavailable; accordingly, a rich literature has focused on extracting causal graphs from observational data only.
Constraint- and score-based methods infer (equivalence classes of) directed acyclic graphs (DAGs) from statistical regularities, but reliability hinges on assumptions and sample size~\citep{spirtes2000causation}, motivating the use of domain knowledge alongside data.
\vspace{-0.05cm}

Formulating such background knowledge is labour-intensive: domain experts must identify relevant variables, specify admissible orientations, and articulate forbidden relations. Existing pipelines either encode these judgements as hard structural constraints, fixing or forbidding edges before discovery~\citep{meek1995causal}, or as Bayesian priors that weight the search over graph structures~\citep{cooper1992bayesian,heckerman_learning_1995}.
Both hard constraints and priors can encode expertise, but offer limited mechanisms to diagnose and resolve conflicts with different sources of evidence such as observational data.
\vspace{-0.05cm}

Causal Assumption-Based Argumentation (Causal ABA)~\citep{russo24causalaba} reframes discovery as an ABA framework~\citep{DBLP:journals/ai/BondarenkoDKT97,DBLP:journals/flap/CyrasFST17} where candidate causal edges are defeasible assumptions, and rules for acyclicity and $d$-separation~\citep{pearl2009causality} induce attacks between incompatible constraints.
Solving it yields DAGs supported by the strongest jointly defensible subset of data- and expert-derived constraints, with provenance via the accepted/defeated assumptions. Crucially, this framework allows to integrate difference sources of knowledge without treating it as ground truth, and to expose conflicts within the input constraints transparently.
\vspace{-0.05cm}

Large language models (LLMs) provide an attractive, yet imperfect, source of knowledge. They encode broad semantic and scientific information and can perform structured reasoning when prompted appropriately \citep{bommasani2021opportunities,kojima2022large}. To leverage this information, one needs to rely on semantically meaningful variable metadata (names and descriptions rather than anonymised identifiers such as $X_1,\dots,X_n$), providing semantic signal unavailable to data-only discovery pipelines. 
Yet, LLMs can hallucinate plausible-sounding but unsupported causal 
links~\citep{ji2023survey}, which goes against the rigour that causal discovery demands. Additionally, their training on web-scale corpora, where many benchmark causal graphs are publicly available, makes it difficult to discern whether suggested dependences stem from genuine reasoning or the verbatim retrieval of the memorised causal graph in its entirety. 

These tensions raise two research questions: how can we exploit the knowledge embedded in LLMs without sacrificing rigour and transparency, and how can we assess LLMs' contributions to causal discovery while mitigating memorisation bias?
We address these questions as follows:
\begin{itemize}
    \item We design a robust \textbf{LLM integration pipeline} that elicits high-precision directional constraints, consensus-filters them, and integrates them defeasibly with data-derived conditional-independence evidence within Causal ABA.
    \item We develop a \textbf{synthetic evaluation protocol} that grounds random DAGs in the \texttt{CauseNet} knowledge graph~\citep{Heindorf2020CauseNet} to better assess LLM generalisation beyond memorised benchmarks.
    \item We show competitive performance on standard benchmarks and on our synthetic protocol.
\end{itemize}
\vspace{-0.1cm}
A key design choice is conservatism: semantic information from LLMs is used to orient (not introduce) edges, and statistical evidence can overrule LLM-derived constraints through defeasible reasoning.
\vspace{-0.2cm}
\section{Related Work}
\vspace{-0.1cm}
We build on three strands of literature: causal discovery from observational data, the integration of expert knowledge into causal discovery, and the use of LLMs for knowledge elicitation and causal discovery.

\vspace{-0.1cm}
\textbf{Causal Discovery. } Causal discovery algorithms aim to reconstruct causal graphs from observational data, often under the assumptions of acyclicity and faithfulness: that the causal structure is a DAG and that all and only the conditional independencies implied by the DAG are present in the data \citep{spirtes2000causation}. A rich literature (see e.g.\ \citep{glymour2019review,ZANGA2022survey,vowels2021doyalike_survey}) has proposed various algorithms with different assumptions, guarantees, and trade-offs between statistical and computational efficiency.
Throughout, we assume causal sufficiency (no unobserved confounders)~\citep{spirtes2000causation}.

\vspace{-0.1cm}
Constraint-based algorithms such as Peter-Clark (PC) \citep{spirtes2000causation} and its improved version, Majority-PC (MPC)~\citep{colombo2014order}, infer Markov equivalence classes (MEC, the upper bound of what can be discovered from observational data alone~\citep{verma1990equivalence}) through conditional independence (CI) tests, whereas score-based procedures such as greedy equivalence search \citep{chickering2002optimal} or its faster counterpart Fast Greedy Search (FGS)~\citep{ramsey2017million} trade combinatorial search for statistical fit criteria. 
We use MPC to derive an efficient number of CI constraints, but our approach is agnostic to the underlying testing procedure.
We also compare against representative score-based (FGS), continuous-optimisation (NOTEARS-MLP~\citep{zheng2020notears_mlp}\footnote{Without claiming its fitness for causal discovery; e.g. see cautionary discussion in \citep{reisach2021beware}.}), and order-based baselines (GRaSP~\citep{lam2022grasp}, BOSS~\citep{andrews2023boss}).

\vspace{-0.1cm}
Hybrid approaches tighten the shortcomings of single types of methods by combining constraint- and score-based approaches~\citep{tsamardinos2006MMHC} while logic-based approaches embed logical inference in the process e.g. via SAT-based encodings~\citep{hyttinen2013discovering}. Causal ABA fits into the logical constraint-based literature using ABA~\citep{DBLP:journals/ai/BondarenkoDKT97} to guarantee that every inferred orientation is backed by an explicit chain of assumptions and rules~\citep{russo24causalaba}. Our work complements these efforts by integrating structured assumptions from LLMs.

\vspace{-0.1cm}
\textbf{Integrating Expert Knowledge.} Prior knowledge has long been recognised as essential to trim the search space and enforce domain restrictions in causal discovery~\citep{cooper1992bayesian,heckerman_learning_1995}. Conventional approaches require experts to enumerate admissible edges, ancestral relations, or forbidden paths before running an algorithm~\citep{meek1995causal}. \citep{hyttinen2013discovering} incorporate domain constraints during search or as post-processing filters, including answer-set-programming (ASP, \citep{GebserKKS17}) and weighted-constraint formulations that can encode background knowledge and resolve inconsistent constraints during optimisation.
By embedding (LLM- or human-sourced) statements into Causal ABA, we can integrate expert knowledge defeasibly and expose conflicts with data-derived constraints. This is complementary to logic programming approaches: LLM- or human-derived statements are represented as assumptions whose acceptance, rejection, and attacks can be inspected, giving provenance for why a semantic constraint survived or was defeated rather than only enforcing it as ordinary hard background knowledge.

\vspace{-0.1cm}
Recent work has moved toward interactive integration of expert knowledge during discovery. This includes approaches that request human input on uncertain edges or constraints as the search progresses~\citep{cao2024causal,kitson_causal_2023} and expert-in-the-loop procedures that iteratively refine a learned graph with domain feedback~\citep{ankan2025expertloop}. These methods depend on ongoing human interaction or repeated model refitting. In contrast, our pipeline elicits structured constraints once (from LLMs or experts) and integrates them defeasibly within Causal ABA, allowing conflicts with data-derived CI evidence to be resolved transparently without iterative expert sessions.

\vspace{-0.1cm}
\textbf{LLMs for Causal Discovery.} 
LLMs exhibit strong few-shot generalisation~\citep{brown2020languagemodelsfewshotlearners} and can be prompted to elicit causal constraints from variable metadata~\citep{kiciman2023causal,zhao2023largelanguagemodelscommonsense}. For causal discovery, this ability enables LLMs to act as experts and \emph{translate metadata into candidate structural constraints}.

A growing literature leverages LLMs in three complementary roles \citep{wan_large_2025}: (i)~querying them to orient edges or assemble full causal graphs \citep{kiciman2023causal,jiralerspong2024efficient,vashishtha2025causal}; (ii)~using them as assistants that critique, refine, or document algorithmic outputs via refinement loops or natural-language analytics \citep{khatibi2024alcmautonomousllmaugmentedcausal,abdulaal2024causal,gkountouras2024languageagentsmeetcausality,liu2024discoveryhiddenworldlarge}; and (iii)~coupling textual constraints with statistical search through hard~\citep{chen2023mitigatingpriorerrorscausal} or soft constraints~\citep{ban2023query}, or even by replacing CI tests with conversational judgments within PC~\citep{cohrs2023large}. This literature also highlights that LLM and human judgements are noisy imperfect-expert signals rather than ground truth~\citep{vashishtha2025causal}. In our approach, MPC provides a set of CI statements from data, while the consensus-filtered LLM constraints are encoded as additional required/forbidden arrow facts and enforced within Causal ABA only when they do not contradict strong statistical counter-evidence (\S\ref{sec:integrate}). As in~\citep{kiciman2023causal,jiralerspong2024efficient,vashishtha2025causal}, we solicit direct causal relations, yet we consolidate the queries into a single prompt for improved efficiency and context. Additionally, we adapt the consensus refinement strategy of~\citep{cohrs2023large} and benchmark against the BFS-style LLM-only method of~\citep{jiralerspong2024efficient}, which iteratively expands a graph via pairwise queries.
Compared to directly enforcing text-derived constraints, our contribution is using argumentation to expose and resolve conflicts, providing a principled integration with data and traceable justifications for accepted and rejected orientations.

\vspace{-0.2cm}
LLM-agentic systems have also been proposed for interactive graph discovery, where agents plan queries and update graphs over multiple steps~\citep{havrilla2025igda}. Our method is non-agentic and avoids iterative prompting loops; instead, we prioritise high-precision, consensus-filtered constraints that can be justified and, if necessary, overturned by the argumentation-based consistency checks.

\vspace{-0.2cm}
Despite this promise, LLM responses remain fallible. Benchmarks reveal gaps between memorised answers and genuine causal reasoning \citep{jin2023cladder,ze2023causal,wan_large_2025}, and repeated studies caution against using LLMs as sole decision-makers for causal discovery \citep{wu2025llmdiscovercausalityrestricted}.
These limitations motivate treating elicited knowledge defeasibly rather than as ground truth and designing evaluation protocols that mitigate memorisation bias. Our empirical study therefore treats LLMs as an input source to the Causal ABA integration mechanism.

\vspace{-0.2cm}
\section{Background}\label{sec:background}
\vspace{-0.2cm}
Let $\DAG=(\nodeSet,\edgSet)$ be a graph with nodes $\nodeSet$ and edges $\edgSet\subseteq\nodeSet\times\nodeSet$.  
Edges can be directed or undirected (for $u,v\in\nodeSet$, if both $(u,v)\in\edgSet$ and $(v,u)\in\edgSet$, then $u$ and $v$ are connected by an undirected edge); the \emph{skeleton} replaces every directed edge with an undirected one. A node's degree is the number of edges linked to it. A \emph{path} $\gpath=x_1\dots x_n$ is a sequence of adjacent nodes; it is \emph{simple} when all nodes are distinct. It is \emph{directed} if $(x_i,x_{i+1})\in\edgSet$ for all $i$, and cyclic if it contains a repeated node, i.e. $x_i=x_j$ for some $i<j$; a directed cycle is a cyclic directed path.
A \emph{directed acyclic graph (DAG)} has only directed edges and no cycles and will serve as our causal graph formalism. A DAG is ascribed a causal semantics by interpreting nodes as random variables and edges as direct causal effects \citep{pearl2009causality}. A Bayesian Network (BN) is a pair $(\DAG,\DGP)$ where $\DGP$ is a joint distribution over $\nodeSet$ that satisfies the Markov condition relative to $\DAG$: every variable is independent of its non-descendants given its parents.
For pairwise disjoint sets $\Xvar,\Yvar,\Zvar\subseteq \nodeSet$  we let  $(\Xvar\!\independent\!\Yvar\!\given\!\Zvar)$ indicate that $\Xvar$ and $\Yvar$ are \emph{independent} given the conditioning set  $\Zvar$; $(\Xvar\!\independent\!\Yvar\!\given\!\emptyset)$ is simply written as $(\Xvar\!\independent\!\Yvar)$ and singleton sets $\{x\}$ with $x\in\nodeSet$ are denoted by $x$ (e.g., $(\{x\}\!\independent\!\{y\}\!\given\!\emptyset)$ is written as $(x\!\independent\!y)$).
The link between DAGs and CI is captured by the $d$-separation criterion~\citep{pearl2009causality}.

\vspace{-0.04cm}
Causal ABA \citep{russo24causalaba} is a framework that combines computational argumentation \citep{Dung95,Toni14tutorialABA} with causal reasoning. 
An ABA framework (ABAF)~\citep{DBLP:journals/ai/BondarenkoDKT97} is a tuple $\langle \mathcal{L}, \mathcal{R}, \mathcal{A}, \contraryempty \rangle$ where $\mathcal{L}$ is a formal \emph{language}, $\mathcal{R}$ is a set of \emph{inference rules} of the form $head \leftarrow body$ where $head \in \mathcal{L}$ and $body$ is a finite (possibly empty) set of elements from $\mathcal{L}$, $\mathcal{A} \subseteq \mathcal{L}$ is a non-empty set of \emph{assumptions}, and $\contraryempty$ is a total mapping from $\mathcal{A}$ into $\mathcal{L}$ (\emph{contrary}). 
In Causal ABA, the language $\mathcal{L}$ includes statements about graphical (causal) relationships (e.g.\ $(x,y)$), paths and independencies; the rules $\mathcal{R}$ encode the principles of acyclicity and $d$-separation, and the assumptions $\mathcal{A}$ represent candidate causal relations and/or independencies that can be accepted or rejected based on the evidence and their relations. An assumption $a \in \mathcal{A}$ \emph{attacks} an assumption $b \in \mathcal{A}$ if and only if $a$ can be used to derive the contrary of $b$, i.e., $\contrary{b}$, using the rules in $\mathcal{R}$. 
\begin{figure*}[ht]
    \centering
    \includegraphics[width=\linewidth]{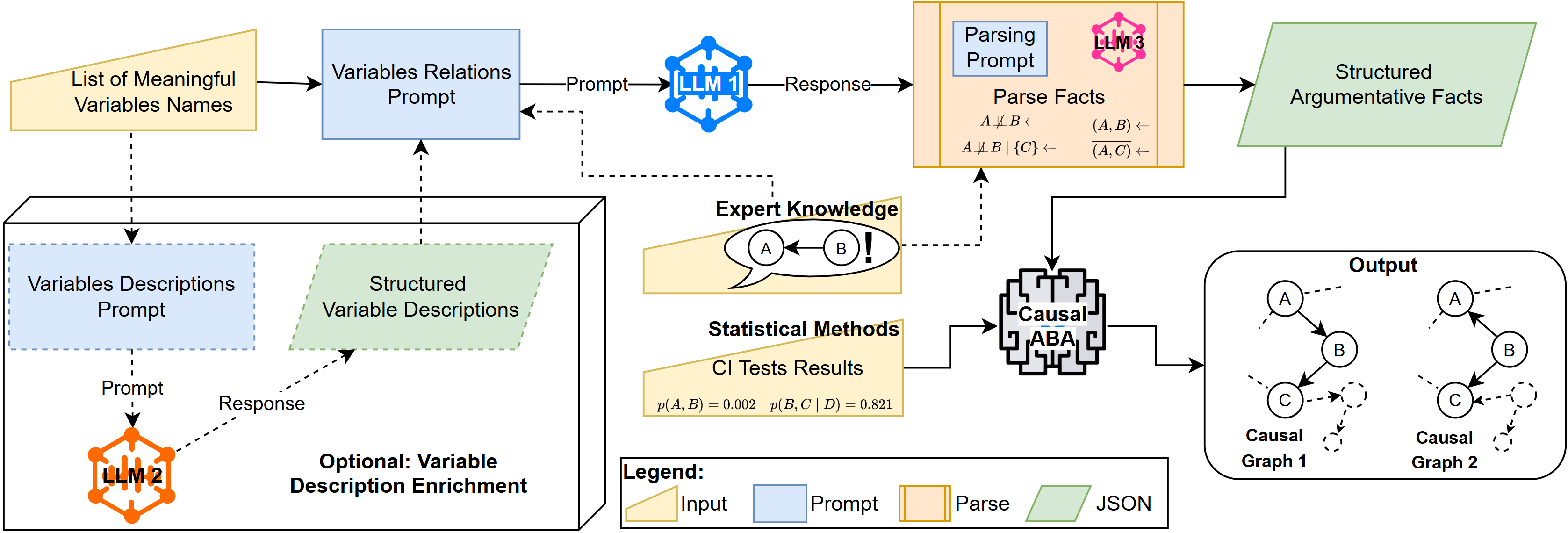}
    \caption{LLM Integration Pipeline: Given a set of variables with their names and (possibly) descriptions, we prompt an LLM to generate pairwise causal statements. These statements are then parsed into structured assumptions for Causal ABA, which combines them with data-derived independence or arrow constraints and background knowledge to infer a set of causal graphs. Expert knowledge can be injected in the LLM prompts or as defeasible facts. Variable descriptions and LLM parsing are optional components of the pipeline (dashed arrows) but enhance the quality of the generated assumptions. Detailed prompts and parsing rules are provided in App.~\ref{sec:appendix_prompts}.}
    \label{fig:llm-pipeline}
\end{figure*}

\vspace{-0.04cm}
An ABAF is evaluated using argumentation \emph{semantics} \citep{baroni_introduction_2011}, which determine which sets of assumptions (called \emph{extensions}) can be collectively accepted based on their ability to cohexist (conflict-freeness) and defend against external attacks (see e.g. \citep{Toni14tutorialABA} for formal definitions and examples). A set of assumptions $S \subseteq \mathcal{A}$ will produce no solution if it contains conflicts, i.e., if there exists $a, b \in S$ such that $a$ attacks $b$. Causal ABA employs \emph{stable semantics} to guarantee a one-to-one correspondence between the accepted assumptions and the $d$-separation relations induced by the output DAG \citep{russo24causalaba}. 
This makes inconsistencies explicit: incompatible constraints attack each other, and only jointly defensible sets are retained.
A bodyless rule ($head \leftarrow$) is called a \emph{fact} and represents an unconditionally true statement in the ABAF i.e., it is by default included in every extension. In the implementation, the abstract rules, assumptions, and facts are \emph{grounded} into a ASP program over the variables and candidate paths before stable extensions are computed; this grounding step is the main source of the scalability trade-offs discussed below.

In \citep{russo24causalaba} ABAPC is proposed as an instantiation of Causal ABA using MPC as the statistical engine to derive CI constraints from data. The CI constraints from MPC are ranked based on a transform of their associated $p$-values, and the algorithm iteratively attempts to construct stable extensions by removing the least \emph{credible} facts (according to the ranking) until one is found. In this way, ABAPC treats CI outcomes as defeasible and selects a high-confidence subset that is jointly consistent with the $d$-separations of one or multiple DAGs.

\begin{example} Consider the simple four-variable DAG in Fig.~\ref{fig:toy_example_graphs}(a): Education ($E$) and Race ($R$) influence Occupation ($O$), which in turn influences Income ($I$); Education also directly affects Income ($E \to I$).
\begin{figure}[ht]
    \centering
    \begin{minipage}{0.48\linewidth}
        \centering
        \begin{tikzpicture}
            \node[cnode] (E) at (0,0.5) {$E$};
            \node[cnode] (R) at (0,-0.5) {$R$};
            \node[cnode] (O) at (1.1,0) {$O$};
            \node[cnode] (I) at (2.5,0) {$I$};

            \path[->,thick,>=stealth]
            (E) edge (O)
            (R) edge (O)
            (O) edge (I)
            (E) edge[out=0,in=155] (I)
            ;
        \end{tikzpicture}
        \par\vspace{-0.18cm}
        \makebox[\linewidth][c]{\hspace{0.22cm}{\small\textbf{(a)}~True graph.}}
    \end{minipage}
    \hfill
    \begin{minipage}{0.48\linewidth}
        \centering
        \begin{tikzpicture}
            \node[cnode] (E) at (0,0.5) {$E$};
            \node[cnode] (R) at (0,-0.5) {$R$};
            \node[cnode] (O) at (1.1,0) {$O$};
            \node[cnode] (I) at (2.5,0) {$I$};

            \path[->,thick,>=stealth]
            (E) edge[-] (O)
            (R) edge (O)
            (E) edge[-,out=0,in=155] (I)
            (I) edge (O)
            ;
        \end{tikzpicture}
        \par\vspace{-0.18cm}
        \makebox[\linewidth][c]{\hspace{0.22cm}{\small\textbf{(b)}~MPC output.}}
    \end{minipage}
    \vspace{-0.28cm}
    \caption{Toy example graphs (true DAG vs.\ MPC output).}
    \label{fig:toy_example_graphs}
\end{figure}
From synthetic data sampled from this ground-truth DAG we obtain 23 CI tests using MPC. These test results, including $E \independent R$ (correct) and $E \independent R \given O$ (wrong wrt\ the $d$-separations in the ground truth DAG), become facts in Causal ABA. Conditioning on the collider $O$ should render $E$ and $R$ dependent, yet PC-style procedures treat both statements as equally trustworthy. When MPC consumes these tests, it returns the partially oriented DAG in Fig.~\ref{fig:toy_example_graphs}(b).
Causal ABA ingests the same CI statements, but reasons about their joint consistency. Each test result is ranked by credibility, and the framework searches for a stable extension. Including both independence statements yields no stable extension: the $d$-separation rules make the statements inconsistent, because $E \independent R \given O$ can hold together with $E \independent R$ only if $R$ and $E$ are disconnected from the other variables---contradicting many of the other CI tests. After progressive exclusions, the stable extension contains only the correct independence statement, yielding the ground-truth DAG.
\end{example}
This toy example highlights the key advantage of ABAPC over PC-style procedures: it can discard a small number of low-credibility CI outcomes that are globally inconsistent, preventing cascading orientation errors.
Having described our statistical and symbolic engines, we turn to our proposed method to integrate LLM-derived knowledge with data.
\section{LLM-Augmented Causal ABA}\label{sec:integrate}
Our methodology integrates semantic knowledge from LLMs and statistical test results with the symbolic rigour of Causal ABA. The core idea is to use an LLM as a proxy domain expert to generate high-precision structural constraints, which are then encoded as facts within the argumentative framework to prune the vast search space of possible causal graphs. This hybrid approach, depicted in Fig.~\ref{fig:llm-pipeline}, aims to enhance both the efficiency and the accuracy of the discovery process. The pipeline consists of a method to elicit structured constraints from LLMs (left), and the formal integration of these constraints into Causal ABA (right). We refer to the resulting LLM-augmented framework as \textbf{ABAPC-LLM}. Human expert knowledge is included as optional, and not currently part of the experiments, but can be readily integrated. Importantly, the LLM never sees the observational samples: it is queried only on variable names/descriptions.
\begin{algorithm}[t]
\caption{ABAPC-LLM}
\label{alg:abapc-llm}
\begin{algorithmic}[1]
\Statex \textbf{Input:} $\mathcal{I},\alpha,|\nodeSet|=d$; metadata $M$, LLM $L$, repetitions $r=5$, MPC skeleton $\SKEL$
\State $\mathbf{T}\leftarrow$ Lines 1--8 of \citep[Alg.~1]{russo24causalaba} on $(\mathcal{I},\alpha,d)$ \Comment{CI facts sorted by strength}
\State $\mathbf{R}\leftarrow [\,]$, $\mathbf{F}\leftarrow [\,]$
\For{$i=1,\ldots,r$}
    \State $(R_i,F_i)\leftarrow L(M)$ \Comment{required and forbidden arrows}
\EndFor
\State $\mathbf{R}\leftarrow \{(x,y):(x,y)\in\cap_i R_i \ \wedge\ x-y\in\SKEL\}$
\State $\mathbf{F}\leftarrow \{\,\overline{(x,y)}:(x,y)\in\cap_i F_i\,\}$
\State $\mathbf{T}\leftarrow \mathbf{T}+[\mathbf{R},\mathbf{F}]$ \Comment{add unanimous compatible LLM facts}
\State $G\leftarrow$ Lines 9--21 of Causal ABA, with \texttt{causalaba} grounded over $\SKEL$ and $\mathbf{T}$
\State \Return $G$
\end{algorithmic}
\end{algorithm}
Algorithm~\ref{alg:abapc-llm} is written as a modification of the original Causal ABA algorithm: the CI-fact construction, relaxation loop, and final graph scoring are unchanged except that the grounded ABAF receives the additional unanimous LLM facts and uses the reduced MPC skeleton.
\vspace{-0.2cm}
\subsection{Constraint Elicitation Pipeline}
\vspace{-0.15cm}
The quality of LLM-derived knowledge is highly dependent on the elicitation process~\citep{anthropicPromptImprover2024}. We propose a robust pipeline to transform unstructured metadata (in the form of variable names and descriptions) into conservative and precise causal constraints.
\vspace{-0.15cm}

\textbf{Eliciting Structural Constraints.} The process begins with a set of variables, each with a meaningful name and an optional longer description. We developed a detailed prompt that instructs a primary LLM (LLM 1 in Fig.~\ref{fig:llm-pipeline}) to act as a domain expert, asking for the generation of two lists containing the following types of information based on established knowledge or logical consistency:
\vspace{-0.15cm}
\begin{itemize}
    \item \textbf{Required Directions}: Causal relationships ($(x,y)$ with $x,y\in\nodeSet$) that the causal graph should contain.
\vspace{-0.15cm}
    \item \textbf{Forbidden Directions}: Causal relationships $\contrary{(x,y)}$ that the causal graph should not contain.
\end{itemize}
\vspace{-0.15cm}
We use these as directional constraints (edge orientations) rather than as claims that an adjacency exists in the data.
The LLM is instructed to structure its response and provide brief justifications for each constraint. The complete prompt is provided in App.~\ref{sec:eli_prompt}. We then apply a schema-guided extraction stage (LLM 3 in Fig.~\ref{fig:llm-pipeline}, using \citep{instructor}) to parse and validate the free-form output into structured required/forbidden arrow lists; details are in App.~\ref{sec:schema-guided_extraction}.

\textbf{Consensus for High Precision.}
To mitigate the stochasticity of LLM outputs and adhere to the high-precision requirement of causal discovery, we adapt the majority voting idea from \citep{cohrs2023large} and introduce a consensus-based filtering step. We query the primary LLM five times independently with the same prompt. The final constraint sets consist only of the intersection of the required and forbidden arrow sets across all five parsed responses. This conservative approach ensures that only the most consistently suggested constraints are passed to Causal ABA, significantly reducing the risk of incorporating spurious constraints (see App.~\ref{sec:appendix-consensus} for details).
Of course, consensus does not guarantee correctness; it is a pragmatic variance-reduction step aimed at improving precision.
\vspace{-0.4cm}
\subsection{Integration into Causal ABA}
\vspace{-0.2cm}
ABAPC-LLM combines CI evidence from data with semantic arrow constraints elicited from LLMs: a set of required and forbidden arrow facts derived from variable metadata and added to the ABAF alongside the MPC-derived CI facts (line 6-8 of Alg.~\ref{alg:abapc-llm}). CI statements are handled within Causal ABA as weighted constraints and are progressively relaxed until the program admits a stable extension (and thus a DAG).
In the current implementation, a consensus LLM direction that enters the framework is treated as a hard directional fact, subject to compatibility with the reduced skeleton: an LLM orientation can be blocked when the corresponding adjacency has been removed from the search space because of a CI test. To treat LLM constraints as weak constraints and make them defeasible not only against direct CI facts, they would need to be calibrated against CI-test confidence. Deriving such calibrated semantic weights, allowing low-trust semantic constraints to be dropped or deferred when conflicts arise, is left to future work.
\vspace{-0.15cm}

\textbf{Data-Driven Skeleton Reduction.} To improve the scalability of Causal ABA we perform an initial, one-off \emph{skeleton reduction step}: we run the MPC testing procedure to obtain an initial search skeleton, and then ground edge and path rules only over the adjacencies that survive that skeleton search. This follows causal theory and the strategy employed by both PC and Causal ABA, with the difference that, during the subsequent iterative relaxation of lower-confidence CI constraints,
these edges are not reintroduced. MPC still runs normally and records the relevant CI tests; the approximation is only in the Causal ABA grounding. Unlike full ABAPC, which reintroduces edge/path rules after relaxing an independence fact, our optimised version does not, so the full correspondence guarantee (Prop. 5.18 in~\citep{russo24causalaba}) of the unoptimised framework does not apply unchanged and final graphs can differ when recovery would require omitted paths. This optimisation is applied to both ABAPC and ABAPC-LLM.
Overall, ABAPC-LLM inherits the CI-testing cost of MPC but adds overhead from solving the ABAF and from a one-off set of LLM calls per dataset; empirical runtimes are discussed in \S\ref{sec:results_analysis}.
\begin{figure*}[h!t]
    \centering
    \includegraphics[width=\linewidth]{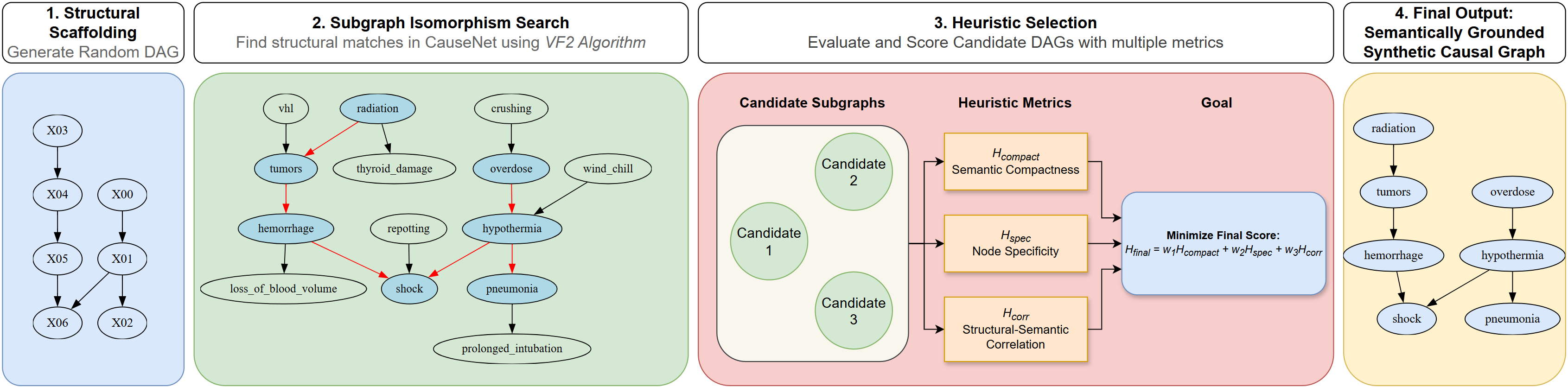}
    \caption{\texttt{CauseNet} synthetic benchmark protocol: 1. generate a random DAG, 2. ground it via induced sub-graph isomorphism in \texttt{CauseNet}, 3. rank candidate mappings with a three-term heuristic, 4. yielding a semantically grounded DAG with named variables for LLM elicitation.\\[-0.7cm]}
    \label{fig:synthetic_pipeline}
\end{figure*}

\vspace{-0.15cm}
\textbf{LLM Constraints Enforcement.} We then add the consensus-filtered constraints to Causal ABA as facts:
\vspace{-0.1cm}
\begin{itemize}
    \item A \textbf{required arrow} (encoded as $(x, y) \leftarrow$) is enforced only if the undirected edge $x\!-\!y$ remains in the reduced skeleton, so semantic knowledge can orient data-supported adjacencies but cannot introduce new ones (yielding conservative, sparse graphs).
    \vspace{-0.2cm}
   \item A \textbf{forbidden arrow} (encoded as $\overline{(x, y)} \leftarrow$) restricts the program from orienting an edge in a certain direction.
\end{itemize}
\vspace{-0.2cm}
This two-stage process leverages statistical evidence to define the initial search space and employs semantic knowledge to further guide the discovery with information that is often orthogonal to what can be inferred from data.
LLMs orient surviving adjacencies only; they cannot add edges. This helps avoiding spurious semantic adjacencies, but limits the gains when CI tests wrongly delete an edge (see~\S\ref{sec:results_analysis}).
\vspace{-0.8cm}
\section{CauseNet Synthetic DAGs}
\label{sec:eval_strategy}
\vspace{-0.3cm}
A fundamental challenge in evaluating LLMs for causal discovery is the risk of data leakage~ \citep{dong2024generalizationmemorizationdatacontamination,balloccu2024leakcheatrepeatdata}. Standard benchmarks, such as \texttt{Asia} \citep{10.1111/j.2517-6161.1988.tb01721.x} or \texttt{Sachs} \citep{doi:10.1126/science.1105809} in the \texttt{bnlearn} repository \citep{bnlearn}, are widely published and likely part of LLMs' pre-training corpora. Consequently, high performance on these benchmarks may reflect non-generalisable memorisation.
Our proposed protocol, illustrated in Fig.~\ref{fig:synthetic_pipeline}, embeds randomly generated DAG structures within \texttt{CauseNet}, a large-scale knowledge graph of cause-and-effect relationships extracted from web text \citep{Heindorf2020CauseNet}. It produces semantically grounded ground-truth DAGs with named variables; we then instantiate each DAG as a BN and sample observational data for structure-learning experiments. The pipeline consists of three main stages, as follows.
\vspace{-0.2cm}

\textbf{1. Structural Scaffolding.}
We begin by generating a structural ``scaffold''. To ensure structural diversity in our benchmarks, we first define a DAG topology using one of three methods: generating Erdős-Rényi (ER)~\citep{Erdos:1959:pmd}, Scale-Free (SF)~\citep{doi:10.1126/science.286.5439.509} graphs, or directly constructing a DAG from a randomly populated lower triangular (LT) adjacency matrix. 
More details, including generation schema and examples, are in App.~\ref{sec:scaffold_graphs}.

\textbf{2. Semantic Grounding via Isomorphism.}
To assign meaningful concepts to the nodes of the random DAG, we use \texttt{CauseNet}. We search for an \textbf{induced sub-graph isomorphism} from the random DAG to the \texttt{CauseNet} graph. Formally, given a random DAG $H=(V_H, E_H)$ and \texttt{CauseNet} $G=(V_G, E_G)$, we find an injective function $f: V_H \to V_G$ such that for any pair of nodes $u, v \in V_H$, an edge $(u,v) \in E_H$ exists if and only if an edge $(f(u), f(v)) \in E_G$ exists. This strategy ensures that the selected sub-graph from \texttt{CauseNet} has the same structure as our random DAG. For this search, we employ the VF2 algorithm of~\citep{cordella2004subgraph}.

\textbf{3. Heuristic-Guided Candidate Selection.}
A single random DAG can have thousands or even millions of isomorphic matches within \texttt{CauseNet}.
To find the most plausible causal scenario, we score and rank all candidate sub-graphs using a heuristic that combines three distinct cost terms into a single objective function to be minimised:
\vspace{-0.2cm}
\begin{itemize}
    \item \textbf{Semantic Compactness:} Measures the thematic coherence of the concepts in a candidate graph. Using pre-computed vector embeddings for all concepts, we calculate the average cosine distance of each concept's vector from their geometric centroid. A low cost indicates that the concepts form a tight semantic cluster (e.g., \texttt{virus}, \texttt{fever}, \texttt{fatigue}).
 \vspace{-0.08cm}
   \item \textbf{Node Specificity:} Penalises candidate graphs that include overly general or ambiguous ``hub'' nodes (e.g., \texttt{illness}, \texttt{problem}). The cost is the average logarithm of each node's degree in the full \texttt{CauseNet} graph, favouring sub-graphs composed of more specific, well-defined concepts.
 \vspace{-0.1cm}
   \item \textbf{Structural-Semantic Correlation:} Induces faithfulness of the graph's topology to the semantic relationships between its concepts. We compute the rank correlation~\citep{Spearman1904rank} between all-pairs shortest-path distances in the graph and the corresponding pairwise cosine distances between semantic embeddings.
    Lower cost (inverse of correlation) means that concepts directly connected in the graph are also semantically very close, ensuring holistic consistency.
\end{itemize}
The selected candidate is the graph that minimises a weighted sum of these three heuristic cost terms. This pipeline produces semantically grounded causal graphs that serve as a robust foundation for evaluating causal discovery involving LLMs. We make our synthetic DAG generator publicly available as \href{https://clarg-group.github.io/CauseNet-graph-generator/}{open-source software} and provide more details in App.~\ref{sec:ground_heuristics}.

\textbf{Why Semantic Grounding Is Necessary.}
Since LLM elicitation relies on variable semantics, anonymised variables (e.g., $X_1,\dots,X_n$ instead of \texttt{virus,...fatigue}) would remove the signal and reduce the model to an uninformed oracle. We therefore ground randomly generated structures in real concepts via \texttt{CauseNet} to obtain semantically meaningful \emph{yet structurally diverse} graphs; while individual CauseNet facts may appear in pre-training corpora, composing them into random, previously unseen subgraphs makes verbatim retrieval of an entire target DAG implausible and helps distinguish memorisation from generalisation. Our protocol does not claim that individual \texttt{CauseNet} relations are unseen by the LLMs. Instead, it mitigates recall of complete public benchmark graphs by randomising topology and composing many local semantic relations into novel DAGs. Accordingly, stronger results on standard \texttt{bnlearn} networks should be interpreted cautiously, since they are consistent with possible benchmark memorisation, whereas \texttt{CauseNet} is intended to reduce full-graph leakage.

\textbf{Sampling Observational Data.}
For structure-learning experiments, we instantiate each grounded DAG as a BN and sample observational data from it. We randomly initialise CPTs since we evaluate structure recovery rather than effect sizes; if needed, CPT parameters can be chosen to better reflect the semantic grounding (details in App.~\ref{sec:data_gen_schema}).
\begin{table*}[t]
    \caption{Structural metrics on \texttt{CauseNet} synthetic datasets, grouped by node count $|\nodeSet|$. Among repeated-run methods, bold marks the best mean and statistically tied methods under BH-corrected Welch tests within each metric/column (Tabs.~\ref{tab:causenet_main_indicator_tests}-\ref{tab:structural_tied_bolding_tests}, App.~\ref{sec:appendix_statistical_tests}). Significance levels for the $p$-values against the next non-bold method in the ranking are: \textnormal{***} $p<0.001$, \textnormal{**} $p<0.01$, \textnormal{*} $p<0.05$. LLM-BFS is shown from one saved run and is excluded from Welch tests and bolding.}
    \label{tab:causenet_dag_full_metrics}
    \centering
    \scriptsize
    \setlength{\tabcolsep}{2pt}
    \resizebox{0.83\textwidth}{!}{%
    \begin{tabular}{l|lcccc}
    \toprule
    & \textbf{Method} & \textbf{SHD $\downarrow$} & \textbf{Precision $\uparrow$} & \textbf{Recall $\uparrow$} & \textbf{F1 $\uparrow$} \\
    \midrule
    \multirow{10}{*}{\rotatebox{90}{\textbf{$|\nodeSet|=5$}}} & Random & \num{6.886 +- 1.478} & \num{0.266 +- 0.145} & \num{0.384 +- 0.230} & \num{0.307 +- 0.171} \\
     & FGS & \num{5.684 +- 1.079} & \num{0.273 +- 0.198} & \num{0.181 +- 0.133} & \num{0.209 +- 0.151} \\
     & NOTEARS-MLP & \num{5.332 +- 0.008} & \num{0.001 +- 0.008} & \num{0.000 +- 0.002} & \num{0.001 +- 0.003} \\
     & GRaSP & \num{3.590 +- 1.404} & \num{0.765 +- 0.257} & \num{0.346 +- 0.238} & \num{0.597 +- 0.125} \\
     & BOSS & \num{3.529 +- 1.305} & \num{0.778 +- 0.224} & \num{0.356 +- 0.226} & \num{0.606 +- 0.117} \\
     & MPC & \num{3.294 +- 0.643} & \num{0.859 +- 0.101} & \num{0.371 +- 0.117} & \num{0.602 +- 0.060} \\
     & MPC-LLM & \num{1.500 +- 0.318} & {\bfseries \num{0.958 +- 0.031}}\,\textnormal{***} & \num{0.710 +- 0.050} & \num{0.795 +- 0.037} \\
     & ABAPC & \num{2.025 +- 0.549} & \num{0.712 +- 0.087} & \num{0.620 +- 0.092} & \num{0.671 +- 0.084} \\
     & ABAPC-LLM & {\bfseries \num{1.230 +- 0.369}}\,\textnormal{***} & \num{0.882 +- 0.046} & {\bfseries \num{0.763 +- 0.062}}\,\textnormal{***} & {\bfseries \num{0.812 +- 0.049}}\,\textnormal{*} \\
     & LLM-BFS & \num{2.667} & \num{0.842} & \num{0.559} & \num{0.628} \\
    \midrule
    \multirow{10}{*}{\rotatebox{90}{\textbf{$|\nodeSet|=10$}}} & Random & \num{28.247 +- 6.281} & \num{0.141 +- 0.069} & \num{0.302 +- 0.177} & \num{0.183 +- 0.091} \\
     & FGS & \num{16.362 +- 1.747} & \num{0.173 +- 0.114} & \num{0.099 +- 0.066} & \num{0.123 +- 0.078} \\
     & NOTEARS-MLP & \num{12.642 +- 0.122} & \num{0.026 +- 0.119} & \num{0.002 +- 0.011} & \num{0.004 +- 0.021} \\
     & GRaSP & \num{8.148 +- 2.752} & \num{0.819 +- 0.257} & \num{0.389 +- 0.187} & \num{0.537 +- 0.176} \\
     & BOSS & \num{7.391 +- 2.627} & {\bfseries \num{0.880 +- 0.197}}\,\textnormal{***} & \num{0.439 +- 0.184} & \num{0.579 +- 0.173} \\
     & MPC & \num{7.055 +- 1.338} & \num{0.854 +- 0.107} & \num{0.453 +- 0.101} & \num{0.579 +- 0.098} \\
     & MPC-LLM & \num{6.206 +- 1.023} & \num{0.856 +- 0.088} & \num{0.511 +- 0.078} & \num{0.627 +- 0.077} \\
     & ABAPC & \num{6.166 +- 1.442} & \num{0.724 +- 0.127} & \num{0.535 +- 0.105} & \num{0.612 +- 0.110} \\
     & ABAPC-LLM & {\bfseries \num{5.626 +- 1.176}}\,\textnormal{***} & \num{0.788 +- 0.089} & {\bfseries \num{0.565 +- 0.086}}\,\textnormal{***} & {\bfseries \num{0.653 +- 0.084}}\,\textnormal{***} \\
     & LLM-BFS & \num{13.278} & \num{0.505} & \num{0.436} & \num{0.439} \\
    \midrule
    \multirow{10}{*}{\rotatebox{90}{\textbf{$|\nodeSet|=15$}}} & Random & \num{63.766 +- 19.364} & \num{0.082 +- 0.036} & \num{0.299 +- 0.171} & \num{0.122 +- 0.053} \\
     & FGS & \num{24.303 +- 2.112} & \num{0.097 +- 0.077} & \num{0.060 +- 0.047} & \num{0.074 +- 0.057} \\
     & NOTEARS-MLP & \num{16.982 +- 0.180} & \num{0.024 +- 0.121} & \num{0.001 +- 0.008} & \num{0.002 +- 0.014} \\
     & GRaSP & \num{10.422 +- 3.014} & \num{0.857 +- 0.217} & \num{0.413 +- 0.153} & \num{0.544 +- 0.158} \\
     & BOSS & \num{9.994 +- 3.010} & \num{0.880 +- 0.191} & \num{0.431 +- 0.156} & \num{0.563 +- 0.162} \\
     & MPC & \num{8.809 +- 1.190} & {\bfseries \num{0.936 +- 0.071}}\,\textnormal{***} & \num{0.479 +- 0.069} & \num{0.621 +- 0.072} \\
     & MPC-LLM & \num{7.874 +- 1.038} & {\bfseries \num{0.929 +- 0.061}}\,\textnormal{***} & \num{0.532 +- 0.061} & {\bfseries \num{0.667 +- 0.058}}\,\textnormal{*} \\
     & ABAPC & \num{7.902 +- 1.537} & \num{0.762 +- 0.100} & \num{0.546 +- 0.087} & \num{0.632 +- 0.090} \\
     & ABAPC-LLM & {\bfseries \num{7.550 +- 1.281}}\,\textnormal{**} & \num{0.789 +- 0.077} & {\bfseries \num{0.565 +- 0.074}}\,\textnormal{***} & \num{0.654 +- 0.072} \\
     & LLM-BFS & \num{18.833} & \num{0.438} & \num{0.351} & \num{0.356} \\
    \bottomrule
    \end{tabular}
    }
\end{table*}
\vspace{-0.3cm}
\section{Experimental Evaluation}
\label{sec:exp_eval}
\vspace{-0.25cm}
We now present an empirical evaluation of our LLM-augmented Causal ABA framework. Our experiments aim at assessing the accuracy and robustness of the inferred causal graphs, particularly in comparison to established causal discovery algorithms and under varying conditions of LLM input quality. Our goal is to use semantic constraints as additional information to orient edges within the MEC; we therefore evaluate the final output against the ground-truth DAG. Additionally, we investigate the interaction between semantic and statistical constraints, and how this interplay affects the discovery process.
\vspace{-0.4cm}
\subsection{Experimental Setup}
\label{sec:exp_setup}
\vspace{-0.3cm}
\textbf{Data.} We evaluate our approach on both standard causal discovery benchmarks and our novel synthetic datasets derived from \texttt{CauseNet}. For each dataset, we generate 5000 observational samples and repeat the experiment 50 times with different random seeds. LLM-derived constraints are generated once per dataset from the metadata and reused across all 50 data repetitions. We use standard pseudo-real BNs from the \texttt{bnlearn} repository \citep{bnlearn}, specifically: \emph{cancer, earthquake, survey, asia, sachs} and \emph{child}. Our synthetic datasets consist of 54 random DAGs with number of nodes $|\nodeSet| \in \{5,10,15\}$, generated using the protocol described in \S\ref{sec:eval_strategy}. Detailed statistics of all datasets are provided in App.~\ref{sec:appendix_graph_data_generation}, and the generated synthetic BNs and full experimental code are available in our \href{https://github.com/briziorusso/ArgCausalDisco}{GitHub repository}.

\textbf{Metadata Preparation and Prompting.} We generated variable descriptions for causal graphs using an LLM (LLM 2 in Fig.~\ref{fig:llm-pipeline}), providing it with the ground-truth graph as a basis for defining each variable's role. The prompt was carefully designed to ensure the resulting descriptions were semantically consistent with the underlying causal structure without explicitly stating any of the causal links. The full prompt is provided in App.~\ref{sec:enrich_prompt}. Because these synthetic descriptions are LLM-generated from a ground-truth graph-level description, they should be viewed as a controlled upper-information condition rather than realistic documentation. Names-only ablations in the appendix (Tabs.~\ref{tab:desc_consensus_ablation} and \ref{tab:desc_consensus_nonzero_ablation}) provide a lower-information stress test; noisy, human-written, and obfuscated metadata are important future robustness tests. We then used these variable names and descriptions as input to our constraint elicitation pipeline (\S\ref{sec:integrate}) to generate the LLM-derived constraints (LLM 1 in Fig.~\ref{fig:llm-pipeline}).
We use two relatively small but mainstream LLMs, Google's \texttt{gemini-2.5-flash} and OpenAI \texttt{gpt-5-mini}, with the same prompts and five-run unanimity rule; \texttt{gemini-2.5-flash-lite} is used for structured output extraction (LLM 3 in Fig.~\ref{fig:llm-pipeline}). We leveraged Anthropic's prompt improver \citep{anthropicPromptImprover2024} to iteratively refine the prompts before running both primary models.
We evaluate \texttt{gemini-2.5-flash} and \texttt{gpt-5-mini} under the same consensus protocol; coverage counts are reported in App.~\ref{sec:appendix-consensus}. We use the same variable names and descriptions for all LLM-based methods.

\textbf{Metrics.} We measure the accuracy of the estimated causal graphs using several metrics: Structural Hamming Distance \citep[SHD]{tsamardinos2006MMHC}, Structural Intervention Distance \citep[SID]{peters2015structural}, precision, recall and F1-score. SHD quantifies the number of edge modifications needed to transform the estimated graph into the true graph, while SID measures the correctness of estimated causal effects under interventions. The F1-score is the harmonic mean of precision and recall with respect to the classification of arrows in the estimated graph against the ground-truth DAG. SID is only used for the \texttt{bnlearn} datasets, as it requires plausible BN parameters that are not available for the \texttt{CauseNet} graphs.
Details in App.~\ref{sec:appendix_metrics_defs}.
\vspace{0.02cm}

\textbf{Baselines.} We compare ABAPC-LLM against several baselines: ABAPC~\citep{russo24causalaba}, MPC~\citep{colombo2014order}, FGS \citep{ramsey2017million}, NOTEARS-MLP \citep{zheng2020notears_mlp}, GRaSP \citep{lam2022grasp}, and BOSS \citep{andrews2023boss}. ABAPC and MPC share our CI-testing setup; comparing ABAPC-LLM to ABAPC isolates the contribution of semantic arrow constraints, while comparing ABAPC to MPC isolates the contribution of the argumentative layer. MPC-LLM injects the same consensus LLM constraints used by ABAPC-LLM as hard background knowledge into MPC through direct background knowledge~\citep{zheng2024causal}, without the Causal ABA layer, separating direct hard-constraint use in MPC from argumentative integration using Causal ABA. The choice of CI testing method is consistent across MPC/-LLM and ABAPC/-LLM: Wilks $G^2$
test~\citep{agresti2018introduction}, with significance level $\alpha=0.01$. Within ABAPC/-LLM, MPC is only used to compute (an efficient number of) CI tests and their associated $p$-values. The remaining baselines span score-based, continuous optimisation, and permutation-based search used with standard hyperparameters. We additionally include an LLM-only baseline~\citep{jiralerspong2024efficient} (LLM-BFS; metadata-only), which isolates LLM performance without CI tests regardless of the method of integration.\footnote{This was run only once per dataset due to API costs.} Finally, we include a random graph generation baseline for a lower bound on performance. Details in App.~\ref{sec:baseline_det}.
\subsection{Results and Analysis}
\label{sec:results_analysis}
\vspace{-0.3cm}
We first report structural recovery, then discuss runtimes and the semantic--statistical constraint interaction. Unless explicitly stated otherwise, the LLM-based main-text results use \texttt{gemini-2.5-flash}; the matched-datasets comparison (Tab.~\ref{tab:gpt_model_ablation_dag}, App.~\ref{sec:appendix_gemini_gpt}) shows that the differences between using the two sets of constraints are often not significant even though \texttt{gpt-5-mini} constraint quality is lower.
\vspace{-0.2cm}

\textbf{Structural Learning Performance.}
Table~\ref{tab:causenet_dag_full_metrics} compares DAG reconstruction on \texttt{CauseNet}. ABAPC-LLM has the lowest SHD and highest recall for every node count, and the best F1 at $|\nodeSet|=\{5,10\}$. The $|\nodeSet|=15$ F1 exception is about the integration method: adding the same LLM constraints raises MPC F1 by 0.046, but ABAPC F1 by only 0.022. The graph-type breakdown in App.~\ref{sec:appendix_additional_metrics} (Tab.~\ref{tab:causenet_dag_v_graph_type}) shows that this shift is concentrated in LT graphs, where MPC-LLM gains 0.09 F1 and reduces SHD by 1.91 over MPC, while ABAPC-LLM gains 0.035 F1 and reduces SHD by 0.60 over ABAPC. This matches the integration policy in \S\ref{sec:integrate}: ABAPC-LLM only uses LLM-required arrows when the reduced skeleton permits them, while MPC-LLM passes them directly as MPC background knowledge. In the $|\nodeSet|=15$ LT cases, 14.4\% of true LLM-required arrow opportunities are blocked by initially wrong CI facts; because blocked arrows are not regrounded after later relaxation, conservative arbitration can leave useful semantic directions unused. Overall, this conservatism preserves the best SHD and recall on average, but gives MPC-LLM an F1 advantage where the direct constraints are especially useful.
A significance level ablation (Tab.~\ref{tab:alpha_ablation_dag}, App.~\ref{sec:appendix_additional_metrics}) is compatible with the same interaction story: at $\alpha=0.05$, the looser CI threshold gives MPC-LLM a larger recall boost, making it the top F1 method; ABAPC-LLM still improves over ABAPC, but its conservative arbitration converts less of the additional semantic signal into F1.
On \texttt{bnlearn}, ABAPC-LLM is also competitive across most structural metrics (Tab.~\ref{tab:bnlearn_dag_all_metrics}, App.~\ref{sec:appendix_bnlearn}).

\textbf{LLM Constraints Quality and Interaction with Data.}
To better understand how our LLM-augmented pipeline improves structural accuracy, we inspect the quality of the LLM constraints entering Causal ABA in relation to the data-driven ones. For each run we compute: (i) an aggregate F1-score of the consensus-filtered required/forbidden arrow constraints against the ground-truth DAG; (ii) the F1-score of the CI statements ultimately retained by ABAPC (data-only, after progressive relaxation), again against ground-truth; and (iii) the resulting change in final graph F1 after adding the semantic constraints ($\Delta$F1). We then bin both constraint-quality F1 into low/mid/high ranges ([0,0.33), [0.33,0.66), [0.66,1]) and report the mean $\Delta$F1 in each cell.

\begin{figure}[t]
    \centering
    \includegraphics[width=1\linewidth]{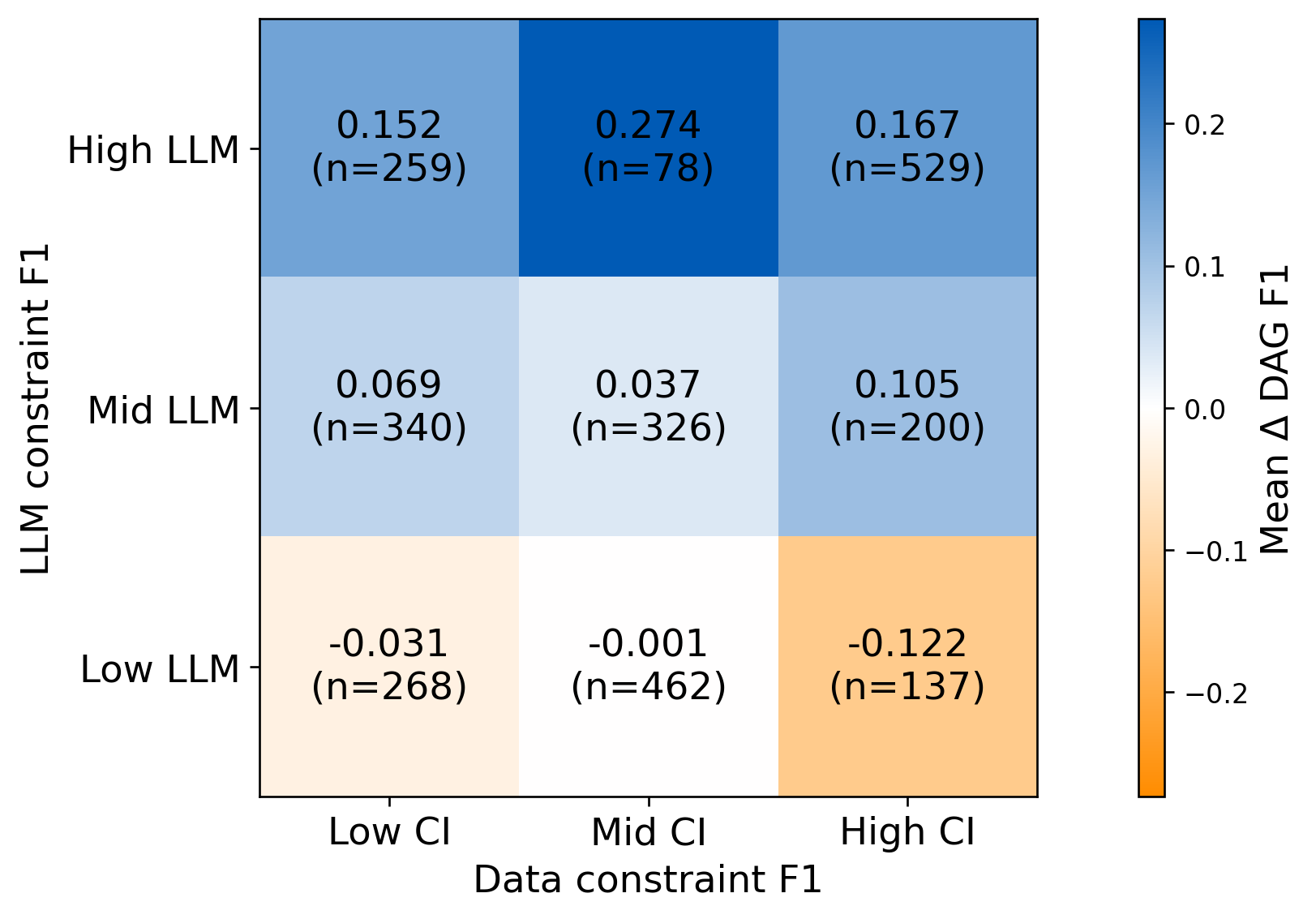}
    \caption{Heatmap of the quality of LLM-derived and data-derived constraints in relation to changes in final graph reconstruction accuracy after integration ($\Delta$F1 against the true DAG; in brackets the number of constraints $n$).\\[-0.8cm]}
    \label{fig:constraint_quality}
\end{figure}

\vspace{-0.1cm}
Fig.~\ref{fig:constraint_quality} shows synergy when the semantic signal is reliable: high-LLM cells are consistently positive. Conversely, low-quality LLM constraints can hurt even with strong CI evidence (Low LLM/High CI: mean $\Delta$F1 $=-0.122$). Although \emph{required} arrows only orient adjacencies surviving skeleton reduction (\S\ref{sec:integrate}), LLM constraints are enforced: conflicts are repaired by relaxing additional CI statements, which can weaken the statistical signal. The conflict is often not a single CI fact directly opposing an LLM arrow: orientations are supported by rules across multiple CI tests. In the current implementation, an LLM direction can be defeated when its adjacency has been removed, but it is not calibrated against a composite CI signal supporting the opposite orientation. This failure mode is most visible with variable descriptions: they increase constraint quality and coverage in many cases (Tabs.~\ref{tab:desc_consensus_ablation} and \ref{tab:desc_consensus_nonzero_ablation}; App.~\ref{sec:appendix_constraint_eval}), but can also introduce misleading context. With names only, negative cells are much weaker (minimum mean $\Delta$F1 $=-0.013$; Fig.~\ref{fig:heapmap_quality_synthetic_no_desc}), supporting the interpretation that richer metadata can both help and mislead.
On the \texttt{bnlearn} benchmarks, the interaction heatmap is almost entirely non-negative and has larger gains in high-LLM cells (e.g., $0.287$ for High LLM/High CI; Fig.~\ref{fig:heapmap_quality_bn}). This is consistent with the higher apparent quality of LLM-derived constraints on public benchmark networks, but should be interpreted cautiously because those networks are likely represented in pre-training data. See App.~\ref{sec:appendix_constraint_eval} for the metadata and consensus ablations, and App.~\ref{sec:appendix_interaction_details} for the per-size, per-type, and per-dataset heatmaps.

\textbf{Runtime.} ABAPC-LLM is slower than MPC due to solving the ABAF, but our skeleton-reduction optimisation makes both ABAPC and ABAPC-LLM tractable with less than 30 variables: excluding external API latency (one-off per dataset), runtimes are in seconds up to 15 nodes on synthetic benchmarks and up to 20 variables on \texttt{bnlearn}. ABAPC-LLM has comparable runtime to ABAPC on small datasets (5 nodes) but reduces solve time by pruning the search space on bigger ones (Tab.~\ref{tab:runtime_scaling_final}, App.~\ref{sec:appendix_runtime}). The bottleneck is grounding and solving the Causal ABAF: CI literals grow with ordered variable pairs and conditioning sets, path rules grow rapidly with the number of simple paths, and stable-extension reasoning is NP-complete in the grounded ABAF size. Skeleton reduction keeps the evaluated cases tractable, but larger dense skeletons remain a failure mode.\vspace{-0.4cm}

\section{Conclusion}
\vspace{-0.3cm}
We introduced a hybrid causal discovery framework that injects LLM-derived semantic knowledge into Causal ABA, combining auditable symbolic conflict resolution with rich and unstructured semantic information. A dedicated elicitation pipeline extracts conservative structural constraints from natural-language metadata, while our \texttt{CauseNet}-grounded synthetic benchmarks provide a setting where memorisation is unlikely and improvements in structural, interventional and precision--recall metrics become apparent. Overall, we advocate using LLMs as auditable sources of causal directions, mediated by formal consistency checks.
\vspace{-0.1cm}

Our findings show that consensus-filtered semantic constraints can boost structural accuracy complementing statistical evidence, motivating several avenues for future work. In particular, we plan to mine knowledge from scientific corpora, extend the dialogue with LLMs to reason about potential unobserved confounders, and endow Causal ABA with adaptive weighting schemes that calibrate LLM constraints against data-driven ones. 

\begin{acknowledgements}
Russo was supported by the ERC under the ERC-POC programme (grant number 101189053).
\end{acknowledgements}

\bibliography{bib}

@inproceedings{russo24causalaba,
  author       = {Fabrizio Russo and
                  Anna Rapberger and
                  Francesca Toni},
  opteditor       = {Pierre Marquis and
                  Magdalena Ortiz and
                  Maurice Pagnucco},
  title        = {Argumentative Causal Discovery},
  optbooktitle    = {Proceedings of the 21st International Conference on Principles of
                  Knowledge Representation and Reasoning, {KR} 2024, Hanoi, Vietnam.
                  November 2-8, 2024},
    booktitle    = {Proc. of {KR} 2024},
  year         = {2024},
  opturl       = {https://doi.org/10.24963/kr.2024/88},
  optdoi          = {10.24963/KR.2024/88},
  note          = {\href{https://doi.org/10.24963/KR.2024/88}{doi: 10.24963/KR.2024/88}},
  bibsource    = {dblp computer science bibliography, https://dblp.org}
}

@book{spirtes2000causation,
  title        = {Causation, Prediction, and Search},
  author       = {Spirtes, Peter and Glymour, Clark and Scheines, Richard},
  publisher    = {MIT Press},
  year         = {2000}
}

@article{zheng2024causal,
  title        = {Causal-learn: Causal Discovery in Python},
  author       = {Zheng, Yujia and Huang, Biwei and Chen, Wei and Ramsey, Joseph and Gong, Mingming and Cai, Ruichu and Shimizu, Shohei and Spirtes, Peter and Zhang, Kun},
  optjournal   = {Journal of Machine Learning Research},
  journal      = {J. Mach. Learn. Res.},
  volume       = {25},
  number       = {60},
  pages        = {1--8},
  year         = {2024},
  note         = {\href{https://jmlr.org/papers/v25/23-0970.html}{JMLR}},
  opturl      = {https://jmlr.org/papers/v25/23-0970.html}
}

@book{pearl2009causality,
  title        = {Causality: Models, Reasoning, and Inference},
  author       = {Pearl, Judea},
  edition      = {2},
  publisher    = {Cambridge University Press},
  year         = {2009}
}

@book{peters2017elements,
author = {Peters, J. and Janzing, D. and Sch\"olkopf, B.},
title = {Elements of Causal Inference: Foundations and Learning Algorithms},
optaddress = {Cambridge, MA, USA},
publisher = {MIT Press},
year = {2017},
opturl = {https://mitpress.mit.edu/9780262037310/elements-of-causal-inference/}
}

@inproceedings{meek1995causal,
  author       = {Christopher Meek},
  opteditor       = {Philippe Besnard and
                  Steve Hanks},
  title        = {Causal inference and causal explanation with background knowledge},
  optbooktitle = {{UAI} '95: Proceedings of the Eleventh Annual Conference on Uncertainty
                  in Artificial Intelligence, Montreal, Quebec, Canada, August 18-20,
                  1995},
  booktitle    = {Proc. of {UAI} 1995},
  pages        = {403--410},
  optpublisher = {Morgan Kaufmann},
  year         = {1995},
  opturl       = {https://doi.org/10.48550/arXiv.1302.4972},
  bibsource    = {dblp computer science bibliography, https://dblp.org},
  note         = {\href{https://dl.acm.org/doi/10.5555/2074158.2074204}{doi: 10.5555/2074158.2074204}}
}

@article{cooper1992bayesian,
  title        = {A Bayesian Method for the Induction of Probabilistic Networks from Data},
  author       = {Cooper, Gregory F. and Herskovits, Edward},
  optjournal   = {Machine Learning},
  journal      = {Mach. Learn.},
  volume       = {9},
  number       = {4},
  pages        = {309--347},
  year         = {1992},
  note = {\href{https://doi.org/10.1007/BF00994110}{doi: 10.1007/BF00994110}},
}

@inproceedings{hyttinen2013discovering,
  author       = {Antti Hyttinen and
                  Frederick Eberhardt and
                  Matti J{\"{a}}rvisalo},
  opteditor       = {Nevin L. Zhang and
                  Jin Tian},
  title        = {Constraint-based Causal Discovery: Conflict Resolution with Answer
                  Set Programming},
  optbooktitle = {Proceedings of the Thirtieth Conference on Uncertainty in Artificial
                  Intelligence, {UAI} 2014, Quebec City, Quebec, Canada, July 23-27,
                  2014},
  booktitle    = {Proc. of {UAI} 2014},
  pages        = {340--349},
  optpublisher = {{AUAI} Press},
  year         = {2014},
  bibsource    = {dblp computer science bibliography, https://dblp.org}
}

@article{bommasani2021opportunities,
  title        = {On the Opportunities and Risks of Foundation Models},
  author       = {Bommasani, Rishi and Hudson, Drew A. and Adeli, Ehsan and Altman, Russ and Arora, Sanjeev and von Arx, Sydney and others},
  optjournal   = {arXiv preprint arXiv:2108.07258},
  journal      = {arXiv preprint \href{https://arxiv.org/abs/2108.07258}{arXiv:2108.07258}},
  year         = {2021}
}

@inproceedings{kojima2022large,
  title        = {Large Language Models are Zero-Shot Reasoners},
  author       = {Kojima, Takeshi and Gu, Shixiang Shane and Reid, Machel and Matsuo, Yutaka and Iwasawa, Yusuke},
  optbooktitle = {Advances in Neural Information Processing Systems},
  booktitle    = {Proc. of {NeurIPS} 2022},
  year         = {2022},
  note         = {\href{https://proceedings.neurips.cc/paper_files/paper/2022/hash/8bb0d291acd4acf06ef112099c16f326-Abstract-Conference.html}{NeurIPS proceedings}}
}

@article{ji2023survey,
  title        = {A Survey on Hallucination in Natural Language Generation},
  author       = {Ji, Ziwei and Lee, Nanyun and Yu, Tianyi and He, Qihui and Zhang, Ruibo and Ren, Xiang and others},
  optjournal   = {ACM Computing Surveys},
  journal      = {ACM Comput. Surv.},
  volume       = {55},
  number       = {12},
  pages        = {1--38},
  year         = {2023},
  note         = {\href{https://doi.org/10.1145/357173}{doi: 10.1145/357173}}
}

@article{DBLP:journals/ai/BondarenkoDKT97,
  author       = {Andrei Bondarenko and
                  Phan Minh Dung and
                  Robert A. Kowalski and
                  Francesca Toni},
  title        = {An Abstract, Argumentation-Theoretic Approach to Default Reasoning},
  journal      = {Artif. Intell.},
  volume       = {93},
  pages        = {63--101},
  year         = {1997},
  opturl       = {https://doi.org/10.1016/S0004-3702(97)00015-5},
  optdoi          = {10.1016/S0004-3702(97)00015-5},
  note          = {\href{https://doi.org/10.1016/S0004-3702(97)00015-5}{doi: 10.1016/S0004-3702(97)00015-5}},
  bibsource    = {dblp computer science bibliography, https://dblp.org}
}

@article{DBLP:journals/flap/CyrasFST17,
  author       = {Kristijonas Cyras and
                  Xiuyi Fan and
                  Claudia Schulz and
                  Francesca Toni},
  title        = {Assumption-based Argumentation: Disputes, Explanations, Preferences},
  journal      = {{FLAP}},
  volume       = {4},
  number       = {8},
  year         = {2017},
  opturl       = {http://www.collegepublications.co.uk/downloads/ifcolog00017.pdf},
  note         = {\href{http://www.collegepublications.co.uk/downloads/ifcolog00017.pdf}{FLAP}},
  bibsource    = {dblp computer science bibliography, https://dblp.org}
}

@article{heckerman_learning_1995,
	title = {Learning {Bayesian} {Networks}: {The} {Combination} of {Knowledge} and {Statistical} {Data}},
	volume = {20},
	optissn = {1573-0565},
	optshorttitle = {Learning {Bayesian} {Networks}},
	opturl = {https://doi.org/10.1023/A:1022623210503},
	optdoi = {10.1023/A:1022623210503},
	note = {\href{https://doi.org/10.1023/A:1022623210503}{doi: 10.1023/A:1022623210503}},
	optabstract = {We describe a Bayesian approach for learning Bayesian networks from a combination of prior knowledge and statistical data. First and foremost, we develop a methodology for assessing informative priors needed for learning. Our approach is derived from a set of assumptions made previously as well as the assumption of likelihood equivalence, which says that data should not help to discriminate network structures that represent the same assertions of conditional independence. We show that likelihood equivalence when combined with previously made assumptions implies that the user's priors for network parameters can be encoded in a single Bayesian network for the next case to be seen—a prior network—and a single measure of confidence for that network. Second, using these priors, we show how to compute the relative posterior probabilities of network structures given data. Third, we describe search methods for identifying network structures with high posterior probabilities. We describe polynomial algorithms for finding the highest-scoring network structures in the special case where every node has at most k = 1 parent. For the general case (k {\textgreater} 1), which is NP-hard, we review heuristic search algorithms including local search, iterative local search, and simulated annealing. Finally, we describe a methodology for evaluating Bayesian-network learning algorithms, and apply this approach to a comparison of various approaches.},
	optlanguage = {en},
	number = {3},
	opturldate = {2022-09-01},
	optjournal = {Machine Learning},
	journal = {Mach. Learn.},
	author = {Heckerman, David and Geiger, Dan and Chickering, David M.},
	year = {1995},
	optkeywords = {Artificial Intelligence, Bayesian networks, Dirichlet, heuristic search, learning, likelihood equivalence, maximum branching},
	pages = {197--243},
	optfile = {Full Text PDF:C\:\\Users\\fabri\\OneDrive - mail.bbk.ac.uk\\Zotero\\storage\\QWKG2WQW\\Heckerman et al. - 1995 - Learning Bayesian networks The combination of knowledge and statistical data.pdf:application/pdf;Heckerman et al. - 1995 - Learning Bayesian Networks The Combination of Kno.pdf:C\:\\Users\\fabri\\OneDrive - mail.bbk.ac.uk\\Zotero\\storage\\M8PF9FY9\\Heckerman et al. - 1995 - Learning Bayesian Networks The Combination of Kno.pdf:application/pdf},
}

@inproceedings{Heindorf2020CauseNet,
  author    = {Heindorf, Stefan and Scholten, Yan and Wachsmuth, Henning and Ngonga Ngomo, Axel-Cyrille and Potthast, Martin},
  title     = {CauseNet: Towards a Causality Graph Extracted from the Web},
  year      = {2020},
  optisbn      = {9781450368599},
  optpublisher = {Association for Computing Machinery},
  optaddress   = {New York, NY, USA},
  opturl       = {https://doi.org/10.1145/3340531.3412763},
  optdoi       = {10.1145/3340531.3412763},
  note       = {\href{https://doi.org/10.1145/3340531.3412763}{doi: 10.1145/3340531.3412763}},
  optabstract  = {Causal knowledge is seen as one of the key ingredients to advance artificial intelligence. Yet, few knowledge bases comprise causal knowledge to date, possibly due to significant efforts required for validation. Notwithstanding this challenge, we compile CauseNet, a large-scale knowledge base of claimed causal relations between causal concepts. By extraction from different semi- and unstructured web sources, we collect more than 11 million causal relations with an estimated extraction precision of 83\% and construct the first large-scale and open-domain causality graph. We analyze the graph to gain insights about causal beliefs expressed on the web and we demonstrate its benefits in basic causal question answering. Future work may use the graph for causal reasoning, computational argumentation, multi-hop question answering, and more.},
  optbooktitle = {Proceedings of the 29th ACM International Conference on Information \& Knowledge Management},
  booktitle = {Proc. of {CIKM} 2020},
  pages     = {3023--3030},
  optnumpages  = {8},
  optkeywords  = {knowledge graph, information extraction, causality},
  optlocation  = {Virtual Event, Ireland},
  optseries    = {CIKM '20}
}

@article{glymour2019review,
  title={Review of causal discovery methods based on graphical models},
  author={Glymour, Clark and Zhang, Kun and Spirtes, Peter},
  optjournal={Front. Genet., 04 June 2019 Sec. Statistical Genetics and Methodology},
  journal={Front. Genet.},
  volume={10},
  pages={524},
  year={2019},
  optpublisher={Frontiers Media SA},
  opturl = {https://doi.org/10.3389/fgene.2019.00524},
  optdoi = {10.3389/fgene.2019.00524},
  note = {\href{https://doi.org/10.3389/fgene.2019.00524}{doi: 10.3389/fgene.2019.00524}},
}

@article{ZANGA2022survey,
  author       = {Alessio Zanga and
                  Elif Ozkirimli and
                  Fabio Stella},
  title        = {A Survey on Causal Discovery: Theory and Practice},
  journal      = {Int. J. Approx. Reason.},
  volume       = {151},
  pages        = {101--129},
  year         = {2022},
  opturl       = {https://doi.org/10.1016/j.ijar.2022.09.004},
  optdoi          = {10.1016/j.ijar.2022.09.004},
  note          = {\href{https://doi.org/10.1016/j.ijar.2022.09.004}{doi: 10.1016/j.ijar.2022.09.004}},
  bibsource    = {dblp computer science bibliography, https://dblp.org}
}

@article{ramsey2017million,
  author       = {Joseph D. Ramsey and
                  Madelyn Glymour and
                  Ruben Sanchez{-}Romero and
                  Clark Glymour},
  title        = {A million variables and more: the Fast Greedy Equivalence Search algorithm
                  for learning high-dimensional graphical causal models, with an application
                  to functional magnetic resonance images},
  journal      = {Int. J. Data Sci. Anal.},
  volume       = {3},
  number       = {2},
  pages        = {121--129},
  year         = {2017},
  opturl       = {https://doi.org/10.1007/s41060-016-0032-z},
  optdoi          = {10.1007/s41060-016-0032-z},
  note          = {\href{https://doi.org/10.1007/s41060-016-0032-z}{doi: 10.1007/s41060-016-0032-z}},
  bibsource    = {dblp computer science bibliography, https://dblp.org}
}

@article{chickering2002optimal,
  author       = {David Maxwell Chickering},
  title        = {Optimal Structure Identification With Greedy Search},
  journal      = {J. Mach. Learn. Res.},
  volume       = {3},
  pages        = {507--554},
  year         = {2002},
  note         = {\href{https://jmlr.org/papers/v3/chickering02b.html}{JMLR}},
  opturl      = {https://jmlr.org/papers/v3/chickering02b.html},
  bibsource    = {dblp computer science bibliography, https://dblp.org}
}

@article{tsamardinos2006MMHC,
  author       = {Ioannis Tsamardinos and
                  Laura E. Brown and
                  Constantin F. Aliferis},
  title        = {The max-min hill-climbing Bayesian network structure learning algorithm},
  journal      = {Mach. Learn.},
  volume       = {65},
  number       = {1},
  pages        = {31--78},
  year         = {2006},
  opturl       = {https://doi.org/10.1007/s10994-006-6889-7},
  optdoi          = {10.1007/s10994-006-6889-7},
  note          = {\href{https://doi.org/10.1007/s10994-006-6889-7}{doi: 10.1007/s10994-006-6889-7}},
  bibsource    = {dblp computer science bibliography, https://dblp.org}
}

@inproceedings{zhao2023largelanguagemodelscommonsense,
  title         = {Large Language Models as Commonsense Knowledge for Large-Scale Task Planning},
  author        = {Zhao, Zirui and Lee, Wee Sun and Hsu, David},
  year          = {2023},
  optbooktitle = {Advances in Neural Information Processing Systems},
  booktitle    = {Proc. of {NeurIPS} 2023},
  opteprint    = {2305.14078},
  optarchiveprefix = {arXiv},
  optprimaryclass  = {cs.RO},
  opturl        = {https://papers.nips.cc/paper_files/paper/2023/hash/65a39213d7d0e1eb5d192aa77e77eeb7-Abstract-Conference.html},
  note          = {\href{https://papers.nips.cc/paper_files/paper/2023/hash/65a39213d7d0e1eb5d192aa77e77eeb7-Abstract-Conference.html}{NeurIPS proceedings}}
}

@inproceedings{wan_large_2025,
  title      = {Large Language Models for Causal Discovery: Current Landscape and Future Directions},
  opturl     = {http://arxiv.org/abs/2402.11068},
  optdoi        = {10.24963/ijcai.2025/1186},
  note        = {\href{https://doi.org/10.24963/ijcai.2025/1186}{doi: 10.24963/ijcai.2025/1186}},
  optdoi     = {10.48550/arXiv.2402.11068},
  optshorttitle = {Large Language Models for Causal Discovery},
  optnumber  = {arXiv:2402.11068},
  optpublisher = {arXiv},
  author     = {Wan, Guangya and Lu, Yunsheng and Wu, Yuqi and Hu, Mengxuan and Li, Sheng},
  year       = {2025},
  optmonth   = {feb},
  booktitle  = {Proc. of {IJCAI} 2025},
  pages      = {10687--10695}
}

@inproceedings{brown2020languagemodelsfewshotlearners,
  title         = {Language Models are Few-Shot Learners},
  author        = {Tom B. Brown and Benjamin Mann and Nick Ryder and Melanie Subbiah and Jared Kaplan and Prafulla Dhariwal and Arvind Neelakantan and Pranav Shyam and Girish Sastry and Amanda Askell and Sandhini Agarwal and Ariel Herbert-Voss and Gretchen Krueger and Tom Henighan and Rewon Child and Aditya Ramesh and Daniel M. Ziegler and Jeffrey Wu and Clemens Winter and Christopher Hesse and Mark Chen and Eric Sigler and Mateusz Litwin and Scott Gray and Benjamin Chess and Jack Clark and Christopher Berner and Sam McCandlish and Alec Radford and Ilya Sutskever and Dario Amodei},
  year          = {2020},
  booktitle     = {Proc. of {NeurIPS} 2020},
  opteprint     = {2005.14165},
  optarchiveprefix = {arXiv},
  optprimaryclass  = {cs.CL},
  opturl        = {https://proceedings.neurips.cc/paper/2020/hash/1457c0d6bfcb4967418bfb8ac142f64a-Abstract.html},
  note          = {\href{https://proceedings.neurips.cc/paper/2020/hash/1457c0d6bfcb4967418bfb8ac142f64a-Abstract.html}{NeurIPS proceedings}}
}

@article{kiciman2023causal,
  title         = {Causal Reasoning and Large Language Models: Opening a New Frontier for Causality},
  author        = {Emre K{\i}c{\i}man and Robert Ness and Amit Sharma and Chenhao Tan},
  year          = {2024},
  journal       = {Trans. Mach. Learn. Res.},
  optyear       = {2023},
  opteprint     = {2305.00050},
  optarchiveprefix = {arXiv},
  optprimaryclass  = {cs.AI},
  opturl        = {https://openreview.net/forum?id=mqoxLkX210},
  note          = {\href{https://openreview.net/forum?id=mqoxLkX210}{OpenReview}}
}

@inproceedings{vashishtha2025causal,
  author    = {Vashishtha, Aniket and Reddy, Abbavaram Gowtham and Kumar, Abhinav and Bachu, Saketh and Balasubramanian, Vineeth N. and Sharma, Amit},
  title     = {Causal Order: The Key to Leveraging Imperfect Experts in Causal Inference},
  optbooktitle = {2025 International Conference on Learning Representations},
  booktitle = {Proc. of {ICLR} 2025},
  year      = {2025},
  optmonth  = {April},
  optabstract  = {Large Language Models (LLMs) have recently been used as experts to infer causal graphs, often by repeatedly applying a pairwise prompt that asks about the causal relationship of each variable pair. However, such experts, including human domain experts, cannot distinguish between direct and indirect effects given a pairwise prompt. Therefore, instead of the graph, we propose that causal order be used as a more stable output interface for utilizing expert knowledge. When querying a perfect expert with a pairwise prompt, we show that the inferred graph can have significant errors whereas the causal order is always correct. In practice, however, LLMs are imperfect experts and we find that pairwise prompts lead to multiple cycles and do not yield a valid order. Hence, we propose a prompting strategy that introduces an auxiliary variable for every variable pair and instructs the LLM to avoid cycles within this triplet. We show, both theoretically and empirically, that such a triplet prompt leads to fewer cycles than the pairwise prompt. Across multiple real-world graphs, the triplet prompt yields a more accurate order using both LLMs and human annotators as experts. By querying the expert with different auxiliary variables for the same variable pair, it also increases robustness---triplet method with much smaller models such as Phi-3 and Llama-3 8B outperforms a pairwise prompt with GPT-4. For practical usage, we show how the estimated causal order from the triplet method can be used to reduce error in downstream discovery and effect inference tasks.},
  note={\href{https://openreview.net/forum?id=9juyeCqL0u}{OpenReview}}
}

@article{chen2023mitigatingpriorerrorscausal,
  author={Chen, Lyuzhou and Ban, Taiyu and Wang, Xiangyu and Lyu, Derui and Chen, Huanhuan},
  journal  = {IEEE Trans. Pattern Anal. Mach. Intell.},
  title={Mitigating Prior Errors in Causal Structure Learning: A Resilient Approach via Bayesian Networks}, 
  year={2025},
  volume={47},
  number={12},
  pages={10990-11002},
  optdoi={10.1109/TPAMI.2025.3594755},
  note={\href{https://doi.org/10.1109/TPAMI.2025.3594755}{doi: 10.1109/TPAMI.2025.3594755}}
}

@article{ban2023query,
  author    = {Ban, T. and Chen, L. and Lyu, D. and Wang, X. and Zhu, Q. and Tu, Q. and Chen, H.},
  title     = {Integrating Large Language Model for Improved Causal Discovery},
  journal   = {IEEE Trans. on Art. Int.},
  year      = {2025},
  volume    = {6},
  number    = {11},
  pages     = {3030--3042},
  optdoi       = {10.1109/tai.2025.3560927},
  opturl       = {https://doi.org},
  optpublisher = {IEEE},
  optissn      = {2691-4581},
  note = {\href{https://doi.org/10.1109/tai.2025.3560927}{doi: 10.1109/tai.2025.3560927}},
}

@inproceedings{cohrs2023large,
  title     = {Large Language Models for Constrained-Based Causal Discovery},
  author    = {Kai-Hendrik Cohrs and Emiliano Diaz and Vasileios Sitokonstantinou and Gherardo Varando and Gustau Camps-Valls},
  optbooktitle = {AAAI 2024 Workshop on ''Are Large Language Models Simply Causal Parrots?''},
  booktitle = {Proc. of {AAAI} Workshop on Large Language Models and Causality 2024},
  year      = {2023},
  note      = {\href{https://openreview.net/forum?id=NEAoZRWHPN}{OpenReview}}
}

@article{khatibi2024alcmautonomousllmaugmentedcausal,
  title         = {ALCM: Autonomous LLM-Augmented Causal Discovery Framework},
  author        = {Elahe Khatibi and Mahyar Abbasian and Zhongqi Yang and Iman Azimi and Amir M. Rahmani},
  year          = {2024},
  journal       = {arXiv preprint \href{https://arxiv.org/abs/2405.01744}{arXiv:2405.01744}},
  opteprint     = {2405.01744},
  optarchiveprefix = {arXiv},
  optprimaryclass  = {cs.LG},
  opturl        = {https://arxiv.org/abs/2405.01744}
}

@inproceedings{abdulaal2024causal,
  title     = {Causal Modelling Agents: Causal Graph Discovery through Synergising Metadata- and Data-driven Reasoning},
  author    = {Ahmed Abdulaal and adamos hadjivasiliou and Nina Montana-Brown and Tiantian He and Ayodeji Ijishakin and Ivana Drobnjak and Daniel C. Castro and Daniel C. Alexander},
  optbooktitle = {The Twelfth International Conference on Learning Representations},
  booktitle = {Proc. of {ICLR} 2024},
  year      = {2024},
  note      = {\href{https://openreview.net/forum?id=pAoqRlTBtY}{OpenReview}}
}

@inproceedings{gkountouras2024languageagentsmeetcausality,
  title         = {Language Agents Meet Causality -- Bridging LLMs and Causal World Models},
  author        = {John Gkountouras and Matthias Lindemann and Phillip Lippe and Efstratios Gavves and Ivan Titov},
  year          = {2025},
  booktitle     = {Proc. of {ICLR} 2025},
  optyear       = {2024},
  opteprint     = {2410.19923},
  optarchiveprefix = {arXiv},
  optprimaryclass  = {cs.AI},
  note          = {\href{https://openreview.net/forum?id=y9A2TpaGsE}{OpenReview}}
}

@inproceedings{liu2024discoveryhiddenworldlarge,
  title         = {Discovery of the Hidden World with Large Language Models},
  author        = {Chenxi Liu and Yongqiang Chen and Tongliang Liu and Mingming Gong and James Cheng and Bo Han and Kun Zhang},
  year          = {2024},
  booktitle     = {Proc. of {NeurIPS} 2024},
  opteprint     = {2402.03941},
  optarchiveprefix = {arXiv},
  optprimaryclass  = {cs.LG},
  opturl        = {https://proceedings.neurips.cc/paper_files/paper/2024/hash/b99a07486702417d3b1bd64ec2cf74ad-Abstract-Conference.html},
  note          = {\href{https://proceedings.neurips.cc/paper_files/paper/2024/hash/b99a07486702417d3b1bd64ec2cf74ad-Abstract-Conference.html}{NeurIPS proceedings}}
}

@inproceedings{jin2023cladder,
  author    = {Zhijing Jin and Yuen Chen and Felix Leeb and Luigi Gresele and Ojasv Kamal and Zhiheng Lyu and Kevin Blin and Fernando Gonzalez and Max Kleiman-Weiner and Mrinmaya Sachan and Bernhard Sch{\"{o}}lkopf},
  title     = {{CL}adder: {A}ssessing Causal Reasoning in Language Models},
  year      = {2023},
  optbooktitle = {NeurIPS},
  booktitle = {Proc. of {NeurIPS} 2023},
  note       = {\href{https://openreview.net/forum?id=e2wtjx0Yqu}{OpenReview}}
}

@article{ze2023causal,
  title   = {{Causal Parrots: Large Language Models May Talk Causality But Are Not Causal}},
  author  = {Matej Ze{\v{c}}evi{\'c} and Moritz Willig and Devendra Singh Dhami and Kristian Kersting},
  optjournal = {Transactions on Machine Learning Research},
  journal = {Trans. Mach. Learn. Res.},
  optissn = {2835-8856},
  year    = {2023},
  opturl  = {https://openreview.net/forum?id=tv46tCzs83},
  note    = {\href{https://openreview.net/forum?id=tv46tCzs83}{OpenReview}}
}

@article{wu2025llmdiscovercausalityrestricted,
  title         = {{LLM Cannot Discover Causality, and Should Be Restricted to Non-Decisional Support in Causal Discovery}},
  author        = {Xingyu Wu and Kui Yu and Jibin Wu and Kay Chen Tan},
  year          = {2025},
  journal       = {arXiv preprint \href{https://arxiv.org/abs/2506.00844}{arXiv:2506.00844}},
  opteprint     = {2506.00844},
  optarchiveprefix = {arXiv},
  optprimaryclass  = {cs.LG},
  opturl        = {https://arxiv.org/abs/2506.00844}
}

@misc{instructor,
  author       = {{567-labs}},
  title        = {Instructor: Structured Outputs for {LLMs}},
  howpublished = {\href{https://github.com/567-labs/instructor}{GitHub repository}},
  optnote      = {GitHub repository},
  year         = {2025},
  opturldate   = {2025-09-27}
}

@article{colombo2014order,
  author       = {Diego Colombo and
                  Marloes H. Maathuis},
  title        = {Order-independent constraint-based causal structure learning},
  journal      = {J. Mach. Learn. Res.},
  volume       = {15},
  number       = {1},
  pages        = {3741--3782},
  year         = {2014},
  opturl       = {https://dl.acm.org/doi/10.5555/2627435.2750365},
  note         = {\href{https://jmlr.org/papers/v15/colombo14a.html}{JMLR}},
  bibsource    = {dblp computer science bibliography, https://dblp.org}
}

@misc{anthropicPromptImprover2024,
  author = {Anthropic},
  title = {Prompt Improver},
  year = {2024},
  opturl = {https://www.anthropic.com/news/prompt-improver},
  note = {\href{https://www.anthropic.com/news/prompt-improver}{Anthropic website}}
}

@inproceedings{tam-etal-2024-speak,
    title = "{Let Me Speak Freely? A Study On The Impact Of Format Restrictions On Large Language Model Performance.}",
    author = "Tam, Zhi Rui  and
      Wu, Cheng-Kuang  and
      Tsai, Yi-Lin  and
      Lin, Chieh-Yen  and
      Lee, Hung-yi  and
      Chen, Yun-Nung",
    opteditor = "Dernoncourt, Franck  and
      Preo{\c{t}}iuc-Pietro, Daniel  and
      Shimorina, Anastasia",
    optbooktitle = "Proceedings of the 2024 Conference on Empirical Methods in Natural Language Processing: Industry Track",
    booktitle = "Proc. of {EMNLP} 2024",
    optmonth = nov,
    year = "2024",
    optaddress = "Miami, Florida, US",
    optpublisher = "Association for Computational Linguistics",
    opturl = "https://aclanthology.org/2024.emnlp-industry.91/",
    optdoi = "10.18653/v1/2024.emnlp-industry.91",
    pages = "1218--1236",
    note = "\href{https://aclanthology.org/2024.emnlp-industry.91/}{ACL Anthology}",
    optabstract = "Structured generation, the process of producing content in standardized formats like JSON and XML, is widely utilized in real-world applications to extract key output information from large language models (LLMs).This study investigates whether such constraints on generation space impact LLMs' abilities, including reasoning and domain knowledge comprehension. Specifically, we evaluate LLMs' performance when restricted to adhere to structured formats versus generating free-form responses across various common tasks. Surprisingly, we observe a significant decline in LLMs' reasoning abilities under format restrictions. Furthermore, we find that stricter format constraints generally lead to greater performance degradation in reasoning tasks."
}

@inproceedings{zheng2018dagstearscontinuousoptimization,
  author       = {Xun Zheng and
                  Bryon Aragam and
                  Pradeep Ravikumar and
                  Eric P. Xing},
  optopteditor       = {Samy Bengio and
                  Hanna M. Wallach and
                  Hugo Larochelle and
                  Kristen Grauman and
                  Nicol{\`{o}} Cesa{-}Bianchi and
                  Roman Garnett},
  title        = {DAGs with {NO} {TEARS:} Continuous imization for Structure Learning},
  optbooktitle    = {Prooceedings of the 31st Annual Conference
                  on Neural Information Processing Systems 2018, NeurIPS 2018, December
                  3-8, 2018, Montr{\'{e}}al, Canada},
  booktitle = {Proc. of {NeurIPS} 2018},
  pages        = {9492--9503},
  year         = {2018},
  opturl       = {https://proceedings.neurips.cc/paper/2018/hash/e347c51419ffb23ca3fd5050202f9c3d-Abstract.html},
  note         = {\href{https://proceedings.neurips.cc/paper/2018/hash/e347c51419ffb23ca3fd5050202f9c3d-Abstract.html}{NeurIPS proceedings}},
  bibsource    = {dblp computer science bibliography, https://dblp.org}
}

@article{Erdos:1959:pmd,
  author = {Erd\"os, P and R\'enyi, A},
  journal = {Publicationes Mathematicae Debrecen},
  optkeywords = {epidemes},
  pages = {290--297},
  title = {On Random Graphs I},
  volume = 6,
  year = 1959
}

@article{
doi:10.1126/science.286.5439.509,
author = {Albert-László Barabási  and Réka Albert },
title = {Emergence of Scaling in Random Networks},
journal = {Science},
volume = {286},
number = {5439},
pages = {509--512},
year = {1999},
optdoi = {10.1126/science.286.5439.509},
note = {\href{https://doi.org/10.1126/science.286.5439.509}{doi: 10.1126/science.286.5439.509}},
optURL = {https://www.science.org/doi/abs/10.1126/science.286.5439.509},
opteprint = {https://www.science.org/doi/pdf/10.1126/science.286.5439.509},
optabstract = {Systems as diverse as genetic networks or the World Wide Web are best described as networks with complex topology. A common property of many large networks is that the vertex connectivities follow a scale-free power-law distribution. This feature was found to be a consequence of two generic mechanisms: (i) networks expand continuously by the addition of new vertices, and (ii) new vertices attach preferentially to sites that are already well connected. A model based on these two ingredients reproduces the observed stationary scale-free distributions, which indicates that the development of large networks is governed by robust self-organizing phenomena that go beyond the particulars of the individual systems.}}

@article{cordella2004subgraph,
  author   = {Cordella, L.P. and Foggia, P. and Sansone, C. and Vento, M.},
  optjournal = {IEEE Transactions on Pattern Analysis and Machine Intelligence},
  journal  = {IEEE Trans. Pattern Anal. Mach. Intell.},
  title    = {A (sub)graph isomorphism algorithm for matching large graphs},
  year     = {2004},
  volume   = {26},
  number   = {10},
  pages    = {1367-1372},
  optkeywords = {Pattern recognition;Pattern matching;Pattern analysis;Application software;NP-complete problem;Performance analysis;Algorithm design and analysis;Testing;Performance evaluation;Relational databases;Index Terms- Graph-subgraph isomorphism;large graphs;attributed relational graphs.},
  optdoi      = {10.1109/TPAMI.2004.75},
  note      = {\href{https://doi.org/10.1109/TPAMI.2004.75}{doi: 10.1109/TPAMI.2004.75}},
}

@Article{bnlearn,
    title = {Learning Bayesian Networks with the {bnlearn} {R}
      Package},
    author = {Marco Scutari},
    optjournal = {Journal of Statistical Software},
    journal = {J. Stat. Softw.},
    year = {2010},
    volume = {35},
    number = {3},
    pages = {1--22},
    optdoi = {10.18637/jss.v035.i03},
    note = {\href{https://doi.org/10.18637/jss.v035.i03}{doi: 10.18637/jss.v035.i03}},
  }

@article{10.1111/j.2517-6161.1988.tb01721.x,
    author = {Lauritzen, S. L. and Spiegelhalter, D. J.},
    title = {Local Computations with Probabilities on Graphical Structures and Their Application to Expert Systems},
    optjournal = {Journal of the Royal Statistical Society: Series B (Methodological)},
    journal = {J. R. Stat. Soc. Ser. B Methodol.},
    volume = {50},
    number = {2},
    pages = {157--194},
    year = {1988},
    optmonth = {12},
    optabstract = {A causal network is used in a number of areas as a depiction of patterns of ‘influence’ among sets of variables. In expert systems it is common to perform ‘inference’ by means of local computations on such large but sparse networks. In general, non-probabilistic methods are used to handle uncertainty when propagating the effects of evidence, and it has appeared that exact probabilistic methods are not computationally feasible. Motivated by an application in electromyography, we counter this claim by exploiting a range of local representations for the joint probability distribution, combined with topological changes to the original network termed ‘marrying’ and ‘filling-in‘. The resulting structure allows efficient algorithms for transfer between representations, providing rapid absorption and propagation of evidence. The scheme is first illustrated on a small, fictitious but challenging example, and the underlying theory and computational aspects are then discussed.},
    optissn = {0035-9246},
    optdoi = {10.1111/j.2517-6161.1988.tb01721.x},
    note = {\href{https://doi.org/10.1111/j.2517-6161.1988.tb01721.x}{doi: 10.1111/j.2517-6161.1988.tb01721.x}},
    opturl = {https://doi.org/10.1111/j.2517-6161.1988.tb01721.x},
    opteprint = {https://academic.oup.com/jrsssb/article-pdf/50/2/157/49097926/jrsssb_50_2_157.pdf},
}

@article{
doi:10.1126/science.1105809,
author = {Karen Sachs  and Omar Perez  and Dana Pe'er  and Douglas A. Lauffenburger  and Garry P. Nolan },
title = {Causal Protein-Signaling Networks Derived from Multiparameter Single-Cell Data},
journal = {Science},
volume = {308},
number = {5721},
pages = {523--529},
year = {2005},
optdoi = {10.1126/science.1105809},
note = {\href{https://doi.org/10.1126/science.1105809}{doi: 10.1126/science.1105809}},
optURL = {https://www.science.org/doi/abs/10.1126/science.1105809},
opteprint = {https://www.science.org/doi/pdf/10.1126/science.1105809},
optabstract = {Machine learning was applied for the automated derivation of causal influences in cellular signaling networks. This derivation relied on the simultaneous measurement of multiple phosphorylated protein and phospholipid components in thousands of individual primary human immune system cells. Perturbing these cells with molecular interventions drove the ordering of connections between pathway components, wherein Bayesian network computational methods automatically elucidated most of the traditionally reported signaling relationships and predicted novel interpathway network causalities, which we verified experimentally. Reconstruction of network models from physiologically relevant primary single cells might be applied to understanding native-state tissue signaling biology, complex drug actions, and dysfunctional signaling in diseased cells.}}

@misc{spiegelhalter1992learning,
  author       = {D. J. Spiegelhalter and R. G. Cowell},
  title        = {Learning in Probabilistic Expert Systems},
  howpublished = {In \emph{Bayesian Statistics 4}, pp. 447--466},
  year         = {1992},
  note         = {\href{https://doi.org/10.1093/oso/9780198522669.003.0025}{doi: 10.1093/oso/9780198522669.003.0025}}
}

@inproceedings{dong2024generalizationmemorizationdatacontamination,
      title={Generalization or Memorization: Data Contamination and Trustworthy Evaluation for Large Language Models}, 
      author={Yihong Dong and Xue Jiang and Huanyu Liu and Zhi Jin and Bin Gu and Mengfei Yang and Ge Li},
      year={2024},
      opteprint={2402.15938},
      optarchivePrefix={arXiv},
      optprimaryClass={cs.CL},
      opturl={https://arxiv.org/abs/2402.15938}, 
      booktitle={Findings of {ACL} 2024},
      pages={12039--12050},
      optdoi={10.18653/V1/2024.FINDINGS-ACL.716},
      note={\href{https://doi.org/10.18653/V1/2024.FINDINGS-ACL.716}{doi: 10.18653/V1/2024.FINDINGS-ACL.716}},
}

@inproceedings{balloccu2024leakcheatrepeatdata,
      title={Leak, Cheat, Repeat: Data Contamination and Evaluation Malpractices in Closed-Source LLMs}, 
      author={Simone Balloccu and Patrícia Schmidtová and Mateusz Lango and Ondřej Dušek},
      year={2024},
      opteprint={2402.03927},
      optarchivePrefix={arXiv},
      optprimaryClass={cs.CL},
      opturl={https://arxiv.org/abs/2402.03927},
      booktitle={Proc. of {EACL} 2024},
      pages={67--93},
      optdoi={10.18653/V1/2024.EACL-LONG.5},
      note={\href{https://doi.org/10.18653/V1/2024.EACL-LONG.5}{doi: 10.18653/V1/2024.EACL-LONG.5}},
}

@inproceedings{zheng2020notears_mlp,
  author       = {Xun Zheng and
                  Chen Dan and
                  Bryon Aragam and
                  Pradeep Ravikumar and
                  Eric P. Xing},
  opteditor       = {Silvia Chiappa and
                  Roberto Calandra},
  title        = {Learning Sparse Nonparametric DAGs},
  optbooktitle    = {The 23rd International Conference on Artificial Intelligence and Statistics,
                  {AISTATS} 2020, 26-28 August 2020, Online [Palermo, Sicily, Italy]},
  booktitle = {Proc. of {AISTATS} 2020},
  optseries       = {Proceedings of Machine Learning Research},
  series = {{PMLR}},
  volume       = {108},
  pages        = {3414--3425},
  optpublisher = {{PMLR}},
  year         = {2020},
  opturl       = {http://proceedings.mlr.press/v108/zheng20a.html},
  note         = {\href{http://proceedings.mlr.press/v108/zheng20a.html}{PMLR}},
  bibsource    = {dblp computer science bibliography, https://dblp.org}
}

@article{vowels2021doyalike_survey,
author = {Vowels, Matthew J. and Camgoz, Necati Cihan and Bowden, Richard},
title = {D’ya {Like} {DAGs}? {A} {Survey} on {Structure} {Learning} and {Causal} {Discovery}},
year = {2022},
optissue_date = {April 2023},
optpublisher = {Association for Computing Machinery},
optaddress = {New York, NY, USA},
volume = {55},
number = {4},
optissn = {0360-0300},
opturl = {https://doi.org/10.1145/3527154},
optdoi = {10.1145/3527154},
note = {\href{https://doi.org/10.1145/3527154}{doi: 10.1145/3527154}},
optabstract = {Causal reasoning is a crucial part of science and human intelligence. In order to discover causal relationships from data, we need structure discovery methods. We provide a review of background theory and a survey of methods for structure discovery. We primarily focus on modern, continuous optimization methods, and provide reference to further resources such as benchmark datasets and software packages. Finally, we discuss the assumptive leap required to take us from structure to causality.},
journal = {ACM Comput. Surv.},
optmonth = nov,
articleno = {82},
pages = {1--35},
optnumpages = {36},
optkeywords = {Causality, causal discovery, directed acyclic graphs, DAGs, structure learning, survey}
}

@inproceedings{jiralerspong2024efficient,
  title={Efficient Causal Graph Discovery Using Large Language Models},
  author={Jiralerspong, Thomas and Chen, Xiaoyin and More, Yash and Shah, Vedant and Bengio, Yoshua},
  optbooktitle={ICLR 2024 Workshop: How Far Are We From AGI},
  booktitle={Proc. of {ICLR} Workshop on How Far Are We From {AGI} 2024},
  year={2024},
  note={\href{https://openreview.net/forum?id=5RBUTx75yr}{OpenReview}}
}

@article{GebserKKS17,
  author       = {Martin Gebser and
                  Roland Kaminski and
                  Benjamin Kaufmann and
                  Torsten Schaub},
  title        = {Multi-shot {ASP} solving with clingo},
  journal      = {Theory Pract. Log. Program.},
  volume       = {19},
  number       = {1},
  pages        = {27--82},
  year         = {2019},
  opturl       = {https://doi.org/10.1017/S1471068418000054},
  optdoi          = {10.1017/S1471068418000054},
  note          = {\href{https://doi.org/10.1017/S1471068418000054}{doi: 10.1017/S1471068418000054}},
  bibsource    = {dblp computer science bibliography, https://dblp.org}
}

@inproceedings{pyAgrum,
      TITLE = {{aGrUM/pyAgrum : a Toolbox to Build Models and Algorithms for Probabilistic Graphical Models in Python}},
      AUTHOR = {Ducamp, Gaspard and Gonzales, Christophe and Wuillemin, Pierre-Henri},
      OPTURL = {https://hal.archives-ouvertes.fr/hal-03135721},
      OPTBOOKTITLE = {{10th International Conference on Probabilistic Graphical Models}},
      BOOKTITLE = {Proc. of {PGM} 2020},
      OPTADDRESS = {Sk{{\o}}rping, Denmark},
      SERIES = {{PMLR}},
      VOLUME = {138},
      PAGES = {609-612},
      YEAR = {2020},
      OPTMONTH = Sep,
      OPTKEYWORDS = {C++ ; Python ; Probabilistic Graphical Models ; Bayesian Networks},
      OPTPDF = {https://hal.archives-ouvertes.fr/hal-03135721/file/ducamp20a.pdf},
      OPTHAL_ID = {hal-03135721},
      OPTHAL_VERSION = {v1},
      NOTE = {\href{https://hal.archives-ouvertes.fr/hal-03135721}{HAL}},
    }

@article{Toni14tutorialABA,
  author       = {Francesca Toni},
  title        = {A tutorial on assumption-based argumentation},
  journal      = {Argument Comput.},
  volume       = {5},
  number       = {1},
  pages        = {89--117},
  year         = {2014},
  opturl       = {https://doi.org/10.1080/19462166.2013.869878},
  optdoi          = {10.1080/19462166.2013.869878},
  note          = {\href{https://doi.org/10.1080/19462166.2013.869878}{doi: 10.1080/19462166.2013.869878}},
  bibsource    = {dblp computer science bibliography, https://dblp.org}
}

@article{Dung95,
  author       = {Phan Minh Dung},
  title        = {On the Acceptability of Arguments and its Fundamental Role in Nonmonotonic
                  Reasoning, Logic Programming and n-Person Games},
  journal      = {Artif. Intell.},
  volume       = {77},
  number       = {2},
  pages        = {321--358},
  year         = {1995},
  opturl       = {https://doi.org/10.1016/0004-3702(94)00041-X},
  optdoi          = {10.1016/0004-3702(94)00041-X},
  note          = {\href{https://doi.org/10.1016/0004-3702(94)00041-X}{doi: 10.1016/0004-3702(94)00041-X}},
  bibsource    = {dblp computer science bibliography, https://dblp.org}
}

@article{baroni_introduction_2011,
  author       = {Pietro Baroni and
                  Martin Caminada and
                  Massimiliano Giacomin},
  title        = {An introduction to argumentation semantics},
  journal      = {Knowl. Eng. Rev.},
  volume       = {26},
  number       = {4},
  pages        = {365--410},
  year         = {2011},
  opturl       = {https://doi.org/10.1017/S0269888911000166},
  optdoi          = {10.1017/S0269888911000166},
  note          = {\href{https://doi.org/10.1017/S0269888911000166}{doi: 10.1017/S0269888911000166}},
  bibsource    = {dblp computer science bibliography, https://dblp.org}
}

@article{Spearman1904rank,
 optISSN = {00029556},
 optURL = {http://www.jstor.org/stable/1412159},
 author = {C. Spearman},
 optjournal = {The American Journal of Psychology},
 journal = {Am. J. Psychol.},
 number = {1},
 pages = {72--101},
 optpublisher = {University of Illinois Press},
 title = {The Proof and Measurement of Association between Two Things},
 opturldate = {2025-10-02},
 volume = {15},
 year = {1904},
 note = {\href{http://www.jstor.org/stable/1412159}{JSTOR}}
}

@article{peters2015structural,
  author       = {Jonas Peters and
                  Peter B{\"{u}}hlmann},
  title        = {Structural Intervention Distance for Evaluating Causal Graphs},
  journal      = {Neural Comput.},
  volume       = {27},
  number       = {3},
  pages        = {771--799},
  year         = {2015},
  opturl       = {https://doi.org/10.1162/NECO\_a\_00708},
  optdoi          = {10.1162/NECO\_a\_00708},
  note          = {\href{https://doi.org/10.1162/NECO_a_00708}{doi: 10.1162/NECO\_a\_00708}},
  bibsource    = {dblp computer science bibliography, https://dblp.org}
}

@book{agresti2018introduction,
  title     = {An Introduction to Categorical Data Analysis},
  author    = {Agresti, Alan},
  edition   = {3},
  publisher = {Wiley},
  year      = {2018}
}

@inproceedings{lam2022grasp,
  title     = {Greedy Relaxations of the Sparsest Permutation Algorithm},
  author    = {Lam, Wai-Yin and Andrews, Bryan and Ramsey, Joseph},
  optbooktitle = {Proceedings of the 38th Conference on Uncertainty in Artificial Intelligence (UAI)},
  booktitle = {Proc. of {UAI} 2022},
  year      = {2022},
  note      = {\href{https://proceedings.mlr.press/v180/lam22a.html}{PMLR}}
}

@inproceedings{andrews2023boss,
  title   = {Fast Scalable and Accurate Discovery of {DAGs} Using the Best Order Score Search and Grow--Shrink Trees},
  author  = {Andrews, Bryan and Ramsey, Joseph and Sanchez Romero, Ruben and Camchong, Jazmin and Kummerfeld, Erich},
  optjournal = {Advances in Neural Information Processing Systems},
  optvolume  = {36},
  booktitle = {Proc. of {NeurIPS} 2023},
  year    = {2023},
  note        = {\href{https://proceedings.neurips.cc/paper_files/paper/2023/file/c9cde817d04811ba28e44071bd9f76a5-Paper-Conference.pdf}{NeurIPS proceedings}}
}

@inproceedings{reisach2021beware,
  title        = {Beware of the Simulated {DAG}! Causal Discovery Benchmarks May Be Easy to Game},
  author       = {Reisach, Alexander G. and Seiler, Christof and Weichwald, Sebastian},
  year         = {2021},
  opteprint    = {2102.13647},
  optarchivePrefix= {arXiv},
  optprimaryClass = {stat.ML},
  booktitle    = {Proc. of {NeurIPS} 2021},
  pages        = {27772--27784},
  note         = {\href{https://proceedings.neurips.cc/paper/2021/hash/e987eff4a7c7b7e580d659feb6f60c1a-Abstract.html}{NeurIPS proceedings}}
}

@article{benjamini1995controlling,
  title={Controlling the false discovery rate: a practical and powerful approach to multiple testing},
  author={Benjamini, Yoav and Hochberg, Yosef},
  optjournal={Journal of the Royal Statistical Society: Series B (Methodological)},
  journal={J. R. Stat. Soc. Ser. B Methodol.},
  volume={57},
  number={1},
  pages={289--300},
  year={1995},
  optpublisher={Wiley Online Library},
  note={\href{https://www.jstor.org/stable/2346101}{JSTOR}}
}

@inproceedings{verma1990equivalence,
author = {Verma, Thomas and Pearl, Judea},
title = {Equivalence and synthesis of causal models},
year = {1990},
optisbn = {0444892648},
optpublisher = {Elsevier Science Inc.},
optaddress = {USA},
optbooktitle = {Proceedings of the Sixth Annual Conference on Uncertainty in Artificial Intelligence},
booktitle = {Proc. of {UAI} 1990},
pages = {255--270},
optnumpages = {16},
optseries = {UAI '90},
note = {\href{https://dl.acm.org/doi/10.5555/647233.719736}{doi: 10.5555/647233.719736}}
}

@inproceedings{cao2024causal,
  title={Causal Discovery with Interactive Human Inputs},
  author={Cao, Jin and Liu, Renxiong},
  optbooktitle={World Congress in Computer Science, Computer Engineering \& Applied Computing},
  booktitle={Proc. of {CSCE} 2024},
  pages={241--256},
  year={2024},
  optorganization={Springer},
  note={\href{https://doi.org/10.1007/978-3-031-85628-0_18}{doi: 10.1007/978-3-031-85628-0-18}}
}

@inproceedings{havrilla2025igda,
  title     = {{IGDA}: Interactive Graph Discovery Through Large Language Model Agents},
  author    = {Alexander Havrilla and David Alvarez-Melis and Nicolo Fusi},
  booktitle = {Proc. of the {ICLR} 2025 Workshop on Reasoning and Planning for Large Language Models},
  year      = {2025},
  opturl    = {https://openreview.net/forum?id=cHV3Iw84AC},
  note      = {\href{https://openreview.net/forum?id=cHV3Iw84AC}{OpenReview}}
}

@inproceedings{ankan2025expertloop,
	title = {Expert-{In}-{The}-{Loop} {Causal} {Discovery}: {Iterative} {Model} {Refinement} {Using} {Expert} {Knowledge}},
	optissn = {2640-3498},
	optshorttitle = {Expert-{In}-{The}-{Loop} {Causal} {Discovery}},
	opturl = {https://proceedings.mlr.press/v286/ankan25a.html},
	optabstract = {Many researchers construct directed acyclic graph (DAG) models manually based on domain knowledge. Although numerous causal discovery algorithms were developed to automatically learn DAGs and other causal models from data, these remain challenging to use due to their tendency to produce results that contradict domain knowledge, among other issues. Here we propose a hybrid, iterative structure learning approach that combines domain knowledge with data-driven insights to assist researchers in constructing DAGs. Our method leverages conditional independence testing to iteratively identify variable pairs where an edge is either missing or superfluous. Based on this information, we can choose to add missing edges with appropriate orientation based on domain knowledge or remove unnecessary ones. We also give a method to rank these missing edges based on their impact on the overall model fit. In a simulation study, we find that this iterative approach to leverage domain knowledge already starts outperforming purely data-driven structure learning if the orientation of new edge is correctly determined in at least two out of three cases. We present a proof-of-concept implementation using a large language model as a domain expert and a graphical user interface designed to assist human experts with DAG construction.},
	optlanguage = {en},
	opturldate = {2025-10-19},
	optbooktitle = {Proceedings of the {Forty}-first {Conference} on {Uncertainty} in {Artificial} {Intelligence}},
	booktitle = {Proc. of {UAI} 2025},
	optpublisher = {PMLR},
	author = {Ankan, Ankur and Textor, Johannes},
	optmonth = jul,
	year = {2025},
	pages = {172--183},
	note = {\href{https://proceedings.mlr.press/v286/ankan25a.html}{PMLR}},
	optfile = {Full Text PDF:C\:\\Users\\fabri\\OneDrive - mail.bbk.ac.uk\\Zotero\\storage\\R7PCABKX\\Ankan and Textor - 2025 - Expert-In-The-Loop Causal Discovery Iterative Model Refinement Using Expert Knowledge.pdf:application/pdf},
}

@article{kitson_causal_2023,
	title = {Causal discovery using dynamically requested knowledge},
	opturl = {http://arxiv.org/abs/2310.11154},
	optdoi = {10.1016/j.knosys.2025.113185},
	note = {\href{https://doi.org/10.1016/j.knosys.2025.113185}{doi: 10.1016/j.knosys.2025.113185}},
	optdoi = {10.48550/arXiv.2310.11154},
	optabstract = {Causal Bayesian Networks (CBNs) are an important tool for reasoning under uncertainty in complex real-world systems. Determining the graphical structure of a CBN remains a key challenge and is undertaken either by eliciting it from humans, using machine learning to learn it from data, or using a combination of these two approaches. In the latter case, human knowledge is generally provided to the algorithm before it starts, but here we investigate a novel approach where the structure learning algorithm itself dynamically identifies and requests knowledge for relationships that the algorithm identifies as uncertain during structure learning. We integrate this approach into the Tabu structure learning algorithm and show that it offers considerable gains in structural accuracy, which are generally larger than those offered by existing approaches for integrating knowledge. We suggest that a variant which requests only arc orientation information may be particularly useful where the practitioner has little preexisting knowledge of the causal relationships. As well as offering improved accuracy, the approach can use human expertise more effectively and contributes to making the structure learning process more transparent.},
	opturldate = {2024-04-29},
	optpublisher = {arXiv},
	author = {Kitson, Neville K. and Constantinou, Anthony C.},
	optmonth = oct,
	year = {2025},
	journal = {Knowl. Based Syst.},
	volume = {314},
	pages = {113185},
	optnote = {arXiv:2310.11154 [cs]},
	optkeywords = {Computer Science - Artificial Intelligence},
	optfile = {arXiv.org Snapshot:C\:\\Users\\fabri\\OneDrive - mail.bbk.ac.uk\\Zotero\\storage\\DKV3KD8L\\2310.html:text/html;Kitson and Constantinou - 2023 - Causal discovery using dynamically requested knowl.pdf:C\:\\Users\\fabri\\OneDrive - mail.bbk.ac.uk\\Zotero\\storage\\4YS4TT6W\\Kitson and Constantinou - 2023 - Causal discovery using dynamically requested knowl.pdf:application/pdf},
}

\clearpage
\appendix
\onecolumn

\title{Leveraging Large Language Models for Causal Discovery: \\a Constraint-based, Argumentation-driven Approach (Supplementary Material)}
\maketitle

In this supplementary material we provide details omitted from the main text due to space limitations. In particular, we describe the prompts used for metadata enrichment and constraint elicitation (App.~\ref{sec:appendix_prompts}), the consensus strategy for stabilising LLM outputs, the generation and semantic grounding of the synthetic CauseNet datasets (App.~\ref{sec:appendix_graph_data_generation}), and the full experimental protocol, metrics, baselines, ablations, and additional quantitative analyses (App.~\ref{sec:appendix_experiment_details}).

\section{Prompting Details}
\label{sec:appendix_prompts}
This section covers the prompts used for the different tasks in which LLMs are involved, see Fig.~\ref{fig:llm-pipeline} in the main text for the pipeline overview. The following prompts were built with various iterations of Anthropic's prompt improver tool~\citep{anthropicPromptImprover2024}.

\subsection{LLM constraints elicitation prompt}\label{sec:eli_prompt}
Below, we provide the main prompt used in our proposed LLM elicitation pipeline. Note that this prompt significantly differs from prior work on LLMs and causal discovery, e.g. \citep{kiciman2023causal,jiralerspong2024efficient}, since it involves a single call for all variables in the dataset. This is enabled by the much longer context window of modern LLMs (e.g., over 1 million tokens for \texttt{gemini-2.5-flash}). This single-call approach is not only more computationally efficient but also allows the LLM to reason holistically about the entire system of variables, potentially capturing higher-order interactions that pairwise queries might miss.

\begin{lstlisting}[style=llmprompt, breaklines]
You will be analyzing a set of causal variables to determine the required and forbidden causal directions between them. Your goal is to achieve very high precision in identifying these relationships.

<causal_variables>
{CAUSAL_VARIABLES}
</causal_variables>

Your task is to analyze these causal variables and generate two sets:
1. **Required directions**: Causal relationships that must exist based on logical necessity, temporal ordering, or fundamental causal principles
2. **Forbidden directions**: Causal relationships that cannot exist due to logical impossibility, temporal constraints, or definitional contradictions

Here are the key principles to follow:

**Required directions** should include:
- Relationships where one variable definitionally or logically must cause another
- Temporal precedence relationships (causes must precede effects)
- Relationships where the causal mechanism is well-established and unavoidable

**Forbidden directions** should include:
- Relationships that would violate temporal ordering (effects cannot cause their own causes)
- Relationships that are logically contradictory or definitionally impossible
- Relationships where the direction would violate established causal mechanisms

**Important guidelines:**
- Only include relationships where you have very high confidence
- When in doubt about a relationship, do not include it in either set
- Consider both direct and indirect causal pathways
- Be precise about the direction of causality (A → B is different from B → A)

Format your final answer as follows:

**Required Directions:**
- [Variable A] → [Variable B]: [Brief justification]
- [Continue for all required directions]

**Forbidden Directions:**  
- [Variable C] → [Variable D]: [Brief justification]
- [Continue for all forbidden directions]

Your final answer should only include the Required Directions and Forbidden Directions sections with their respective causal relationships and justifications. Aim for very high precision - only include relationships where you are highly confident in the causal direction requirement or prohibition.
\end{lstlisting}

\subsection{Schema-Guided Extraction of Constraints}
\label{sec:schema-guided_extraction}

To ensure reliability and avoid constraining the LLM's reasoning process with rigid formatting requirements \citep{tam-etal-2024-speak}, we decouple constraint generation from structured output extraction. The primary LLM produces a semi-structured natural-language answer with required/forbidden directions and brief rationales; we then use a schema-enforcing tool \citep{instructor} with a lightweight model (\texttt{gemini-2.5-flash-lite}, LLM 3 in Fig.~\ref{fig:llm-pipeline}) to map this text into typed required/forbidden arrow lists. The extracted constraints are validated and normalised (e.g., variable-name matching and de-duplication) before downstream consensus filtering and integration.

\subsection{LLM Consensus Details}
\label{sec:appendix-consensus}
The consensus mechanism is designed to generate high-precision causal constraints by aggregating the outputs from multiple independent LLM queries. This is due to the inherent stochasticity of LLMs, and inspired by the majority voting strategy proposed by~\citep{cohrs2023large}. Both \texttt{gemini-2.5-flash} and \texttt{gpt-5-mini} use the same elicitation prompt and the same five-run unanimity rule.

For each causal discovery problem, the LLM is queried five times, yielding five distinct sets of constraints. Let $R_i$ and $F_i$ represent the set of ``required'' and ``forbidden'' arrows, respectively, from run $i \in \{1, \dots, 5\}$.
The final high-confidence consensus sets are obtained by taking the intersection across the five runs: $ R_{\text{consensus}} = \bigcap_{i=1}^{5} R_i $ and $ F_{\text{consensus}} = \bigcap_{i=1}^{5} F_i $.
This guarantees that an arrow appears in the final constraints only if it was proposed unanimously across all independent evaluations. The repetition count of five was chosen empirically to balance precision and recall, with enough repeats to filter out stochastic or low-confidence suggestions while avoiding excessive computational cost. The conservative intersection rule substantially improves the precision of the resulting constraints, as empirically validated in Appendix \ref{sec:appendix_constraint_eval}. Across the 54 synthetic datasets, \texttt{gemini-2.5-flash} yields 476 unanimous constraints and non-empty consensus sets for 53 datasets, while \texttt{gpt-5-mini} yields 129 unanimous constraints and non-empty consensus sets for 43 datasets.

\subsection{Metadata Enrichment Prompt (Adding Variables' Descriptions)}\label{sec:enrich_prompt}
Below, we provide the prompt used for eliciting the variable descriptions before interrogating the main LLM (LLM 1 in Fig.~\ref{fig:llm-pipeline}) about causal relations amongst variables. The following prompt is related to LLM 2 in Fig.~\ref{fig:llm-pipeline}, which we deem an optional block as its usefulness may depend on context. Ablations on the importance of variable descriptions are in \S\ref{fig:constraint_quality}. As noted in the main text, for synthetic data, this prompt receives a graph-level description generated from the ground-truth graph, making the resulting metadata a controlled high-information condition rather than a realistic documentation setting. Names-only, noisy, human-written, and obfuscated metadata are separate robustness regimes.

\begin{lstlisting}[style=llmprompt, breaklines]
You are tasked with generating descriptions for causal variables in a randomly generated causal graph used for synthetic dataset evaluations of causal discovery algorithms. Your goal is to create meaningful descriptions for each variable that accurately reflect their role in the graph without revealing their relationships with other variables.

Here is the description of the randomly generated causal graph:'

<causal_graph>
{CAUSAL_GRAPH_DESCRIPTION}
</causal_graph>

To complete this task, follow these steps:

1. Carefully read and analyze the causal graph description.

2. For each variable mentioned in the graph, create a description that:
   a) Precisely describes its meaning within the context of the graph
   b) Does NOT spoil its relationships with other variables, either explicitly or implicitly
   c) Maintains a consistent context or scenario across all variable descriptions

3. If a variable has different contextual meanings in its relationships with other variables, provide a general description that could encompass these different meanings without revealing the specific relationships.

4. After generating all variable descriptions, assess the quality of the random causal graph based on how well the variables' meanings align with each other and how realistic the overall scenario is.

5. Present your output in the following format:

<title>
[Provide a concise title for the causal graph]
</title>

<variable_descriptions>
[List each variable and its description]
</variable_descriptions>

<graph_quality_assessment>
[Provide your assessment of the graph quality, including how well the variables' meanings align and how realistic the overall scenario is]
</graph_quality_assessment>

Remember, your primary goal is to create meaningful descriptions without revealing any causal relationships. Be creative in developing a consistent context that could plausibly connect all the variables.
\end{lstlisting}

\section{Graph and Data Generation Details}
\label{sec:appendix_graph_data_generation}

The generation of our synthetic datasets is a two-stage process designed to create structurally diverse and semantically plausible causal graphs. First, we generate a structural scaffold (a DAG) using one of three different topology generation methods (\S\ref{sec:scaffold_graphs}). Second, we ground this abstract structure in real-world concepts by finding a matching sub-graph within the \texttt{CauseNet} knowledge graph, guided by specific heuristics (\S\ref{sec:ground_heuristics}). We provide an open-source UI implementing this pipeline at \href{https://clarg-group.github.io/CauseNet-graph-generator/}{https://clarg-group.github.io/CauseNet-graph-generator/}, with the full code included in our \href{https://github.com/briziorusso/ArgCausalDisco/tree/llm-uai2026-freeze/synthetic_graph_demo}{repository}. In \S\ref{sec:data_gen_schema}, we describe the schema used to generate the synthetic datasets and sample observational data from the grounded DAGs, while in \S\ref{sec:data_gen_bnlearn} we provide details on the \texttt{bnlearn} datasets used in the experiments within this supplementary material.

\subsection{Structural Scaffolding and Graph Types}\label{sec:scaffold_graphs}
We generate the initial DAG topologies using three distinct methods to ensure a variety of graph structures:

\begin{itemize}
    \item \textbf{Erd\H{o}s-Rényi (ER) and Scale-Free (SF):} These two methods are adapted from the well-established NOTEARS codebase \citep{zheng2018dagstearscontinuousoptimization}. The ER model \citep{Erdos:1959:pmd} produces graphs with a random, uniform edge distribution, while the SF model \citep{doi:10.1126/science.286.5439.509} generates graphs with power-law degree distributions, characterised by a few high-degree ``hub'' nodes.
    \item \textbf{Lower Triangle (LT):} This custom method constructs a connected DAG by directly populating a lower triangular adjacency matrix. The algorithm proceeds in two stages. First, to guarantee connectivity, it ensures every node from $1$ to $n-1$ has at least one parent by randomly adding a single incoming edge for each from a node with a lower index. Second, it distributes the remaining specified number of edges by randomly placing them in the unoccupied positions of the lower triangle. This two-step process ensures both connectivity and the desired edge density.
\end{itemize}

While the ER and SF implementations from NOTEARS can produce graphs with a slightly varied number of nodes and edges, the LT method allows for precise control over these parameters.

\subsection{Semantic Grounding and Heuristics}\label{sec:ground_heuristics}
Once a structural scaffold is generated, we imbue it with semantic meaning by finding an isomorphic sub-graph within the \texttt{CauseNet} knowledge graph~\citep{Heindorf2020CauseNet}. Since multiple isomorphic sub-graphs can exist, we employ one of three heuristics to select the most suitable candidate match.

\begin{itemize}
    \item \textbf{\texttt{none}:} This baseline heuristic serves as a control. It performs no optimisation and simply selects the first isomorphic sub-graph returned by the search algorithm.
    \item \textbf{\texttt{degrees}:} This heuristic aims to find the most specific and least central concepts. It selects the candidate sub-graph that minimises the sum of the in-degrees and out-degrees of all its nodes within the larger \texttt{CauseNet} graph. The intuition is that nodes with lower total degrees represent more niche concepts.
    \item \textbf{\texttt{semantics}:} This heuristic, as detailed in the main text, selects the candidate sub-graph that minimises a weighted score over three metrics. For a given candidate subgraph with concept (node) set $\mathcal{S}_c$, the metrics are:
    \begin{itemize}
        \item \textbf{Semantic Compactness ($H_{compact}$):} Measures thematic coherence by calculating the average cosine distance of each concept's vector embedding from their geometric centroid:
        $$
        H_{compact}(\mathcal{S}_c) = \operatorname{avg}_{v \in \mathcal{S}_c} \left( d_{\cos}(\operatorname{emb}(v), \boldsymbol{\mu}_{\mathcal{S}_c}) \right)
        $$
        where $\operatorname{emb}(v)$ is the vector embedding of concept $v$, $\boldsymbol{\mu}_{\mathcal{S}_c}$ is the centroid vector of all concept embeddings in the set $\mathcal{S}_c$, and $d_{\cos}$ is the cosine distance.

        \item \textbf{Node Specificity ($H_{spec}$):} Penalises overly general ``hub'' nodes using the average log-degree of each node in the full \texttt{CauseNet} graph:
        $$
        H_{spec}(\mathcal{S}_c) = \operatorname{avg}_{v \in \mathcal{S}_c} \left( \log(\operatorname{deg}(v) + 1) \right)
        $$
        where $\operatorname{deg}(v)$ is the degree (sum of in- and out-degrees) of node $v$ in the full \texttt{CauseNet} graph. The logarithm dampens the penalty for extremely high-degree nodes.

        \item \textbf{Structural-Semantic Correlation ($H_{corr}$):} Ensures alignment between graph topology and semantic relationships. The score to be minimised is based on the Spearman's rank correlation:
        $$
        H_{corr}(\mathcal{S}_c) = 1 - \operatorname{SpearmanCorr}(d_{\text{graph}}, d_{\text{sem}})
        $$
        where $d_{\text{graph}}$ is the set of all pairwise shortest-path distances between nodes in the candidate graph, and $d_{\text{sem}}$ is the corresponding set of pairwise cosine distances between their semantic embeddings. A high correlation (low score) indicates that structurally close nodes are also semantically close.
    \end{itemize}
    The final score to be minimised is a weighted sum $H_{final} = w_1 H_{compact} + w_2 H_{spec} + w_3 H_{corr}$, where the weights $w_i$ are customisable. In the generation code, the \texttt{semantics} heuristic defaults to equal weights $(w_1,w_2,w_3)=(1,1,1)$.
\end{itemize}

Additionally, our implementation supports user-defined heuristics, allowing researchers to specify custom scoring functions to rank candidate sub-graphs according to their own criteria or domain-specific preferences.
\vspace{-0.2cm}
\subsection{CauseNet Datasets Generation Schema}\label{sec:data_gen_schema}
\begin{figure*}[ht]
    \centering
    \includegraphics[width=\textwidth]{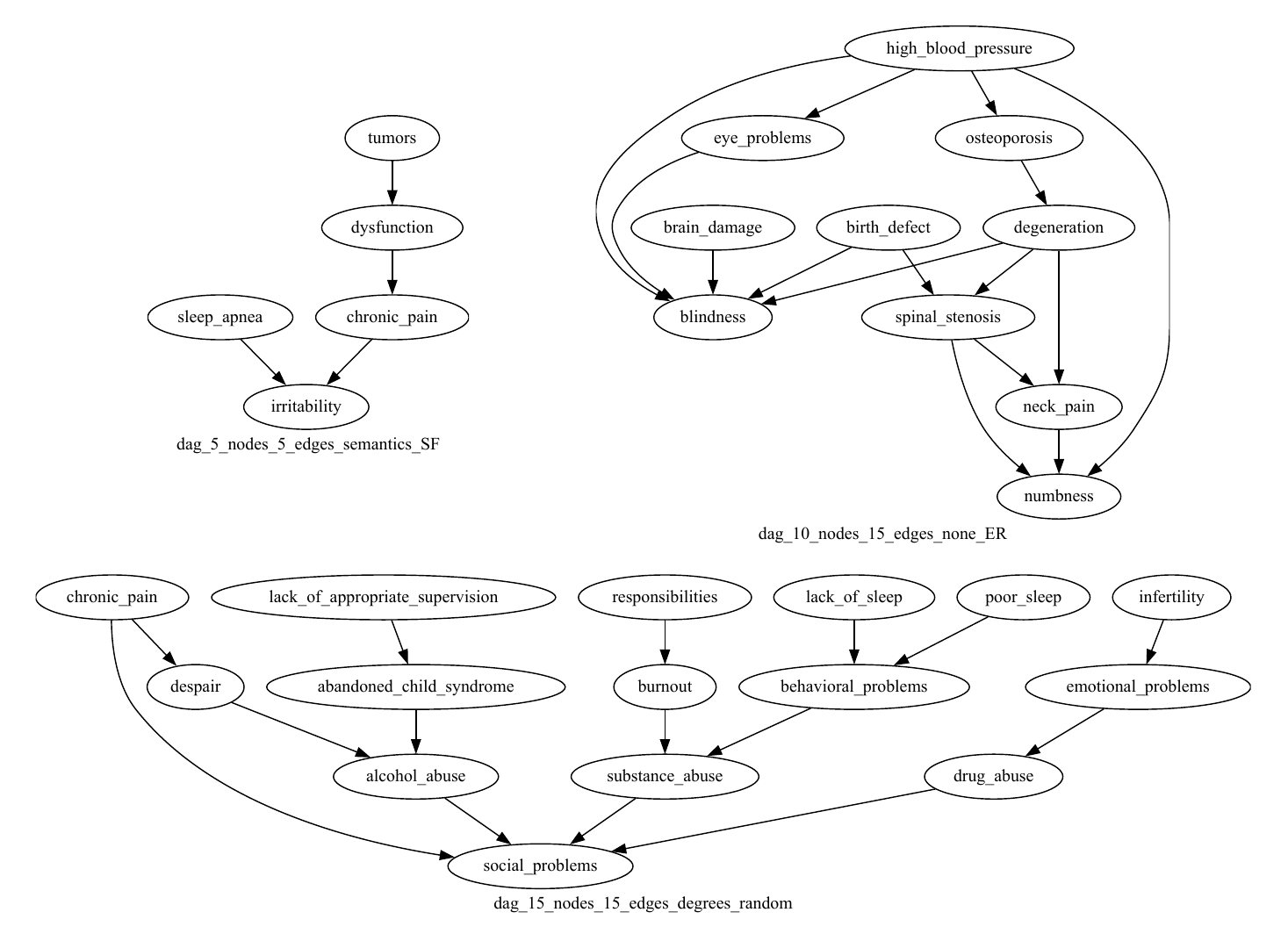}
    \caption{Examples of semantically grounded DAGs from our synthetic data pipeline, generated with different structural methods (ER, SF, LT) and semantic heuristics (\texttt{none}, \texttt{degrees}, \texttt{semantics}).}
    \label{fig:synthetic_examples}
\end{figure*}

\vspace{-0.2cm}
The full suite of 54 synthetic datasets was generated by creating one graph for each unique combination of the following parameters:
\begin{itemize}
    \item \textbf{Number of Nodes:} \{5, 10, 15\}
    \item \textbf{Graph Density:} Two levels of edge counts for each node size:
        \begin{itemize}
            \item \textbf{5 nodes:} 5 and 7 edges
            \item \textbf{10 nodes:} 10 and 15 edges
            \item \textbf{15 nodes:} 15 and 22 edges
        \end{itemize}
    \item \textbf{Graph Type:} \{ER, SF, LT\}
    \item \textbf{Heuristic:} \{none, degrees, semantics\}
\end{itemize}
Some examples of the generated graphs are shown in Fig.~\ref{fig:synthetic_examples}.
\vspace{-0.2cm}
\subsubsection{CPT and Data Generation}
\vspace{-0.2cm}
After a semantically grounded DAG is finalised, we use the \href{https://pyagrum.readthedocs.io/en/latest/}{\texttt{pyAgrum} library} to construct the corresponding Bayesian Network. For each variable in the BN, the domain size is set to 2, making all variables binary. The Conditional Probability Tables (CPTs) for the BN are randomly generated during the initialisation process.
The full collection of the 54 generated BNs in BIFXML and PNG format is available in the project's \href{https://github.com/briziorusso/ArgCausalDisco}{public repository}, allowing for full reproducibility of our experiments.

The randomly initialised CPTs are not intended to reflect realistic effect sizes or context-specific independencies. Their sole purpose is to generate observational data that respects the sampled graph structure.
Our evaluation targets structural recovery (presence and orientation of edges), not the calibration of causal effect magnitudes.
Mismatches between semantic strength and statistical effect size are therefore expected and mirror low-signal regimes encountered in practice.
\begin{table}[h!t]
    \caption{Details of the \texttt{bnlearn} benchmark networks.}
    \label{tab:real_data_det}    
    \centering
    \begin{tabular}{r|ccc}
Dataset Name & $|\nodeSet|$ &  $|\edgSet|$  & $|\edgSet|/|\nodeSet|$ \\
\hline
\href{https://www.bnlearn.com/bnrepository/discrete-small.html#cancer}{CANCER} & 5 & 4 & 0.8\\
\href{https://www.bnlearn.com/bnrepository/discrete-small.html#earthquake}{EARTHQUAKE} & 5 & 4  & 0.8\\
\href{https://www.bnlearn.com/bnrepository/discrete-small.html#survey}{SURVEY} & 6 & 6 & 1\\
\href{https://www.bnlearn.com/bnrepository/discrete-small.html#asia}{ASIA} & 8 & 8  & 1\\
\href{https://www.bnlearn.com/bnrepository/discrete-small.html#sachs}{SACHS} & 11 & 17 & 1.55\\
\href{https://www.bnlearn.com/bnrepository/discrete-medium.html#child}{CHILD} & 20 & 25 & 1.25\\
    \end{tabular}
\end{table}
\subsection{Bnlearn Datasets Generation}\label{sec:data_gen_bnlearn}
For the complementary benchmark evaluation (see \S\ref{sec:exp_eval}), we use six discrete Bayesian networks from the \texttt{bnlearn} repository~\citep{bnlearn}.\footnote{\url{https://www.bnlearn.com/bnrepository/}} The repository is widely used in causal-discovery and Bayesian-network structure-learning work, and collects standard toy, expert-designed, and experimentally derived networks. 
Specifically, we use \texttt{asia}, \texttt{cancer}, \texttt{earthquake}, \texttt{survey}, \texttt{sachs}, and \texttt{child} (Table~\ref{tab:real_data_det}). This set includes small diagnostic or illustrative networks, the Sachs protein-signalling network~\citep{doi:10.1126/science.1105809}, and the Child diagnosis network~\citep{spiegelhalter1992learning}; the Asia network traces back to the classic expert-system example of \citet{10.1111/j.2517-6161.1988.tb01721.x}. 
For each network, we load the BIF structure and associated conditional probability tables from the repository and sample 5000 observations with 50 different random seeds to measure variance and confidence intervals.
\vspace{-0.2cm}
\section{Details on Experiments}
\label{sec:appendix_experiment_details}
\vspace{-0.2cm}
We provide here additional details on the experimental setup, metrics, and results that were omitted from the main text due to space constraints.
\vspace{-0.2cm}
\subsection{Metrics Definitions}
\label{sec:appendix_metrics_defs}
\vspace{-0.3cm}
We evaluate the performance of causal discovery algorithms using standard metrics that capture different aspects of structural accuracy. Let $G_{true} = (V, E_{true})$ be the ground truth and $G_{est} = (V, E_{est})$ the estimated DAG.
\begin{itemize}
    \item \textbf{Structural Hamming Distance (SHD):} The SHD \citep{tsamardinos2006MMHC} measures the structural difference between two graphs. It is defined as the number of edge operations (additions, deletions, or reversals) required to transform $G_{est}$ into $G_{true}$. A lower SHD indicates a better structural match.
    \item \textbf{Structural Intervention Distance (SID):} The SID \citep{peters2015structural} is a metric that quantifies the difference in the interventional distributions implied by two causal graphs. It counts the number of pairs $(i, j)$ for which the causal effect of intervening on node $i$ on node $j$ is incorrectly predicted by $G_{est}$ compared to $G_{true}$. A lower SID indicates a better match in terms of causal predictions.
    \item \textbf{Precision, Recall, and F1-Score:} These metrics evaluate the accuracy of the learned directed edge set. Let TP (True Positives) be the number of true arrows recovered with the correct orientation, FP (False Positives) be the number of predicted arrows not present in the true directed edge set, and FN (False Negatives) be the number of true arrows not recovered with the correct orientation.
    \begin{itemize}
        \item \textbf{Precision} is the fraction of correctly identified arrows among all arrows in the estimated graph: $\text{Precision} = \frac{\text{TP}}{\text{TP} + \text{FP}}$.
        \item \textbf{Recall} is the fraction of true arrows that were correctly identified: $\text{Recall} = \frac{\text{TP}}{\text{TP} + \text{FN}}$.
        \item \textbf{F1-Score} is the harmonic mean of Precision and Recall: $F1 = 2 \cdot \frac{\text{Precision} \cdot \text{Recall}}{\text{Precision} + \text{Recall}}$.
    \end{itemize}
\end{itemize}
\subsection{Baselines}\label{sec:baseline_det}
We used the following nine baselines with respective implementations (see \S\ref{sec:exp_eval} for context):
\begin{itemize}
    \item A Random baseline (RND) as in \citep{russo24causalaba}, by sampling 10 random ER DAGs with the same number of nodes $d$ as the ground truth. For each draw, the number of edges is sampled as
    $S \sim \mathrm{Unif}\{d,\ldots,d(d-1)/2\}$; conditional on $S=s$, the undirected skeleton is sampled uniformly from all $s$-edge subsets of the $\binom{d}{2}$ possible unordered pairs, and then oriented acyclically by a uniformly random node ordering, i.e. for order $\pi$, $\{i,j\}$ is oriented as $i\to j$ iff $\pi(i)>\pi(j)$.
    \item Fast Greedy Equivalence Search\footnote{\url{https://github.com/bd2kccd/py-causal}} (FGS)~\citep{ramsey2017million} is a score-based causal discovery algorithm. It is a fast implementation of GES \citep{chickering2002optimal}: graphs are evaluated using the Bayesian Information Criterion (BIC) as edges are greedily inserted and removed in forward and backward phases.
    \item NOTEARS-MLP\footnote{\url{https://github.com/xunzheng/notears}} (NT)~\citep{zheng2020notears_mlp} learns a non-linear SEM via continuous optimisation. Having a Multi-Layer Perceptron (MLP) at its core, this method should adapt to different functional dependencies among the variables. The optimisation is carried out via augmented Lagrangian with a continuous formulation of acyclicity \citep{zheng2018dagstearscontinuousoptimization}.
    \item GRaSP\footnote{ \url{https://github.com/py-why/causal-learn/blob/main/causallearn/search/PermutationBased/GRaSP.py}}~\citep{lam2022grasp} is a permutation-based method that searches over variable orderings as a scalable relaxation of the sparsest permutation criterion. For each candidate permutation it constructs a subgraph-minimal DAG by selecting parents only among preceding variables (using a grow--shrink strategy) and scores the projected DAG via BIC; it then greedily explores neighbouring permutations via \emph{tuck} operations (a generalisation of local transpositions) to improve sparsity/score.
    \item BOSS\footnote{ \url{https://github.com/py-why/causal-learn/blob/main/causallearn/search/PermutationBased/BOSS.py}}~\citep{andrews2023boss} is a permutation-based score search that greedily traverses variable orderings, projects each ordering to a subgraph-minimal DAG using grow--shrink, and returns the MEC of the highest-scoring projected DAG (BIC). To scale to larger graphs, it uses grow--shrink trees to cache intermediate computations across permutations, avoiding redundant scoring.
    \item Majority-PC\footnote{\url{https://github.com/briziorusso/ArgCausalDisco/blob/llm-uai2026-freeze/cd_algorithms/PC.py}} (MPC)~\citep{colombo2014order} is a constraint-based causal discovery algorithm. It uses CI tests and graphical rules based on $d$-separation to extract a CPDAG from the data. MPC is an improved version of the original Peter-Clark (PC) algorithm~\citep{spirtes2000causation} that renders it order-independent while maintaining soundness and completeness with infinite data. This constitutes the statistical engine underlying ABAPC, isolating the impact of Causal ABA. 
    \item MPC-LLM\footnote{\url{https://github.com/briziorusso/ArgCausalDisco/blob/llm-uai2026-freeze/cd_algorithms/models.py}} runs MPC with the same consensus required/forbidden arrow constraints supplied through the hard background-knowledge functionality of \texttt{causal-learn}~\citep{zheng2024causal}. This enforces the specified directions and prohibitions in the returned graph, with no subsequent arbitration or relaxation step.
    \item ABAPC\footnote{\url{https://github.com/briziorusso/ArgCausalDisco/blob/llm-uai2026-freeze/abapc.py}} (ABAPC)~\citep{russo24causalaba} is an instantiation of Causal ABA, a logic- and constraint-based causal discovery algorithm that uses an Assumption-Based Argumentation framework to resolve conflicts and ambiguities within observational data. The method uses the CI tests from the MPC algorithm as inputs into the Causal ABA (hence ABAPC) and excludes a progressively larger number of low-confidence tests to find a stable extension of the ABAF that contains a DAG which is guaranteed to imply the $d$-separations supported by the retained tests. This is the non-LLM variant of our proposed method, allowing us to isolate the impact of LLM-derived constraints.
    \item LLM-BFS\footnote{\url{https://github.com/superkaiba/causal-llm-bfs}} (BFS)~\citep{jiralerspong2024efficient} is a breadth-first search-based approach that leverages LLMs for causal discovery. It explores the graph structure by iteratively expanding the search space and incorporating LLM-generated insights to build the causal graph, only leveraging variable names and descriptions. This method is the LLM-only baseline, allowing us to isolate the impact of data-driven constraints.
\end{itemize}

\subsection{Additional Details on Experiments and Results}
\label{sec:appendix_additional_experiments}

The main text reports the most compact \texttt{CauseNet} DAG view. This appendix gives the complementary breakdowns in the order they are referenced in the main text: additional \texttt{CauseNet} structure tables (\S\ref{sec:appendix_additional_metrics}), \texttt{bnlearn} results (\S\ref{sec:appendix_bnlearn}), runtime (\S\ref{sec:appendix_runtime}), LLM-constraint ablations (\S\ref{sec:appendix_constraint_eval}), LLM-source comparisons (\S\ref{sec:appendix_gemini_gpt}), and interaction heatmaps (\S\ref{sec:appendix_interaction_details}). Statistical-test tables are collected at the end because they support multiple preceding results (\S\ref{sec:appendix_statistical_tests}).

\subsubsection{Additional CauseNet Results}
\label{sec:appendix_additional_metrics}
Table~\ref{tab:causenet_dag_v_graph_type} refines the main \texttt{CauseNet} result by crossing node count with graph type. The average columns show that ABAPC-LLM has the best marginal SHD and F1 for $|\nodeSet|=5$ and $|\nodeSet|=10$, and the best marginal SHD for $|\nodeSet|=15$; the only node-count marginal exception is $|\nodeSet|=15$ F1, where MPC-LLM is higher. Averaging also over graph type and size, ABAPC-LLM remains best overall for both SHD and F1. The individual cells make clear where direct hard constraints are most competitive: MPC-LLM is strong for F1 on larger LT graphs and some $|\nodeSet|=15$ settings, matching the mechanism discussed in the main text. It uses LLM-required arrows directly, whereas ABAPC-LLM only orients them after skeleton compatibility and conflict resolution.
\begin{table*}[t]
    \caption{Structural metrics on \texttt{CauseNet} by graph type and $|\nodeSet|$, with marginal averages. Among repeated-run methods, bold marks the best mean and statistically tied methods under BH-corrected Welch tests within each metric/column. Significance levels for the $p$-values against the next non-bold method in the ranking are: \textnormal{***} $p<0.001$, \textnormal{**} $p<0.01$, \textnormal{*} $p<0.05$, \textnormal{.} $p<0.1$. LLM-BFS is shown from one saved run and is excluded from Welch tests and bolding. Tests: Table~\ref{tab:causenet_type_size_indicator_tests}; tied bolding: Table~\ref{tab:structural_tied_bolding_tests}.}
    \label{tab:causenet_dag_v_graph_type}
    \centering
    \scriptsize
    \setlength{\tabcolsep}{2pt}
    \renewcommand{\arraystretch}{0.68}
    \begin{tabular}{@{}cclcccc@{}}
    \toprule
     & & \textbf{Method} & \textbf{ER} & \textbf{SF} & \textbf{LT} & \textbf{Average} \\
    \midrule
    \multirow{20}{*}{\rotatebox{90}{\textbf{$|\nodeSet|=5$}}} & \multirow{10}{*}{\textbf{SHD $\downarrow$}} & Random & \num{6.717 +- 1.517} & \num{7.133 +- 1.453} & \num{6.807 +- 1.465} & \num{6.886 +- 1.478} \\
     &  & FGS & \num{6.570 +- 0.972} & \num{4.580 +- 0.907} & \num{5.903 +- 1.358} & \num{5.684 +- 1.079} \\
     &  & NOTEARS-MLP & \num{6.000 +- 0.000} & \num{3.997 +- 0.023} & \num{6.000 +- 0.000} & \num{5.332 +- 0.008} \\
     &  & GRaSP & \num{3.750 +- 1.553} & \num{2.910 +- 1.038} & \num{4.110 +- 1.620} & \num{3.590 +- 1.404} \\
     &  & BOSS & \num{3.843 +- 1.515} & \num{2.740 +- 1.008} & \num{4.003 +- 1.392} & \num{3.529 +- 1.305} \\
     &  & MPC & \num{2.727 +- 0.658} & \num{3.130 +- 0.468} & \num{4.027 +- 0.802} & \num{3.294 +- 0.643} \\
     &  & MPC-LLM & \num{1.680 +- 0.305} & \num{1.603 +- 0.305} & {\bfseries \num{1.217 +- 0.345}}\,\textnormal{***} & \num{1.500 +- 0.318} \\
     &  & ABAPC & \num{1.515 +- 0.542} & \num{1.850 +- 0.383} & \num{2.709 +- 0.722} & \num{2.025 +- 0.549} \\
     &  & ABAPC-LLM & {\bfseries \num{1.054 +- 0.372}}\,\textnormal{***} & {\bfseries \num{1.302 +- 0.278}}\,\textnormal{**} & {\bfseries \num{1.335 +- 0.457}}\,\textnormal{***} & {\bfseries \num{1.230 +- 0.369}}\,\textnormal{***} \\
     &  & LLM-BFS & \num{2.667} & \num{2.000} & \num{3.333} & \num{2.667} \\
    \cmidrule(lr){2-7}
     & \multirow{10}{*}{\textbf{F1 $\uparrow$}} & Random & \num{0.337 +- 0.178} & \num{0.252 +- 0.163} & \num{0.332 +- 0.170} & \num{0.307 +- 0.171} \\
     &  & FGS & \num{0.175 +- 0.140} & \num{0.197 +- 0.157} & \num{0.257 +- 0.157} & \num{0.209 +- 0.151} \\
     &  & NOTEARS-MLP & \num{0.000 +- 0.000} & \num{0.002 +- 0.010} & \num{0.000 +- 0.000} & \num{0.001 +- 0.003} \\
     &  & GRaSP & \num{0.612 +- 0.167} & \num{0.577 +- 0.042} & \num{0.603 +- 0.167} & \num{0.597 +- 0.125} \\
     &  & BOSS & \num{0.605 +- 0.152} & \num{0.593 +- 0.042} & \num{0.620 +- 0.157} & \num{0.606 +- 0.117} \\
     &  & MPC & \num{0.692 +- 0.063} & \num{0.615 +- 0.007} & \num{0.500 +- 0.110} & \num{0.602 +- 0.060} \\
     &  & MPC-LLM & \num{0.805 +- 0.025} & {\bfseries \num{0.697 +- 0.055}}\,\textnormal{***} & {\bfseries \num{0.883 +- 0.030}}\,\textnormal{**} & \num{0.795 +- 0.037} \\
     &  & ABAPC & \num{0.766 +- 0.078} & \num{0.632 +- 0.075} & \num{0.617 +- 0.098} & \num{0.671 +- 0.084} \\
     &  & ABAPC-LLM & {\bfseries \num{0.843 +- 0.044}}\,\textnormal{***} & {\bfseries \num{0.734 +- 0.046}}\,\textnormal{***} & \num{0.858 +- 0.057} & {\bfseries \num{0.812 +- 0.049}}\,\textnormal{*} \\
     &  & LLM-BFS & \num{0.653} & \num{0.630} & \num{0.601} & \num{0.628} \\
    \midrule
    \multirow{20}{*}{\rotatebox{90}{\textbf{$|\nodeSet|=10$}}} & \multirow{10}{*}{\textbf{SHD $\downarrow$}} & Random & \num{27.053 +- 5.917} & \num{28.720 +- 6.325} & \num{28.967 +- 6.600} & \num{28.247 +- 6.281} \\
     &  & FGS & \num{16.350 +- 1.835} & \num{15.923 +- 1.612} & \num{16.813 +- 1.795} & \num{16.362 +- 1.747} \\
     &  & NOTEARS-MLP & \num{12.477 +- 0.137} & \num{12.963 +- 0.147} & \num{12.487 +- 0.082} & \num{12.642 +- 0.122} \\
     &  & GRaSP & \num{7.913 +- 2.340} & \num{9.513 +- 2.852} & \num{7.017 +- 3.065} & \num{8.148 +- 2.752} \\
     &  & BOSS & \num{7.143 +- 2.267} & \num{8.530 +- 2.622} & \num{6.500 +- 2.993} & \num{7.391 +- 2.627} \\
     &  & MPC & \num{7.188 +- 1.275} & \num{6.883 +- 1.527} & \num{7.093 +- 1.212} & \num{7.055 +- 1.338} \\
     &  & MPC-LLM & \num{5.488 +- 0.772} & {\bfseries \num{6.112 +- 1.198}} & \num{7.017 +- 1.098} & \num{6.206 +- 1.023} \\
     &  & ABAPC & \num{5.720 +- 1.161} & \num{6.221 +- 1.493} & \num{6.557 +- 1.671} & \num{6.166 +- 1.442} \\
     &  & ABAPC-LLM & {\bfseries \num{5.061 +- 0.791}}\,\textnormal{**} & {\bfseries \num{5.711 +- 1.302}}\,\textnormal{*} & {\bfseries \num{6.105 +- 1.437}}\,\textnormal{*} & {\bfseries \num{5.626 +- 1.176}}\,\textnormal{***} \\
     &  & LLM-BFS & \num{9.500} & \num{17.167} & \num{13.167} & \num{13.278} \\
    \cmidrule(lr){2-7}
     & \multirow{10}{*}{\textbf{F1 $\uparrow$}} & Random & \num{0.192 +- 0.098} & \num{0.183 +- 0.095} & \num{0.175 +- 0.078} & \num{0.183 +- 0.091} \\
     &  & FGS & \num{0.127 +- 0.087} & \num{0.122 +- 0.072} & \num{0.120 +- 0.077} & \num{0.123 +- 0.078} \\
     &  & NOTEARS-MLP & \num{0.002 +- 0.022} & \num{0.007 +- 0.028} & \num{0.003 +- 0.013} & \num{0.004 +- 0.021} \\
     &  & GRaSP & \num{0.542 +- 0.162} & \num{0.467 +- 0.160} & \num{0.602 +- 0.205} & \num{0.537 +- 0.176} \\
     &  & BOSS & \num{0.588 +- 0.160} & \num{0.512 +- 0.167} & {\bfseries \num{0.638 +- 0.192}}\,\textnormal{***} & \num{0.579 +- 0.173} \\
     &  & MPC & \num{0.560 +- 0.100} & \num{0.620 +- 0.098} & \num{0.558 +- 0.097} & \num{0.579 +- 0.098} \\
     &  & MPC-LLM & {\bfseries \num{0.663 +- 0.057}}\,\textnormal{***} & \num{0.657 +- 0.092} & \num{0.560 +- 0.083} & \num{0.627 +- 0.077} \\
     &  & ABAPC & \num{0.616 +- 0.094} & \num{0.643 +- 0.101} & \num{0.575 +- 0.135} & \num{0.612 +- 0.110} \\
     &  & ABAPC-LLM & {\bfseries \num{0.656 +- 0.057}}\,\textnormal{***} & {\bfseries \num{0.681 +- 0.086}}\,\textnormal{**} & {\bfseries \num{0.623 +- 0.111}}\,\textnormal{*} & {\bfseries \num{0.653 +- 0.084}}\,\textnormal{***} \\
     &  & LLM-BFS & \num{0.448} & \num{0.383} & \num{0.497} & \num{0.439} \\
    \midrule
    \multirow{20}{*}{\rotatebox{90}{\textbf{$|\nodeSet|=15$}}} & \multirow{10}{*}{\textbf{SHD $\downarrow$}} & Random & \num{62.333 +- 18.323} & \num{64.417 +- 20.652} & \num{64.547 +- 19.117} & \num{63.766 +- 19.364} \\
     &  & FGS & \num{26.243 +- 2.217} & \num{20.207 +- 1.870} & \num{26.460 +- 2.250} & \num{24.303 +- 2.112} \\
     &  & NOTEARS-MLP & \num{18.483 +- 0.090} & \num{13.977 +- 0.265} & \num{18.487 +- 0.185} & \num{16.982 +- 0.180} \\
     &  & GRaSP & \num{11.150 +- 2.942} & \num{9.317 +- 2.747} & \num{10.800 +- 3.353} & \num{10.422 +- 3.014} \\
     &  & BOSS & \num{10.947 +- 3.300} & \num{9.067 +- 2.800} & \num{9.970 +- 2.930} & \num{9.994 +- 3.010} \\
     &  & MPC & \num{9.673 +- 1.293} & \num{8.340 +- 0.955} & \num{8.413 +- 1.322} & \num{8.809 +- 1.190} \\
     &  & MPC-LLM & \num{8.815 +- 1.132} & \num{8.307 +- 0.978} & {\bfseries \num{6.502 +- 1.003}}\,\textnormal{**} & \num{7.874 +- 1.038} \\
     &  & ABAPC & \num{8.843 +- 1.696} & {\bfseries \num{7.103 +- 1.144}}\,\textnormal{***} & \num{7.761 +- 1.770} & \num{7.902 +- 1.537} \\
     &  & ABAPC-LLM & {\bfseries \num{8.259 +- 1.245}}\,\textnormal{*} & {\bfseries \num{7.226 +- 1.151}}\,\textnormal{***} & \num{7.165 +- 1.447} & {\bfseries \num{7.550 +- 1.281}}\,\textnormal{**} \\
     &  & LLM-BFS & \num{18.000} & \num{21.500} & \num{17.000} & \num{18.833} \\
    \cmidrule(lr){2-7}
     & \multirow{10}{*}{\textbf{F1 $\uparrow$}} & Random & \num{0.130 +- 0.055} & \num{0.102 +- 0.050} & \num{0.133 +- 0.055} & \num{0.122 +- 0.053} \\
     &  & FGS & \num{0.085 +- 0.060} & \num{0.042 +- 0.055} & \num{0.095 +- 0.055} & \num{0.074 +- 0.057} \\
     &  & NOTEARS-MLP & \num{0.000 +- 0.008} & \num{0.005 +- 0.022} & \num{0.000 +- 0.013} & \num{0.002 +- 0.014} \\
     &  & GRaSP & \num{0.540 +- 0.150} & \num{0.507 +- 0.168} & \num{0.585 +- 0.157} & \num{0.544 +- 0.158} \\
     &  & BOSS & \num{0.555 +- 0.160} & \num{0.515 +- 0.177} & \num{0.618 +- 0.148} & \num{0.563 +- 0.162} \\
     &  & MPC & \num{0.622 +- 0.075} & \num{0.557 +- 0.073} & \num{0.685 +- 0.067} & \num{0.621 +- 0.072} \\
     &  & MPC-LLM & {\bfseries \num{0.677 +- 0.062}}\,\textnormal{**} & \num{0.548 +- 0.072} & {\bfseries \num{0.775 +- 0.042}}\,\textnormal{***} & {\bfseries \num{0.667 +- 0.058}}\,\textnormal{*} \\
     &  & ABAPC & \num{0.619 +- 0.098} & {\bfseries \num{0.601 +- 0.080}}\,\textnormal{***} & \num{0.677 +- 0.092} & \num{0.632 +- 0.090} \\
     &  & ABAPC-LLM & \num{0.660 +- 0.068} & {\bfseries \num{0.590 +- 0.077}}\,\textnormal{***} & \num{0.712 +- 0.070} & \num{0.654 +- 0.072} \\
     &  & LLM-BFS & \num{0.317} & \num{0.339} & \num{0.411} & \num{0.356} \\
    \midrule
    \multirow{20}{*}{\rotatebox{90}{\textbf{Avg.}}} & \multirow{10}{*}{\textbf{SHD $\downarrow$}} & Random & \num{32.034 +- 8.586} & \num{33.423 +- 9.477} & \num{33.440 +- 9.061} & \num{32.966 +- 9.041} \\
     &  & FGS & \num{16.388 +- 1.674} & \num{13.570 +- 1.463} & \num{16.392 +- 1.801} & \num{15.450 +- 1.646} \\
     &  & NOTEARS-MLP & \num{12.320 +- 0.076} & \num{10.312 +- 0.145} & \num{12.324 +- 0.089} & \num{11.652 +- 0.103} \\
     &  & GRaSP & \num{7.604 +- 2.278} & \num{7.247 +- 2.212} & \num{7.309 +- 2.679} & \num{7.387 +- 2.390} \\
     &  & BOSS & \num{7.311 +- 2.361} & \num{6.779 +- 2.143} & \num{6.824 +- 2.438} & \num{6.971 +- 2.314} \\
     &  & MPC & \num{6.529 +- 1.076} & \num{6.118 +- 0.983} & \num{6.511 +- 1.112} & \num{6.386 +- 1.057} \\
     &  & MPC-LLM & \num{5.328 +- 0.736} & \num{5.341 +- 0.827} & {\bfseries \num{4.912 +- 0.816}}\,\textnormal{***} & \num{5.193 +- 0.793} \\
     &  & ABAPC & \num{5.359 +- 1.133} & \num{5.058 +- 1.007} & \num{5.676 +- 1.387} & \num{5.364 +- 1.176} \\
     &  & ABAPC-LLM & {\bfseries \num{4.791 +- 0.803}}\,\textnormal{**} & {\bfseries \num{4.747 +- 0.910}}\,\textnormal{*} & {\bfseries \num{4.868 +- 1.114}}\,\textnormal{***} & {\bfseries \num{4.802 +- 0.942}}\,\textnormal{***} \\
     &  & LLM-BFS & \num{10.056} & \num{13.556} & \num{11.167} & \num{11.593} \\
    \cmidrule(lr){2-7}
     & \multirow{10}{*}{\textbf{F1 $\uparrow$}} & Random & \num{0.219 +- 0.111} & \num{0.179 +- 0.103} & \num{0.213 +- 0.101} & \num{0.204 +- 0.105} \\
     &  & FGS & \num{0.129 +- 0.096} & \num{0.120 +- 0.094} & \num{0.157 +- 0.096} & \num{0.135 +- 0.095} \\
     &  & NOTEARS-MLP & \num{0.001 +- 0.010} & \num{0.004 +- 0.020} & \num{0.001 +- 0.009} & \num{0.002 +- 0.013} \\
     &  & GRaSP & \num{0.564 +- 0.159} & \num{0.517 +- 0.123} & \num{0.597 +- 0.176} & \num{0.559 +- 0.153} \\
     &  & BOSS & \num{0.583 +- 0.157} & \num{0.540 +- 0.128} & \num{0.626 +- 0.166} & \num{0.583 +- 0.150} \\
     &  & MPC & \num{0.624 +- 0.079} & \num{0.597 +- 0.059} & \num{0.581 +- 0.091} & \num{0.601 +- 0.077} \\
     &  & MPC-LLM & {\bfseries \num{0.715 +- 0.048}}\,\textnormal{***} & \num{0.634 +- 0.073} & {\bfseries \num{0.739 +- 0.052}}\,\textnormal{***} & \num{0.696 +- 0.057} \\
     &  & ABAPC & \num{0.667 +- 0.090} & \num{0.625 +- 0.085} & \num{0.623 +- 0.108} & \num{0.638 +- 0.095} \\
     &  & ABAPC-LLM & {\bfseries \num{0.720 +- 0.056}}\,\textnormal{***} & {\bfseries \num{0.668 +- 0.070}}\,\textnormal{***} & {\bfseries \num{0.731 +- 0.079}}\,\textnormal{***} & {\bfseries \num{0.706 +- 0.069}}\,\textnormal{*} \\
     &  & LLM-BFS & \num{0.473} & \num{0.451} & \num{0.503} & \num{0.475} \\
    \bottomrule
    \end{tabular}
\end{table*}
Table~\ref{tab:alpha_ablation_dag} compares the headline metrics under the main $\alpha=0.01$ CI threshold and the archived $\alpha=0.05$ run. Moving to $\alpha=0.05$ changes the CI-derived skeleton and the downstream compatibility checks faced by semantic constraints. The result is a useful stress test for the integration policy: MPC-LLM becomes the top F1 method at all node counts, while ABAPC-LLM still improves over ABAPC but by a smaller margin, consistent with direct MPC background knowledge extracting more of the usable LLM signal. Specifically, going from $\alpha=0.01$ to $\alpha=0.05$, mean relaxed facts increase from 209 to 270 at $|\nodeSet|=10$ and from 255 to 360 at $|\nodeSet|=15$, while retained-CI F1 falls from 0.519 to 0.477 and from 0.505 to 0.455, respectively. A less conservative significance threshold $\alpha=0.05$ leaves a denser CI-derived skeleton, so more true adjacencies remain available for the final graph. At the same time, the retained CI evidence is less clean: ABAPC-LLM has to drop more CI facts before a stable extension exists, and those retained facts match the ground truth less well. This weakens the benefit of its conservative semantic integration, while MPC-LLM still applies the same LLM constraints directly as hard background knowledge.
\begin{table*}[t]
    \caption{Significance threshold ablation on \texttt{CauseNet}. The main runs use $\alpha=0.01$; the comparison uses $\alpha=0.05$ runs. Among repeated-run methods, bold marks the best mean and statistically tied methods under BH-corrected Welch tests within each metric/column. Significance levels for the $p$-values against the next non-bold method in the ranking are: \textnormal{***} $p<0.001$, \textnormal{**} $p<0.01$, \textnormal{*} $p<0.05$, \textnormal{.} $p<0.1$. LLM-BFS is shown from one saved run and is excluded from Welch tests and bolding. Tests: Table~\ref{tab:alpha_indicator_tests}; tied bolding: Table~\ref{tab:structural_tied_bolding_tests}.}
    \label{tab:alpha_ablation_dag}
    \centering
    \scriptsize
    \setlength{\tabcolsep}{2pt}
    \resizebox{0.8\textwidth}{!}{%
    \begin{tabular}{cllccc}
    \toprule
    \textbf{$\alpha$} & \textbf{Metric} & \textbf{Method} & \textbf{$|\nodeSet|=5$} & \textbf{$|\nodeSet|=10$} & \textbf{$|\nodeSet|=15$} \\
    \midrule
    \multirow{20}{*}{\rotatebox{90}{\textbf{$\alpha=0.01$}}} & \multirow{10}{*}{\textbf{SHD $\downarrow$}} & Random & \num{6.886 +- 1.478} & \num{28.247 +- 6.281} & \num{63.766 +- 19.364} \\
     &  & FGS & \num{5.684 +- 1.079} & \num{16.362 +- 1.747} & \num{24.303 +- 2.112} \\
     &  & NOTEARS-MLP & \num{5.332 +- 0.008} & \num{12.642 +- 0.122} & \num{16.982 +- 0.180} \\
     &  & GRaSP & \num{3.590 +- 1.404} & \num{8.148 +- 2.752} & \num{10.422 +- 3.014} \\
     &  & BOSS & \num{3.529 +- 1.305} & \num{7.391 +- 2.627} & \num{9.994 +- 3.010} \\
     &  & MPC & \num{3.294 +- 0.643} & \num{7.055 +- 1.338} & \num{8.809 +- 1.190} \\
     &  & MPC-LLM & \num{1.500 +- 0.318} & \num{6.206 +- 1.023} & \num{7.874 +- 1.038} \\
     &  & ABAPC & \num{2.025 +- 0.549} & \num{6.166 +- 1.442} & \num{7.902 +- 1.537} \\
     &  & ABAPC-LLM & {\bfseries \num{1.230 +- 0.369}}\,\textnormal{***} & {\bfseries \num{5.626 +- 1.176}}\,\textnormal{***} & {\bfseries \num{7.550 +- 1.281}}\,\textnormal{**} \\
     &  & LLM-BFS & \num{2.667} & \num{13.278} & \num{18.833} \\
    \cmidrule(lr){2-6}
     & \multirow{10}{*}{\textbf{F1 $\uparrow$}} & Random & \num{0.307 +- 0.171} & \num{0.183 +- 0.091} & \num{0.122 +- 0.053} \\
     &  & FGS & \num{0.209 +- 0.151} & \num{0.123 +- 0.078} & \num{0.074 +- 0.057} \\
     &  & NOTEARS-MLP & \num{0.001 +- 0.003} & \num{0.004 +- 0.021} & \num{0.002 +- 0.014} \\
     &  & GRaSP & \num{0.597 +- 0.125} & \num{0.537 +- 0.176} & \num{0.544 +- 0.158} \\
     &  & BOSS & \num{0.606 +- 0.117} & \num{0.579 +- 0.173} & \num{0.563 +- 0.162} \\
     &  & MPC & \num{0.602 +- 0.060} & \num{0.579 +- 0.098} & \num{0.621 +- 0.072} \\
     &  & MPC-LLM & \num{0.795 +- 0.037} & \num{0.627 +- 0.077} & {\bfseries \num{0.667 +- 0.058}}\,\textnormal{*} \\
     &  & ABAPC & \num{0.671 +- 0.084} & \num{0.612 +- 0.110} & \num{0.632 +- 0.090} \\
     &  & ABAPC-LLM & {\bfseries \num{0.812 +- 0.049}}\,\textnormal{*} & {\bfseries \num{0.653 +- 0.084}}\,\textnormal{***} & \num{0.654 +- 0.072} \\
     &  & LLM-BFS & \num{0.628} & \num{0.439} & \num{0.356} \\
    \midrule
    \multirow{20}{*}{\rotatebox{90}{\textbf{$\alpha=0.05$}}} & \multirow{10}{*}{\textbf{SHD $\downarrow$}} & Random & \num{6.886 +- 1.478} & \num{28.247 +- 6.281} & \num{63.766 +- 19.364} \\
     &  & FGS & \num{5.684 +- 1.079} & \num{16.362 +- 1.747} & \num{24.303 +- 2.112} \\
     &  & NOTEARS-MLP & \num{5.332 +- 0.008} & \num{12.642 +- 0.122} & \num{16.982 +- 0.180} \\
     &  & GRaSP & \num{3.590 +- 1.404} & \num{8.148 +- 2.752} & \num{10.422 +- 3.014} \\
     &  & BOSS & \num{3.529 +- 1.305} & \num{7.391 +- 2.627} & \num{9.994 +- 3.010} \\
     &  & MPC & \num{2.913 +- 0.868} & \num{5.267 +- 1.791} & \num{7.289 +- 1.971} \\
     &  & MPC-LLM & {\bfseries \num{1.170 +- 0.583}}\,\textnormal{***} & {\bfseries \num{4.531 +- 1.516}}\,\textnormal{***} & {\bfseries \num{6.439 +- 1.769}}\,\textnormal{***} \\
     &  & ABAPC & \num{1.894 +- 0.695} & \num{5.833 +- 1.583} & \num{7.893 +- 1.882} \\
     &  & ABAPC-LLM & {\bfseries \num{1.233 +- 0.514}}\,\textnormal{***} & \num{5.414 +- 1.381} & \num{7.504 +- 1.670} \\
     &  & LLM-BFS & \num{2.667} & \num{13.278} & \num{18.833} \\
    \cmidrule(lr){2-6}
     & \multirow{10}{*}{\textbf{F1 $\uparrow$}} & Random & \num{0.307 +- 0.171} & \num{0.183 +- 0.091} & \num{0.122 +- 0.053} \\
     &  & FGS & \num{0.209 +- 0.151} & \num{0.123 +- 0.078} & \num{0.074 +- 0.057} \\
     &  & NOTEARS-MLP & \num{0.001 +- 0.003} & \num{0.004 +- 0.021} & \num{0.002 +- 0.014} \\
     &  & GRaSP & \num{0.597 +- 0.125} & \num{0.537 +- 0.176} & \num{0.544 +- 0.158} \\
     &  & BOSS & \num{0.606 +- 0.117} & \num{0.579 +- 0.173} & \num{0.563 +- 0.162} \\
     &  & MPC & \num{0.653 +- 0.110} & \num{0.691 +- 0.131} & \num{0.684 +- 0.106} \\
     &  & MPC-LLM & {\bfseries \num{0.850 +- 0.073}}\,\textnormal{***} & {\bfseries \num{0.734 +- 0.103}}\,\textnormal{***} & {\bfseries \num{0.721 +- 0.092}}\,\textnormal{***} \\
     &  & ABAPC & \num{0.695 +- 0.099} & \num{0.636 +- 0.111} & \num{0.640 +- 0.103} \\
     &  & ABAPC-LLM & \num{0.817 +- 0.064} & \num{0.672 +- 0.093} & \num{0.663 +- 0.088} \\
     &  & LLM-BFS & \num{0.628} & \num{0.439} & \num{0.356} \\
    \bottomrule
    \end{tabular}
    }
\end{table*}
\clearpage

\subsubsection{Performance on \texttt{bnlearn} Benchmarks}
\label{sec:appendix_bnlearn}
Table~\ref{tab:bnlearn_dag_all_metrics} reports the full structural metric table for the standard \texttt{bnlearn} benchmarks, with $|\nodeSet|$ and $|\edgSet|$ included in the dataset labels. These benchmarks are useful for comparability with the causal-discovery literature, but they are public and likely present in LLM pre-training data; the results should therefore be interpreted as a complementary pseudo-real benchmark rather than as the main evidence for generalisation.

The pattern is nevertheless informative. ABAPC-LLM is strongest or statistically tied on the larger and more difficult \texttt{sachs} and \texttt{child} networks, and remains tied with the best group on the easy \texttt{earthquake} network where several methods nearly recover the graph. MPC-LLM is best on \texttt{asia} and \texttt{survey}, and tied with ABAPC-LLM on \texttt{cancer}; this mirrors the \texttt{CauseNet} story that direct LLM background knowledge can help when the semantic constraints are highly reliable. The benefit of the argumentative layer is clearest where purely hard constraints would otherwise over-commit or where many statistical and semantic facts must be reconciled.

The LLM-only baseline achieves very high recall on several of these public networks, but at lower precision; this is useful as a diagnostic, not as a deployable strategy. It suggests that LLM-derived constraints can contain real signal, although maybe confounded by memorisation, while also reinforcing the need for a data-driven and conflict-aware layer before accepting a graph.
\begin{table*}[t]
    \caption{Structural metrics on \texttt{bnlearn} benchmarks; dataset labels report $|\nodeSet|$ and $|\edgSet|$. Among repeated-run methods, bold marks the best mean and statistically tied methods under BH-corrected Welch tests within each metric/column. Significance levels for the $p$-values against the next non-bold method in the ranking are: \textnormal{***} $p<0.001$, \textnormal{**} $p<0.01$, \textnormal{*} $p<0.05$, \textnormal{.} $p<0.1$. LLM-BFS is shown from one saved run and is excluded from Welch tests and bolding. Tests: Tables~\ref{tab:bnlearn_earthquake_indicator_tests}, \ref{tab:bnlearn_other_indicator_tests}, and \ref{tab:bnlearn_child_indicator_tests}; tied bolding: Table~\ref{tab:bnlearn_tied_bolding_tests}.}
    \label{tab:bnlearn_dag_all_metrics}
    \centering
    \scriptsize
    \setlength{\tabcolsep}{2pt}
    \resizebox{\textwidth}{!}{%
    \begin{tabular}{llccccc}
    \toprule
    \textbf{Dataset} & \textbf{Method} & \textbf{SHD $\downarrow$} & \textbf{SID $\downarrow$} & \textbf{Precision $\uparrow$} & \textbf{Recall $\uparrow$} & \textbf{F1 $\uparrow$} \\
    \midrule
    \multirow{10}{*}{\shortstack[l]{\texttt{cancer}\\$|\nodeSet|=5$\\$|\edgSet|=4$}} & Random & \num{5.600 +- 1.650} & \num{11.260 +- 4.400} & \num{0.200 +- 0.220} & \num{0.200 +- 0.220} & \num{0.360 +- 0.160} \\
     & FGS & \num{2.320 +- 0.470} & \num{11.280 +- 1.880} & \num{0.450 +- 0.080} & \num{0.420 +- 0.120} & \num{0.430 +- 0.100} \\
     & NOTEARS-MLP & \num{3.980 +- 0.140} & \num{9.980 +- 0.620} & \num{0.250 +- 0.500} & \num{0.000 +- 0.040} & \num{0.400} \\
     & GRaSP & \num{3.800 +- 0.760} & \num{9.740 +- 1.880} & \num{0.580 +- 0.370} & \num{0.050 +- 0.190} & \num{0.660 +- 0.280} \\
     & BOSS & \num{3.600 +- 1.110} & \num{9.160 +- 2.620} & \num{0.860 +- 0.240} & \num{0.100 +- 0.280} & \num{0.770 +- 0.300} \\
     & MPC & \num{3.420 +- 1.400} & \num{8.520 +- 3.520} & \num{0.940 +- 0.180} & \num{0.150 +- 0.350} & {\bfseries \num{0.940 +- 0.180}}\,\textnormal{**} \\
     & MPC-LLM & {\bfseries \num{0.680 +- 0.590}}\,\textnormal{***} & {\bfseries \num{0.780 +- 1.000}}\,\textnormal{***} & {\bfseries \num{0.990 +- 0.050}}\,\textnormal{.} & {\bfseries \num{0.840 +- 0.130}}\,\textnormal{***} & {\bfseries \num{0.900 +- 0.080}}\,\textnormal{**} \\
     & ABAPC & \num{2.106 +- 1.148} & \num{8.070 +- 4.491} & \num{0.556 +- 0.272} & \num{0.478 +- 0.276} & \num{0.604 +- 0.179} \\
     & ABAPC-LLM & {\bfseries \num{0.680 +- 0.587}}\,\textnormal{***} & {\bfseries \num{0.960 +- 1.324}}\,\textnormal{***} & {\bfseries \num{0.996 +- 0.028}}\,\textnormal{.} & {\bfseries \num{0.835 +- 0.148}}\,\textnormal{***} & {\bfseries \num{0.901 +- 0.091}}\,\textnormal{**} \\
     & LLM-BFS & \num{4.000} & \num{0.000} & \num{0.500} & \num{1.000} & \num{0.667} \\
    \midrule
    \multirow{10}{*}{\shortstack[l]{\texttt{earthquake}\\$|\nodeSet|=5$\\$|\edgSet|=4$}} & Random & \num{5.700 +- 1.420} & \num{10.560 +- 3.550} & \num{0.170 +- 0.160} & \num{0.170 +- 0.160} & \num{0.290 +- 0.100} \\
     & FGS & \num{2.620 +- 0.640} & \num{9.940 +- 0.240} & \num{0.440 +- 0.060} & \num{0.500 +- 0.000} & \num{0.470 +- 0.030} \\
     & NOTEARS-MLP & \num{2.120 +- 1.170} & \num{9.760 +- 5.380} & \num{0.500 +- 0.270} & \num{0.500 +- 0.280} & \num{0.500 +- 0.270} \\
     & GRaSP & {\bfseries \num{0.160 +- 0.790}}\,\textnormal{***} & {\bfseries \num{0.400 +- 1.980}} & {\bfseries \num{1.000 +- 0.000}}\,\textnormal{***} & {\bfseries \num{0.960 +- 0.200}} & {\bfseries \num{1.000 +- 0.000}}\,\textnormal{***} \\
     & BOSS & {\bfseries \num{0.400 +- 1.210}}\,\textnormal{***} & \num{1.000 +- 3.030} & {\bfseries \num{1.000 +- 0.000}}\,\textnormal{***} & \num{0.900 +- 0.300} & {\bfseries \num{1.000 +- 0.000}}\,\textnormal{***} \\
     & MPC & {\bfseries \num{0.040 +- 0.200}}\,\textnormal{***} & {\bfseries \num{0.000 +- 0.000}}\,\textnormal{*} & {\bfseries \num{0.990 +- 0.040}}\,\textnormal{***} & {\bfseries \num{1.000 +- 0.000}}\,\textnormal{*} & {\bfseries \num{1.000 +- 0.020}}\,\textnormal{***} \\
     & MPC-LLM & {\bfseries \num{0.040 +- 0.200}}\,\textnormal{***} & {\bfseries \num{0.000 +- 0.000}}\,\textnormal{*} & {\bfseries \num{0.990 +- 0.040}}\,\textnormal{***} & {\bfseries \num{1.000 +- 0.000}}\,\textnormal{*} & {\bfseries \num{1.000 +- 0.020}}\,\textnormal{***} \\
     & ABAPC & {\bfseries \num{0.100 +- 0.463}}\,\textnormal{***} & {\bfseries \num{0.160 +- 1.131}}\,\textnormal{.} & {\bfseries \num{0.982 +- 0.080}}\,\textnormal{***} & {\bfseries \num{0.990 +- 0.071}}\,\textnormal{*} & {\bfseries \num{0.986 +- 0.073}}\,\textnormal{***} \\
     & ABAPC-LLM & {\bfseries \num{0.080 +- 0.340}}\,\textnormal{***} & {\bfseries \num{0.080 +- 0.566}}\,\textnormal{.} & {\bfseries \num{0.987 +- 0.052}}\,\textnormal{***} & {\bfseries \num{0.995 +- 0.035}}\,\textnormal{*} & {\bfseries \num{0.991 +- 0.041}}\,\textnormal{***} \\
     & LLM-BFS & \num{4.000} & \num{0.000} & \num{0.500} & \num{1.000} & \num{0.667} \\
    \midrule
    \multirow{10}{*}{\shortstack[l]{\texttt{survey}\\$|\nodeSet|=6$\\$|\edgSet|=6$}} & Random & \num{7.980 +- 1.770} & \num{17.540 +- 6.060} & \num{0.240 +- 0.160} & \num{0.240 +- 0.160} & \num{0.280 +- 0.140} \\
     & FGS & \num{6.680 +- 1.000} & \num{14.260 +- 3.680} & \num{0.400 +- 0.160} & \num{0.210 +- 0.080} & \num{0.270 +- 0.100} \\
     & NOTEARS-MLP & \num{6.000 +- 0.000} & \num{17.000 +- 0.000} & -- & \num{0.000 +- 0.000} & -- \\
     & GRaSP & \num{5.920 +- 0.400} & \num{16.920 +- 0.400} & {\bfseries \num{1.000 +- 0.000}}\,\textnormal{***} & \num{0.010 +- 0.070} & \num{0.500 +- 0.000} \\
     & BOSS & \num{5.880 +- 0.480} & \num{16.880 +- 0.480} & {\bfseries \num{1.000 +- 0.000}}\,\textnormal{***} & \num{0.020 +- 0.080} & \num{0.500 +- 0.000} \\
     & MPC & \num{4.620 +- 1.880} & \num{14.120 +- 4.040} & \num{0.830 +- 0.140} & \num{0.230 +- 0.310} & \num{0.680 +- 0.140} \\
     & MPC-LLM & {\bfseries \num{2.060 +- 1.040}}\,\textnormal{***} & {\bfseries \num{6.320 +- 4.180}}\,\textnormal{***} & \num{0.890 +- 0.120} & {\bfseries \num{0.670 +- 0.160}}\,\textnormal{***} & {\bfseries \num{0.750 +- 0.140}}\,\textnormal{*} \\
     & ABAPC & \num{3.675 +- 1.027} & \num{15.860 +- 4.087} & \num{0.518 +- 0.153} & \num{0.394 +- 0.165} & \num{0.453 +- 0.145} \\
     & ABAPC-LLM & \num{3.297 +- 0.981} & \num{13.477 +- 3.868} & \num{0.612 +- 0.165} & \num{0.457 +- 0.155} & \num{0.517 +- 0.147} \\
     & LLM-BFS & \num{6.000} & \num{0.000} & \num{0.500} & \num{1.000} & \num{0.667} \\
    \midrule
    \multirow{10}{*}{\shortstack[l]{\texttt{asia}\\$|\nodeSet|=8$\\$|\edgSet|=8$}} & Random & \num{12.660 +- 1.810} & \num{30.360 +- 7.940} & \num{0.140 +- 0.110} & \num{0.140 +- 0.110} & \num{0.190 +- 0.080} \\
     & FGS & \num{11.240 +- 0.770} & \num{38.160 +- 2.780} & \num{0.250 +- 0.050} & \num{0.240 +- 0.030} & \num{0.250 +- 0.040} \\
     & NOTEARS-MLP & \num{7.180 +- 1.260} & \num{32.000 +- 8.000} & \num{0.310 +- 0.200} & \num{0.230 +- 0.160} & \num{0.300 +- 0.150} \\
     & GRaSP & \num{6.800 +- 1.500} & \num{17.080 +- 6.220} & \num{0.670 +- 0.250} & \num{0.360 +- 0.130} & \num{0.450 +- 0.110} \\
     & BOSS & \num{6.220 +- 1.000} & \num{24.740 +- 6.740} & \num{0.600 +- 0.350} & \num{0.240 +- 0.140} & \num{0.400 +- 0.110} \\
     & MPC & \num{6.000 +- 0.000} & \num{23.040 +- 0.830} & {\bfseries \num{0.980 +- 0.080}}\,\textnormal{***} & \num{0.250 +- 0.020} & \num{0.400 +- 0.020} \\
     & MPC-LLM & {\bfseries \num{2.840 +- 0.470}}\,\textnormal{***} & {\bfseries \num{13.720 +- 0.540}}\,\textnormal{***} & {\bfseries \num{0.990 +- 0.030}}\,\textnormal{***} & {\bfseries \num{0.650 +- 0.050}}\,\textnormal{***} & {\bfseries \num{0.780 +- 0.040}}\,\textnormal{***} \\
     & ABAPC & \num{5.047 +- 1.028} & \num{25.730 +- 6.848} & \num{0.684 +- 0.207} & \num{0.372 +- 0.122} & \num{0.477 +- 0.149} \\
     & ABAPC-LLM & \num{3.580 +- 0.710} & \num{15.500 +- 2.308} & {\bfseries \num{0.992 +- 0.040}}\,\textnormal{***} & \num{0.557 +- 0.082} & \num{0.710 +- 0.068} \\
     & LLM-BFS & \num{5.000} & \num{0.000} & \num{0.615} & \num{1.000} & \num{0.762} \\
    \midrule
    \multirow{10}{*}{\shortstack[l]{\texttt{sachs}\\$|\nodeSet|=11$\\$|\edgSet|=17$}} & Random & \num{26.020 +- 2.490} & \num{50.400 +- 5.870} & \num{0.160 +- 0.080} & \num{0.160 +- 0.080} & \num{0.170 +- 0.080} \\
     & FGS & \num{23.820 +- 1.640} & \num{48.700 +- 3.850} & \num{0.140 +- 0.040} & \num{0.130 +- 0.040} & \num{0.140 +- 0.030} \\
     & NOTEARS-MLP & \num{13.580 +- 1.200} & \num{52.360 +- 6.610} & \num{0.600 +- 0.190} & \num{0.200 +- 0.070} & \num{0.300 +- 0.100} \\
     & GRaSP & \num{16.420 +- 1.110} & \num{52.320 +- 1.870} & \num{0.550 +- 0.130} & \num{0.030 +- 0.070} & \num{0.220 +- 0.050} \\
     & BOSS & \num{17.000 +- 0.000} & \num{53.000 +- 0.000} & -- & \num{0.000 +- 0.000} & -- \\
     & MPC & \num{16.000 +- 2.740} & \num{50.560 +- 6.680} & \num{0.730 +- 0.010} & \num{0.060 +- 0.160} & \num{0.590 +- 0.030} \\
     & MPC-LLM & \num{13.280 +- 1.970} & \num{47.840 +- 5.910} & \num{0.740 +- 0.020} & \num{0.220 +- 0.120} & \num{0.320 +- 0.100} \\
     & ABAPC & \num{5.244 +- 1.684} & \num{26.212 +- 9.472} & \num{0.721 +- 0.101} & \num{0.693 +- 0.100} & \num{0.706 +- 0.100} \\
     & ABAPC-LLM & {\bfseries \num{4.460 +- 1.264}}\,\textnormal{**} & {\bfseries \num{21.056 +- 6.983}}\,\textnormal{**} & {\bfseries \num{0.759 +- 0.063}}\,\textnormal{*} & {\bfseries \num{0.739 +- 0.075}}\,\textnormal{**} & {\bfseries \num{0.748 +- 0.069}}\,\textnormal{*} \\
     & LLM-BFS & \num{20.000} & \num{50.000} & \num{0.312} & \num{0.294} & \num{0.303} \\
    \midrule
    \multirow{10}{*}{\shortstack[l]{\texttt{child}\\$|\nodeSet|=20$\\$|\edgSet|=25$}} & Random & \num{45.260 +- 2.350} & \num{304.4 +- 22.620} & \num{0.070 +- 0.040} & \num{0.070 +- 0.040} & \num{0.080 +- 0.040} \\
     & FGS & \num{57.000 +- 4.110} & \num{271.7 +- 28.450} & \num{0.110 +- 0.040} & \num{0.190 +- 0.070} & \num{0.140 +- 0.050} \\
     & NOTEARS-MLP & \num{20.660 +- 2.400} & \num{220.3 +- 29.260} & \num{0.450 +- 0.100} & \num{0.370 +- 0.080} & \num{0.400 +- 0.080} \\
     & GRaSP & \num{16.320 +- 3.390} & \num{232.4 +- 36.750} & \num{0.880 +- 0.160} & \num{0.370 +- 0.130} & \num{0.510 +- 0.140} \\
     & BOSS & \num{14.220 +- 2.360} & \num{210.5 +- 28.260} & {\bfseries \num{0.960 +- 0.080}}\,\textnormal{***} & \num{0.440 +- 0.090} & \num{0.600 +- 0.100} \\
     & MPC & \num{9.160 +- 4.130} & \num{164.3 +- 63.590} & \num{0.840 +- 0.070} & \num{0.630 +- 0.170} & \num{0.710 +- 0.120} \\
     & MPC-LLM & {\bfseries \num{5.060 +- 1.800}}\,\textnormal{***} & \num{87.820 +- 26.340} & \num{0.880 +- 0.050} & {\bfseries \num{0.800 +- 0.070}}\,\textnormal{***} & \num{0.830 +- 0.050} \\
     & ABAPC & {\bfseries \num{5.372 +- 3.250}}\,\textnormal{***} & {\bfseries \num{84.833 +- 45.226}} & \num{0.851 +- 0.089} & {\bfseries \num{0.786 +- 0.129}}\,\textnormal{***} & \num{0.815 +- 0.105} \\
     & ABAPC-LLM & {\bfseries \num{4.417 +- 2.009}}\,\textnormal{***} & {\bfseries \num{72.653 +- 24.836}}\,\textnormal{**} & \num{0.894 +- 0.050} & {\bfseries \num{0.824 +- 0.080}}\,\textnormal{***} & {\bfseries \num{0.855 +- 0.052}}\,\textnormal{*} \\
     & LLM-BFS & \num{56.000} & \num{114.0} & \num{0.242} & \num{0.600} & \num{0.345} \\
    \bottomrule
    \end{tabular}
    }
\end{table*}
\clearpage
\subsubsection{Runtime}
\label{sec:appendix_runtime}
Table~\ref{tab:runtime_scaling_final} reports wall-clock runtimes excluding LLM API latency, which is incurred once per dataset when the LLM-derived constraints are generated. The relevant comparison is therefore the solver/search cost after those constraints are available. ABAPC and ABAPC-LLM are slower than MPC because they ground and solve an argumentative program, but the skeleton-reduction step keeps the evaluated networks tractable. The LLM constraints can also reduce the effective search space, so ABAPC-LLM is not uniformly slower than ABAPC despite adding semantic facts.
\begin{table*}[t]
    \caption{Runtime in seconds on \texttt{CauseNet} synthetic datasets (mean$\pm$std; LLM API latency excluded).}
    \label{tab:runtime_scaling_final}
    \centering
    \normalsize
    \setlength{\tabcolsep}{8pt}
    \begin{tabular}{lccc}
    \toprule
    \textbf{Method} & \textbf{$|\nodeSet|=5$} & \textbf{$|\nodeSet|=10$} & \textbf{$|\nodeSet|=15$} \\
    \midrule
    Random & \num{0.00 +- 0.00} & \num{0.00 +- 0.00} & \num{0.00 +- 0.00} \\
    FGS & \num{0.11 +- 0.03} & \num{0.14 +- 0.04} & \num{0.15 +- 0.04} \\
    NOTEARS-MLP & \num{0.19 +- 0.10} & \num{0.64 +- 0.25} & \num{1.20 +- 0.77} \\
    GRaSP & \num{1.10 +- 0.27} & \num{5.45 +- 2.88} & \num{14.06 +- 5.69} \\
    BOSS & \num{1.29 +- 0.32} & \num{7.22 +- 3.41} & \num{18.63 +- 7.64} \\
    MPC & \num{0.01 +- 0.00} & \num{0.04 +- 0.03} & \num{0.05 +- 0.03} \\
    MPC-LLM & \num{0.01 +- 0.00} & \num{0.04 +- 0.03} & \num{0.05 +- 0.03} \\
    ABAPC & \num{0.05 +- 0.03} & \num{1.96 +- 2.88} & \num{4.42 +- 4.06} \\
    ABAPC-LLM & \num{0.06 +- 0.05} & \num{1.81 +- 2.54} & \num{4.33 +- 3.97} \\
    \bottomrule
    \end{tabular}
\end{table*}

\subsection{LLM Constraints Evaluation}
\label{sec:appendix_constraint_eval}
In this section we evaluate the LLM-derived required and forbidden arrows before they are integrated with CI tests, compare single-run averages with the five-run unanimity consensus used in the main experiments, and test whether variable descriptions help beyond variable names. These analyses explain when the semantic layer helps structural learning and when it can become a source of conflict.

Table~\ref{tab:desc_consensus_ablation} summarises LLM-derived constraint quality for average single runs and consensus filters, with and without variable descriptions. The description effect depends strongly on the metadata regime. On \texttt{bnlearn}, descriptions are valuable because some variable names are cryptic (e.g., Survey $A,S,E,O,R,T$ and Sachs protein abbreviations); both forbidden and required constraint F1 improve in the average-run setting. On \texttt{CauseNet}, variable names are already meaningful, so descriptions add less and can sometimes introduce misleading context. The consensus filter is deliberately conservative: it keeps fewer constraints and often lowers recall, but raises reliability when a constraint survives unanimity. This is the intended operating point for ABAPC-LLM, because false semantic facts are costly once they enter the conflict-resolution.
Table~\ref{tab:desc_consensus_nonzero_ablation} repeats the same analysis after removing side-specific zero-constraint cases. This separates the quality of emitted constraints from consensus coverage. When unanimity emits a non-empty constraint set, precision is often very high, especially for forbidden constraints.
\begin{table*}[h!b]
    \caption{LLM-derived constraint quality by dataset, consensus strategy, and description use. Bold compares Average vs. Consensus within each setting; green marks the row best; arrows show the description effect.}
    \label{tab:desc_consensus_ablation}
    \centering
    \scriptsize
    \setlength{\tabcolsep}{3.5pt}
    \resizebox{1\textwidth}{!}{%
    \begin{tabular}{lcccccccc}
    \toprule
     & \multicolumn{4}{c}{\textbf{bnlearn}} & \multicolumn{4}{c}{\textbf{CauseNet}} \\
    \cmidrule(lr){2-5} \cmidrule(lr){6-9}
    \textbf{Metric} & \multicolumn{2}{c}{\textbf{Average}} & \multicolumn{2}{c}{\textbf{Consensus}} & \multicolumn{2}{c}{\textbf{Average}} & \multicolumn{2}{c}{\textbf{Consensus}} \\
    \cmidrule(lr){2-3} \cmidrule(lr){4-5} \cmidrule(lr){6-7} \cmidrule(lr){8-9}
     & \textbf{w/o desc} & \textbf{with desc} & \textbf{w/o desc} & \textbf{with desc} & \textbf{w/o desc} & \textbf{with desc} & \textbf{w/o desc} & \textbf{with desc} \\
    \midrule
    \multicolumn{9}{c}{\textbf{Forbidden constraints}} \\
    \midrule
    \# constraints & \num{19.80 +- 13.90} & \num{25.47 +- 24.39} & \num{8.00 +- 6.60} & \num{8.17 +- 1.72} & \num{17.59 +- 13.02} & \num{20.64 +- 16.60} & \num{4.85 +- 4.83} & \num{4.07 +- 3.42} \\
    Precision & {\bfseries \num{0.927 +- 0.186}} & \best{\num{0.973 +- 0.041}}\,\up & \num{0.833 +- 0.408} & \num{0.948 +- 0.085}\,\up & {\bfseries \num{0.952 +- 0.132}} & {\bfseries \num{0.942 +- 0.155}}\,\down & \num{0.855 +- 0.339} & \num{0.877 +- 0.317}\,\up \\
    Recall & {\bfseries \num{0.409 +- 0.234}} & \best{\num{0.472 +- 0.219}}\,\up & \num{0.241 +- 0.240} & \num{0.275 +- 0.216}\,\up & {\bfseries \num{0.293 +- 0.216}} & {\bfseries \num{0.312 +- 0.221}}\,\up & \num{0.123 +- 0.164} & \num{0.115 +- 0.168}\,\down \\
    F1 & {\bfseries \num{0.533 +- 0.260}} & \best{\num{0.606 +- 0.224}}\,\up & \num{0.338 +- 0.316} & \num{0.392 +- 0.281}\,\up & {\bfseries \num{0.409 +- 0.232}} & {\bfseries \num{0.431 +- 0.234}}\,\up & \num{0.189 +- 0.216} & \num{0.174 +- 0.218}\,\down \\
    \midrule
    \multicolumn{9}{c}{\textbf{Required constraints}} \\
    \midrule
    \# constraints & \num{8.27 +- 6.97} & \num{10.80 +- 11.71} & \num{4.50 +- 3.73} & \num{3.67 +- 4.13} & \num{12.57 +- 9.60} & \num{14.56 +- 11.52} & \num{4.11 +- 3.29} & \num{4.74 +- 3.33} \\
    Precision & {\bfseries \num{0.468 +- 0.288}} & \best{\num{0.597 +- 0.235}}\,\up & \num{0.375 +- 0.327} & \num{0.505 +- 0.422}\,\up & \num{0.487 +- 0.249} & \num{0.464 +- 0.230}\,\down & {\bfseries \num{0.568 +- 0.353}} & {\bfseries \num{0.514 +- 0.335}}\,\down \\
    Recall & {\bfseries \num{0.445 +- 0.352}} & \best{\num{0.611 +- 0.328}}\,\up & \num{0.303 +- 0.388} & \num{0.281 +- 0.375}\,\down & {\bfseries \num{0.486 +- 0.255}} & {\bfseries \num{0.516 +- 0.242}}\,\up & \num{0.244 +- 0.210} & \num{0.278 +- 0.244}\,\up \\
    F1 & {\bfseries \num{0.401 +- 0.240}} & \best{\num{0.555 +- 0.232}}\,\up & \num{0.284 +- 0.284} & \num{0.321 +- 0.337}\,\up & {\bfseries \num{0.450 +- 0.206}} & {\bfseries \num{0.453 +- 0.190}}\,\up & \num{0.314 +- 0.228} & \num{0.334 +- 0.249}\,\up \\
    \bottomrule
    \end{tabular}
    }
\end{table*}
\begin{table*}[t]
    \caption{LLM-derived constraint quality excluding zero-constraint cases. Formatting follows Table~\ref{tab:desc_consensus_ablation}: bold compares Average vs. Consensus within each setting; green marks the row best; arrows show the description effect.}
    \label{tab:desc_consensus_nonzero_ablation}
    \centering
    \scriptsize
    \setlength{\tabcolsep}{3.5pt}
    \resizebox{1\textwidth}{!}{%
    \begin{tabular}{lcccccccc}
    \toprule
     & \multicolumn{4}{c}{\textbf{bnlearn}} & \multicolumn{4}{c}{\textbf{CauseNet}} \\
    \cmidrule(lr){2-5} \cmidrule(lr){6-9}
    \textbf{Metric} & \multicolumn{2}{c}{\textbf{Average}} & \multicolumn{2}{c}{\textbf{Consensus}} & \multicolumn{2}{c}{\textbf{Average}} & \multicolumn{2}{c}{\textbf{Consensus}} \\
    \cmidrule(lr){2-3} \cmidrule(lr){4-5} \cmidrule(lr){6-7} \cmidrule(lr){8-9}
     & \textbf{w/o desc} & \textbf{with desc} & \textbf{w/o desc} & \textbf{with desc} & \textbf{w/o desc} & \textbf{with desc} & \textbf{w/o desc} & \textbf{with desc} \\
    \midrule
    \multicolumn{9}{c}{\textbf{Forbidden constraints}} \\
    \midrule
    \# constraints & \num{20.48 +- 13.63} & \num{25.47 +- 24.39} & \num{9.60 +- 5.94} & \num{8.17 +- 1.72} & \num{17.86 +- 12.94} & \num{20.79 +- 16.57} & \num{5.57 +- 4.78} & \num{4.58 +- 3.29} \\
    Precision & \num{0.959 +- 0.062} & {\bfseries \num{0.973 +- 0.041}}\,\up & \best{\num{1.000 +- 0.000}} & \num{0.948 +- 0.085}\,\down & \num{0.966 +- 0.061} & \num{0.949 +- 0.132}\,\down & {\bfseries \num{0.982 +- 0.069}} & {\bfseries \num{0.986 +- 0.053}}\,\up \\
    Recall & {\bfseries \num{0.423 +- 0.225}} & \best{\num{0.472 +- 0.219}}\,\up & \num{0.289 +- 0.234} & \num{0.275 +- 0.216}\,\down & {\bfseries \num{0.297 +- 0.215}} & {\bfseries \num{0.314 +- 0.220}}\,\up & \num{0.141 +- 0.168} & \num{0.129 +- 0.172}\,\down \\
    F1 & {\bfseries \num{0.551 +- 0.244}} & \best{\num{0.606 +- 0.224}}\,\up & \num{0.405 +- 0.300} & \num{0.392 +- 0.281}\,\down & {\bfseries \num{0.415 +- 0.228}} & {\bfseries \num{0.434 +- 0.231}}\,\up & \num{0.217 +- 0.218} & \num{0.196 +- 0.222}\,\down \\
    \midrule
    \multicolumn{9}{c}{\textbf{Required constraints}} \\
    \midrule
    \# constraints & \num{9.92 +- 6.45} & \num{11.17 +- 11.74} & \num{6.75 +- 1.71} & \num{5.50 +- 3.87} & \num{12.66 +- 9.57} & \num{14.66 +- 11.50} & \num{4.53 +- 3.17} & \num{5.02 +- 3.22} \\
    Precision & \num{0.562 +- 0.213} & \num{0.618 +- 0.210}\,\up & {\bfseries \num{0.562 +- 0.195}} & \best{\num{0.757 +- 0.206}}\,\up & \num{0.491 +- 0.247} & \num{0.468 +- 0.227}\,\down & {\bfseries \num{0.626 +- 0.317}} & {\bfseries \num{0.544 +- 0.320}}\,\down \\
    Recall & {\bfseries \num{0.535 +- 0.316}} & \best{\num{0.632 +- 0.312}}\,\up & \num{0.454 +- 0.399} & \num{0.422 +- 0.394}\,\down & {\bfseries \num{0.490 +- 0.252}} & {\bfseries \num{0.520 +- 0.239}}\,\up & \num{0.269 +- 0.204} & \num{0.294 +- 0.241}\,\up \\
    F1 & {\bfseries \num{0.482 +- 0.172}} & \best{\num{0.574 +- 0.211}}\,\up & \num{0.426 +- 0.233} & \num{0.481 +- 0.294}\,\up & {\bfseries \num{0.454 +- 0.203}} & {\bfseries \num{0.457 +- 0.187}}\,\up & \num{0.346 +- 0.214} & \num{0.354 +- 0.242}\,\up \\
    \bottomrule
    \end{tabular}
    }
\end{table*}
\subsubsection{Gemini and GPT-5-mini Constraint Sources}
\label{sec:appendix_gemini_gpt}
We report \texttt{Gemini-2.5-flash} and \texttt{GPT-5-mini} under the same prompting, parsing, and five-run consensus protocol. \texttt{Gemini-2.5-flash} is used for the broader ablation figures because most constraint-quality and metadata ablations were run with \texttt{Gemini-2.5-flash} and because it gives broader non-empty consensus coverage on the synthetic datasets: 476 unanimous constraints and non-empty consensus sets on 53/54 datasets, compared with 129 unanimous constraints and non-empty consensus sets on 43/54 datasets for \texttt{GPT-5-mini}.
\begin{table*}[t]
    \caption{Structural metrics on the 43 matched \texttt{CauseNet} datasets with Gemini and GPT-5-mini constraints, grouped by $|\nodeSet|$. LLM-constraint variants are labelled by source. Among repeated-run methods, bold marks the best mean and statistically tied methods under BH-corrected Welch tests within each metric/column. Significance levels for the $p$-values against the next non-bold method in the ranking are: \textnormal{***} $p<0.001$, \textnormal{**} $p<0.01$, \textnormal{*} $p<0.05$, \textnormal{.} $p<0.1$. LLM-BFS is shown from one saved run and is excluded from Welch tests and bolding. Tests: Table~\ref{tab:matched_llm_indicator_tests}; tied bolding: Table~\ref{tab:structural_tied_bolding_tests}.}
    \label{tab:gpt_model_ablation_dag}
    \centering
    \scriptsize
    \setlength{\tabcolsep}{1.8pt}
    \resizebox{0.92\textwidth}{!}{%
    \begin{tabular}{l|lcccc}
    \toprule
    & \textbf{Method} & \textbf{SHD $\downarrow$} & \textbf{Precision $\uparrow$} & \textbf{Recall $\uparrow$} & \textbf{F1 $\uparrow$} \\
    \midrule
    \multirow{12}{*}{\rotatebox{90}{\textbf{$|\nodeSet|=5$}}} & Random & \num{6.857 +- 1.470} & \num{0.273 +- 0.145} & \num{0.385 +- 0.225} & \num{0.311 +- 0.168} \\
     & FGS & \num{5.610 +- 1.101} & \num{0.295 +- 0.199} & \num{0.192 +- 0.130} & \num{0.224 +- 0.149} \\
     & NOTEARS-MLP & \num{5.467 +- 0.000} & \num{0.000 +- 0.000} & \num{0.000 +- 0.000} & \num{0.000 +- 0.000} \\
     & GRaSP & \num{3.677 +- 1.465} & \num{0.737 +- 0.266} & \num{0.347 +- 0.242} & \num{0.591 +- 0.137} \\
     & BOSS & \num{3.633 +- 1.367} & \num{0.746 +- 0.232} & \num{0.355 +- 0.229} & \num{0.601 +- 0.127} \\
     & MPC & \num{3.395 +- 0.643} & \num{0.836 +- 0.107} & \num{0.367 +- 0.111} & \num{0.592 +- 0.065} \\
     & MPC-LLM (Gemini) & \num{1.328 +- 0.331} & {\bfseries \num{0.952 +- 0.027}}\,\textnormal{***} & \num{0.753 +- 0.053} & {\bfseries \num{0.822 +- 0.041}}\,\textnormal{***} \\
     & MPC-LLM (GPT) & \num{2.213 +- 0.479} & \num{0.835 +- 0.088} & \num{0.589 +- 0.089} & \num{0.713 +- 0.083} \\
     & ABAPC & \num{2.127 +- 0.566} & \num{0.702 +- 0.084} & \num{0.609 +- 0.090} & \num{0.663 +- 0.083} \\
     & ABAPC-LLM (Gemini) & {\bfseries \num{1.186 +- 0.375}}\,\textnormal{*} & \num{0.902 +- 0.043} & {\bfseries \num{0.778 +- 0.063}}\,\textnormal{*} & {\bfseries \num{0.828 +- 0.048}}\,\textnormal{***} \\
     & ABAPC-LLM (GPT) & \num{1.604 +- 0.546} & \num{0.807 +- 0.090} & \num{0.702 +- 0.095} & \num{0.755 +- 0.084} \\
     & LLM-BFS (Gemini) & \num{2.800} & \num{0.877} & \num{0.558} & \num{0.631} \\
    \midrule
    \multirow{12}{*}{\rotatebox{90}{\textbf{$|\nodeSet|=10$}}} & Random & \num{28.076 +- 6.230} & \num{0.143 +- 0.071} & \num{0.300 +- 0.176} & \num{0.184 +- 0.091} \\
     & FGS & \num{16.127 +- 1.747} & \num{0.189 +- 0.122} & \num{0.107 +- 0.069} & \num{0.132 +- 0.081} \\
     & NOTEARS-MLP & \num{12.687 +- 0.126} & \num{0.029 +- 0.123} & \num{0.002 +- 0.011} & \num{0.005 +- 0.022} \\
     & GRaSP & \num{8.466 +- 2.772} & \num{0.805 +- 0.274} & \num{0.366 +- 0.184} & \num{0.519 +- 0.174} \\
     & BOSS & \num{7.560 +- 2.604} & {\bfseries \num{0.884 +- 0.201}}\,\textnormal{***} & \num{0.426 +- 0.180} & \num{0.569 +- 0.171} \\
     & MPC & \num{7.173 +- 1.314} & \num{0.836 +- 0.109} & \num{0.446 +- 0.098} & \num{0.574 +- 0.096} \\
     & MPC-LLM (Gemini) & \num{6.197 +- 1.095} & \num{0.843 +- 0.093} & \num{0.519 +- 0.081} & \num{0.634 +- 0.084} \\
     & MPC-LLM (GPT) & \num{6.630 +- 1.231} & \num{0.860 +- 0.095} & \num{0.492 +- 0.092} & \num{0.621 +- 0.084} \\
     & ABAPC & \num{6.345 +- 1.449} & \num{0.723 +- 0.126} & \num{0.526 +- 0.102} & \num{0.605 +- 0.107} \\
     & ABAPC-LLM (Gemini) & {\bfseries \num{5.984 +- 1.257}}\,\textnormal{.} & \num{0.776 +- 0.096} & \num{0.542 +- 0.090} & \num{0.633 +- 0.089} \\
     & ABAPC-LLM (GPT) & {\bfseries \num{5.755 +- 1.417}}\,\textnormal{***} & \num{0.787 +- 0.114} & {\bfseries \num{0.569 +- 0.101}}\,\textnormal{***} & {\bfseries \num{0.655 +- 0.102}}\,\textnormal{***} \\
     & LLM-BFS (Gemini) & \num{12.857} & \num{0.489} & \num{0.432} & \num{0.446} \\
    \midrule
    \multirow{12}{*}{\rotatebox{90}{\textbf{$|\nodeSet|=15$}}} & Random & \num{63.989 +- 19.393} & \num{0.083 +- 0.034} & \num{0.302 +- 0.171} & \num{0.123 +- 0.052} \\
     & FGS & \num{24.747 +- 2.130} & \num{0.095 +- 0.074} & \num{0.059 +- 0.046} & \num{0.073 +- 0.056} \\
     & NOTEARS-MLP & \num{17.197 +- 0.177} & \num{0.024 +- 0.121} & \num{0.001 +- 0.008} & \num{0.001 +- 0.014} \\
     & GRaSP & \num{10.230 +- 3.184} & \num{0.854 +- 0.217} & \num{0.436 +- 0.161} & \num{0.564 +- 0.163} \\
     & BOSS & \num{9.653 +- 3.086} & \num{0.892 +- 0.176} & \num{0.460 +- 0.161} & \num{0.589 +- 0.165} \\
     & MPC & \num{8.677 +- 1.207} & {\bfseries \num{0.934 +- 0.068}}\,\textnormal{***} & \num{0.497 +- 0.069} & \num{0.640 +- 0.070} \\
     & MPC-LLM (Gemini) & {\bfseries \num{7.616 +- 1.052}}\,\textnormal{.} & {\bfseries \num{0.934 +- 0.049}}\,\textnormal{***} & \num{0.557 +- 0.060} & {\bfseries \num{0.691 +- 0.055}}\,\textnormal{***} \\
     & MPC-LLM (GPT) & \num{8.059 +- 1.217} & {\bfseries \num{0.941 +- 0.064}}\,\textnormal{***} & \num{0.532 +- 0.070} & \num{0.671 +- 0.068} \\
     & ABAPC & \num{7.863 +- 1.602} & \num{0.770 +- 0.101} & \num{0.554 +- 0.089} & \num{0.641 +- 0.092} \\
     & ABAPC-LLM (Gemini) & {\bfseries \num{7.508 +- 1.292}}\,\textnormal{*} & \num{0.799 +- 0.072} & {\bfseries \num{0.573 +- 0.073}}\,\textnormal{**} & \num{0.663 +- 0.070} \\
     & ABAPC-LLM (GPT) & {\bfseries \num{7.575 +- 1.513}}\,\textnormal{*} & \num{0.790 +- 0.093} & {\bfseries \num{0.570 +- 0.086}}\,\textnormal{*} & \num{0.658 +- 0.087} \\
     & LLM-BFS (Gemini) & \num{16.429} & \num{0.448} & \num{0.341} & \num{0.357} \\
    \bottomrule
    \end{tabular}
    }
\end{table*}
Table~\ref{tab:gpt_model_ablation_dag} reports the matched structural comparison on the 43 synthetic datasets where both \texttt{Gemini-2.5-flash} and \texttt{GPT-5-mini} consensus runs are available. ABAPC-LLM is stable across the two LLM sources, while MPC-LLM remains competitive mainly where direct hard constraints help. \texttt{GPT-5-mini} produces more zero-F1 consensus sets than \texttt{Gemini-2.5-flash} (33.3\% vs.\ 1.9\%) and fewer high-quality sets (3.7\% vs.\ 13.0\%).
\clearpage

\subsubsection{Interaction Between Constraint Quality and Structural Performance}\label{sec:appendix_interaction_details}
The interaction heatmaps ask whether semantic constraints help because they are independently accurate, because the CI evidence is accurate, or because both signals are accurate together. Each cell bins runs by LLM-constraints F1 and retained CI-constraints F1 and reports the mean change in final DAG F1 after adding semantic constraints to ABAPC.

\begin{figure}[ht]
    \centering
    \begin{minipage}{0.5\linewidth}
        \centering
        \includegraphics[width=\linewidth]{figures/interaction_heatmap.png}
        \subcaption{\texttt{CauseNet} synthetic datasets.}
        \label{fig:heapmap_quality_synthetic}
    \end{minipage}
    \hfill
    \begin{minipage}{0.48\linewidth}
        \centering
        \includegraphics[width=\linewidth]{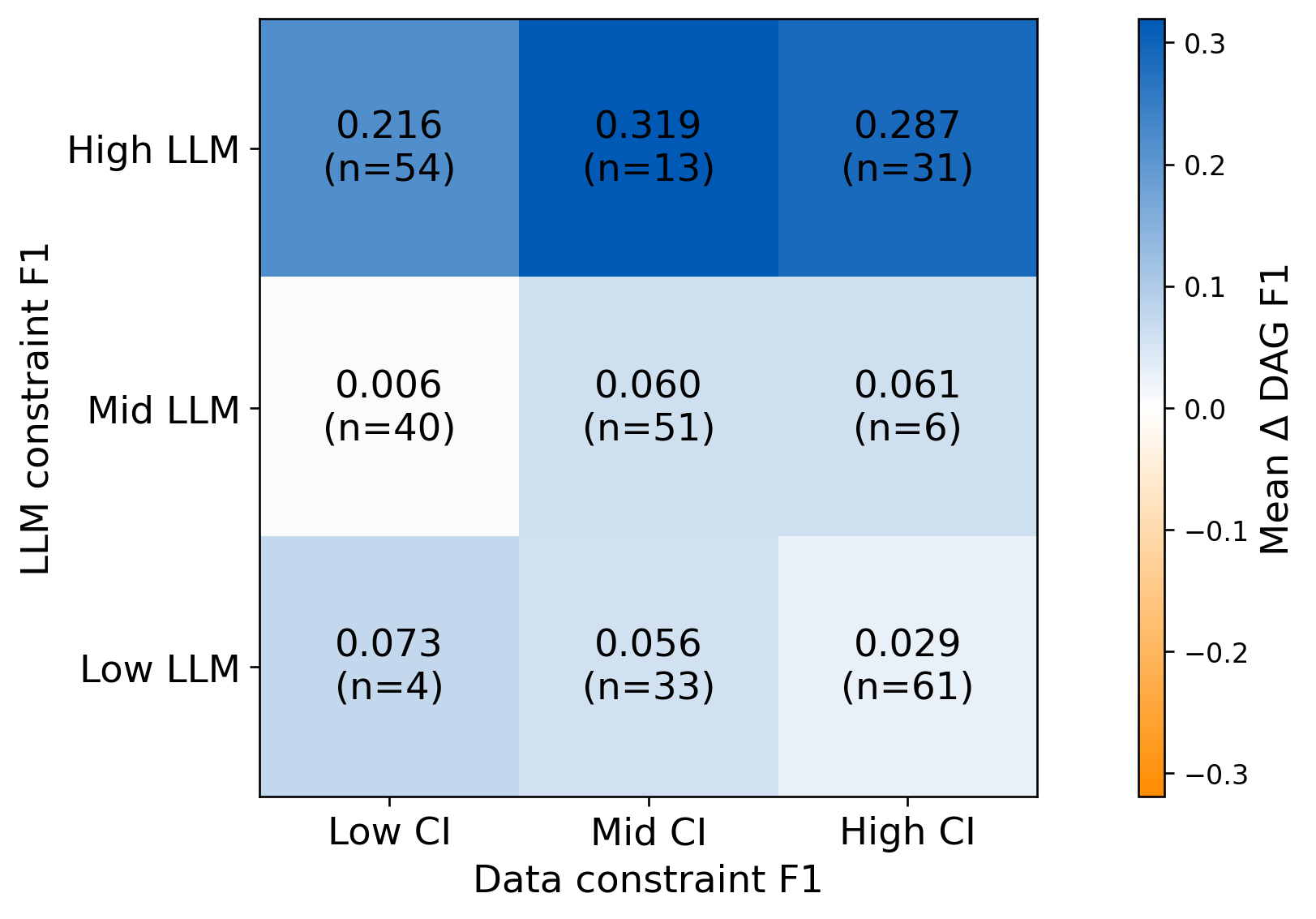}
        \subcaption{\texttt{bnlearn} benchmarks.}
        \label{fig:heapmap_quality_bn}
    \end{minipage}
    \caption{With variable descriptions: interaction between LLM-constraint F1 and retained-CI F1. Cell values are mean $\Delta$ DAG F1 after adding semantic constraints to ABAPC; counts are in parentheses.}
    \label{fig:heapmap_quality_all}
\end{figure}

Figure~\ref{fig:heapmap_quality_all} compares \texttt{CauseNet} and \texttt{bnlearn} with descriptions. On \texttt{CauseNet}, the highest LLM-quality row is positive across all CI bins, while the low-LLM/high-CI cell is negative. This is the central failure mode of the current integration policy: misleading semantic constraints can still force the solver to relax otherwise useful statistical facts. On \texttt{bnlearn}, the cells are non-negative and the high-LLM row is strongly positive, which is consistent with the much cleaner LLM constraints obtained on these public benchmark networks.

\begin{figure}[ht]
    \centering
    \begin{minipage}{0.5\linewidth}
        \centering
        \includegraphics[width=\linewidth]{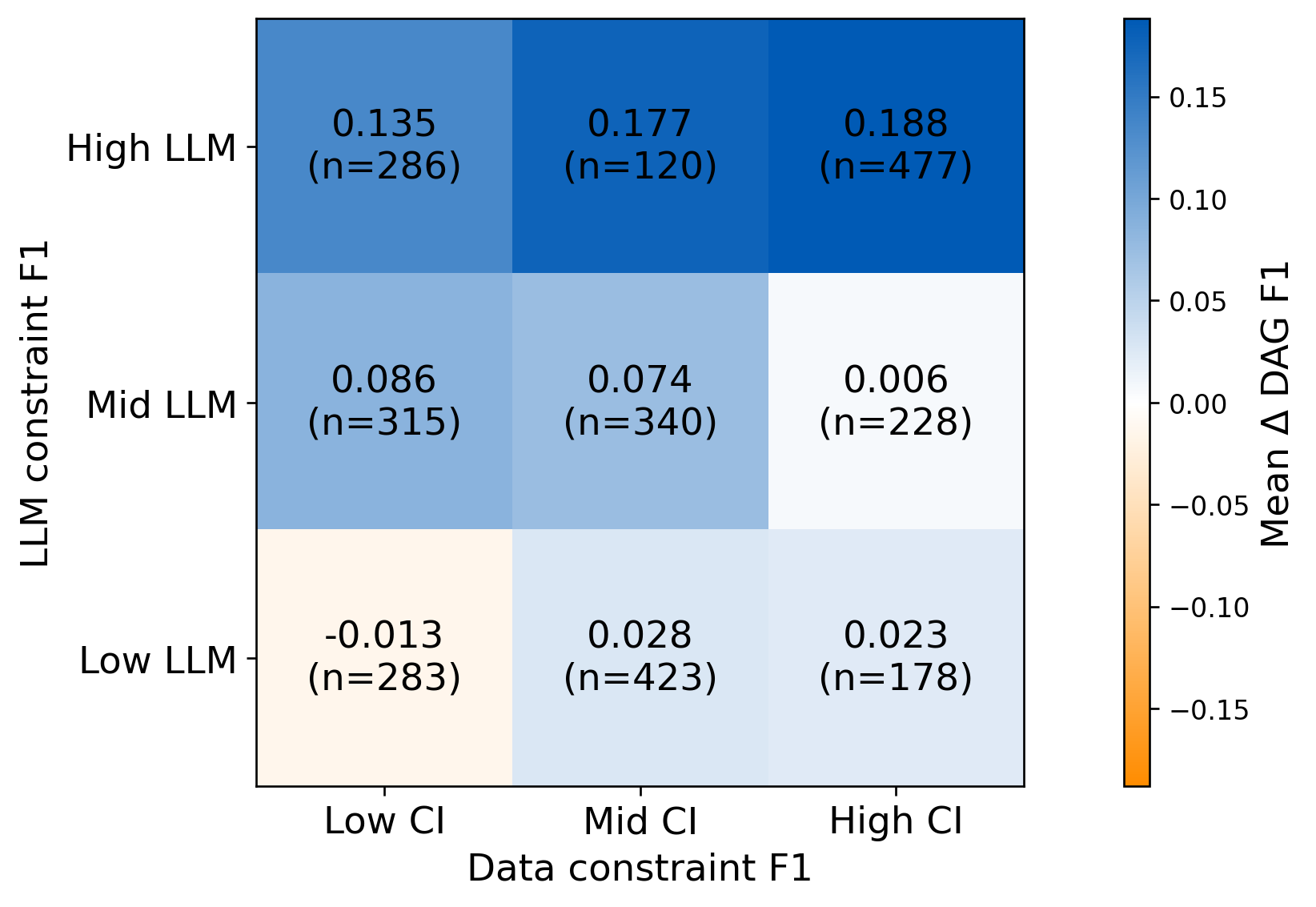}
        \subcaption{\texttt{CauseNet} synthetic datasets (names only).}
        \label{fig:heapmap_quality_synthetic_no_desc}
    \end{minipage}
    \hfill
    \begin{minipage}{0.48\linewidth}
        \centering
        \includegraphics[width=\linewidth]{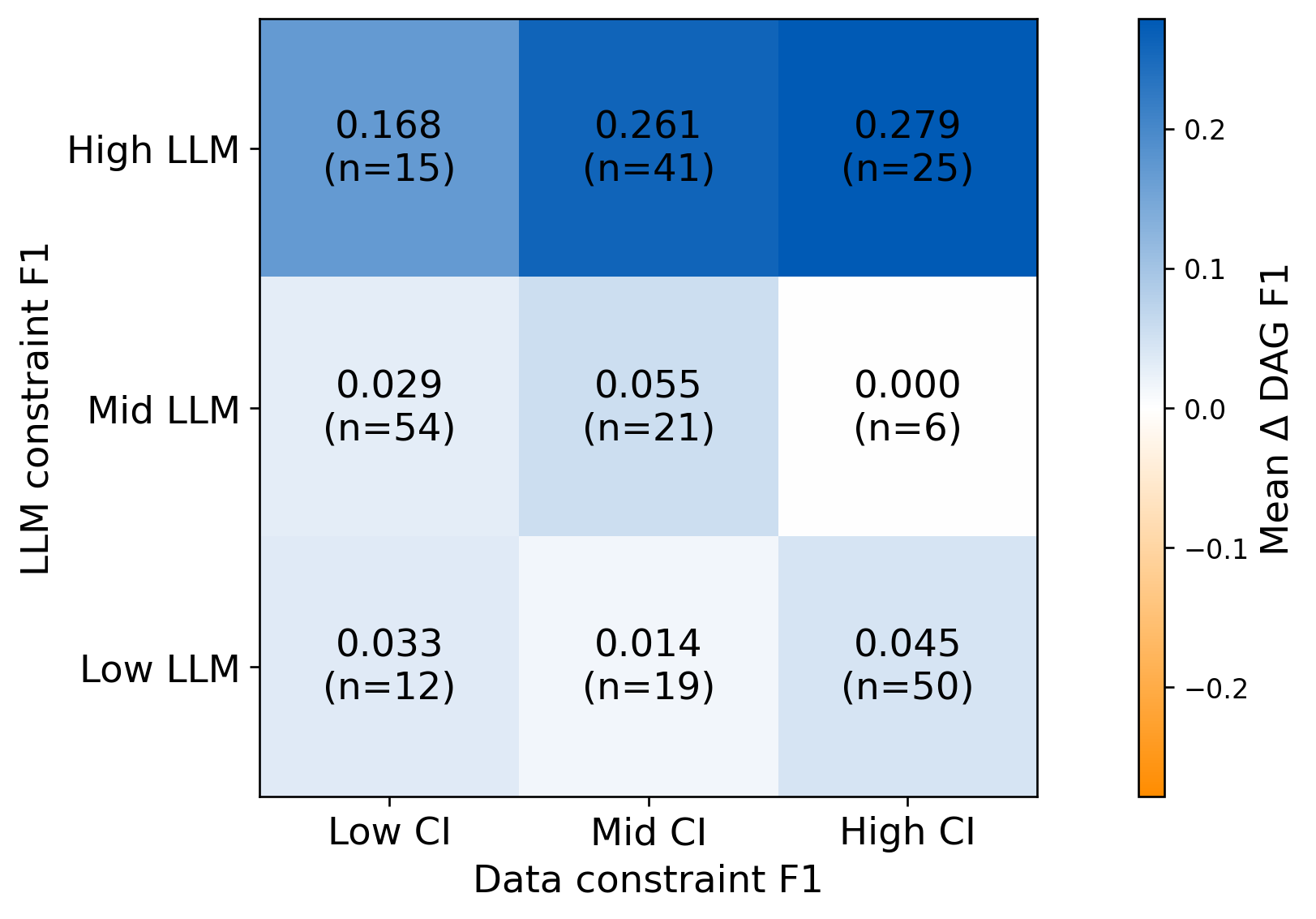}
        \subcaption{\texttt{bnlearn} benchmarks (names only).}
        \label{fig:heapmap_quality_bn_no_desc}
    \end{minipage}
    \caption{Without variable descriptions: interaction between LLM-constraint F1 and retained-CI F1. Names-only metadata weakens the negative \texttt{CauseNet} cells but also removes useful contextual information for some \texttt{bnlearn} labels.}
    \label{fig:heapmap_quality_all_no_desc}
\end{figure}

Figure~\ref{fig:heapmap_quality_all_no_desc} repeats the comparison with names only. The \texttt{CauseNet} heatmap becomes less extreme: gains remain positive in the high-LLM row and the most negative cell is close to zero. This supports the interpretation from Table~\ref{tab:desc_consensus_ablation}: descriptions add useful information in many cases, but they can also add misleading context when variable names already carry strong semantic signal. For \texttt{bnlearn}, names-only LLM-derived constraints remain useful but are less informative for datasets with terse labels, especially \texttt{survey} (single-letter variables $A,S,E,O,R,T$) and \texttt{sachs} (protein-signalling abbreviations such as \texttt{Akt}, \texttt{Mek}, and \texttt{PKC}).

\begin{figure*}[ht]
    \centering
    \includegraphics[width=\linewidth]{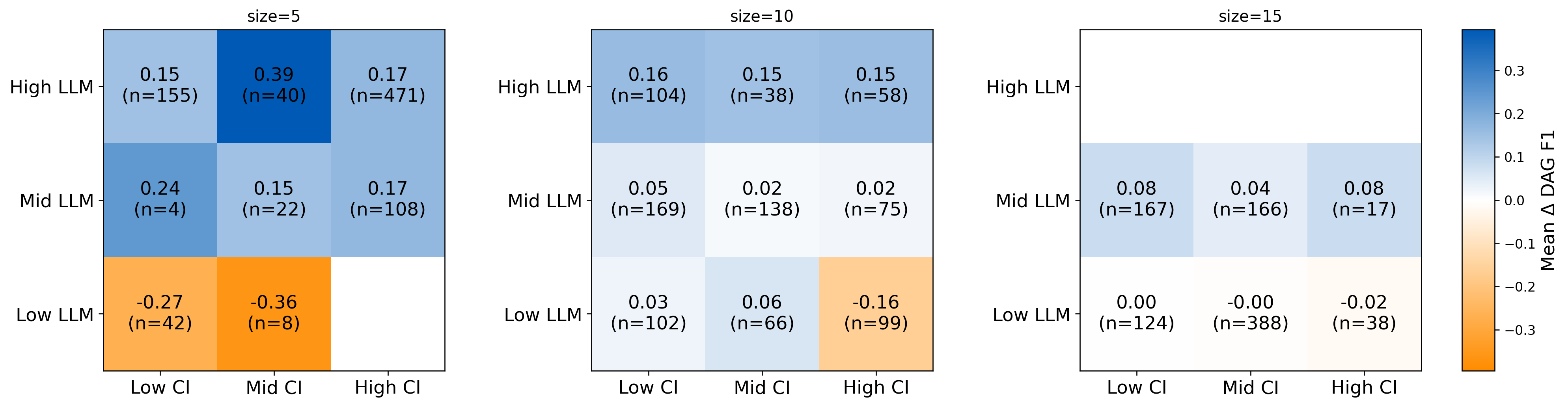}
    \caption{\texttt{CauseNet}: interaction heatmaps split by node count, using the same global bins as Fig.~\ref{fig:constraint_quality}.}
    \label{fig:constraint_quality_per_nodesize}
\end{figure*}

\begin{figure*}[ht]
    \centering
    \includegraphics[width=\linewidth]{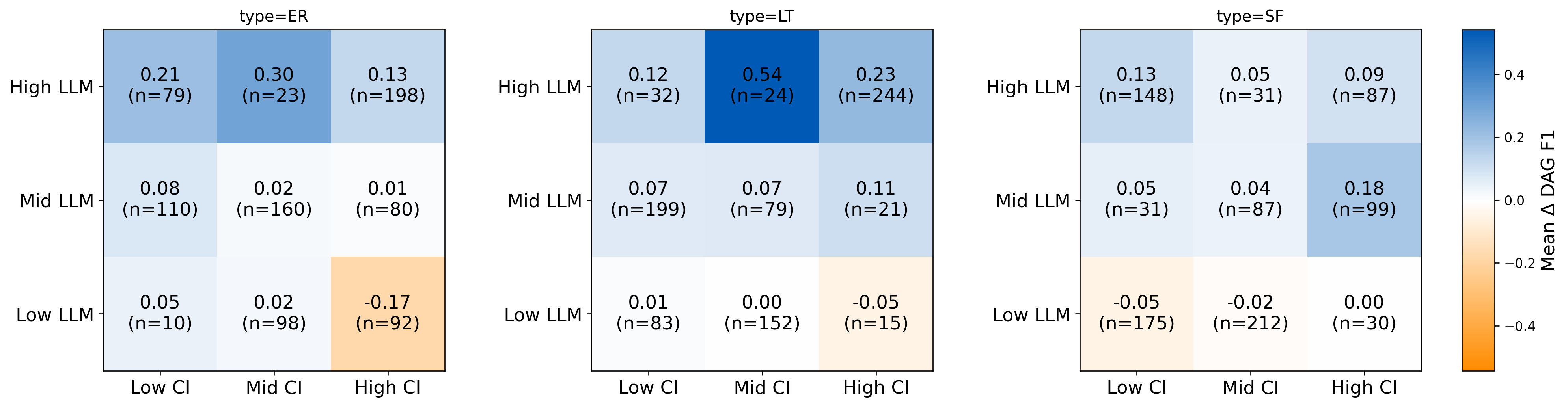}
    \caption{\texttt{CauseNet}: interaction heatmaps split by graph type, using the same global bins as Fig.~\ref{fig:constraint_quality}.}
    \label{fig:constraint_quality_per_graph_type}
\end{figure*}

Figures~\ref{fig:constraint_quality_per_nodesize} and \ref{fig:constraint_quality_per_graph_type} provide the same \texttt{CauseNet} analysis split by node count and graph type. They show that the qualitative pattern is not an artefact of one graph family: high-quality semantic constraints tend to help across sizes and ER/SF/LT topologies, while low-quality semantic constraints are risky precisely when the data-derived signal is already informative.

\begin{figure*}[ht]
    \centering
    \includegraphics[width=\linewidth]{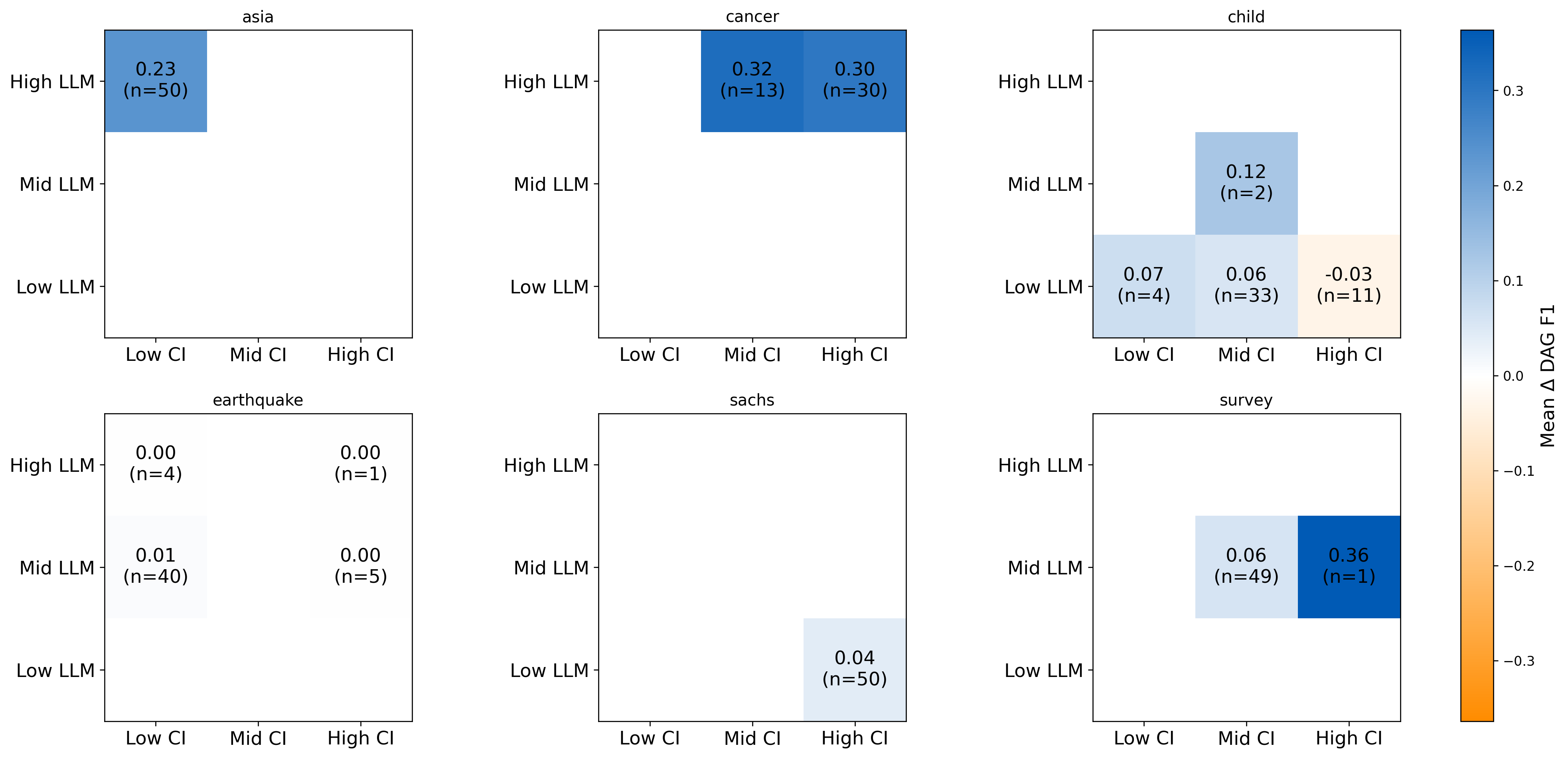}
    \caption{\texttt{bnlearn}: per-dataset interaction heatmaps using the same global bins as Fig.~\ref{fig:heapmap_quality_bn}, allowing absolute LLM-constraint-quality comparisons across benchmarks.}
    \label{fig:heatmap_per_bndataset_globcut}
\end{figure*}

\begin{figure*}[ht]
    \centering
    \includegraphics[width=\linewidth]{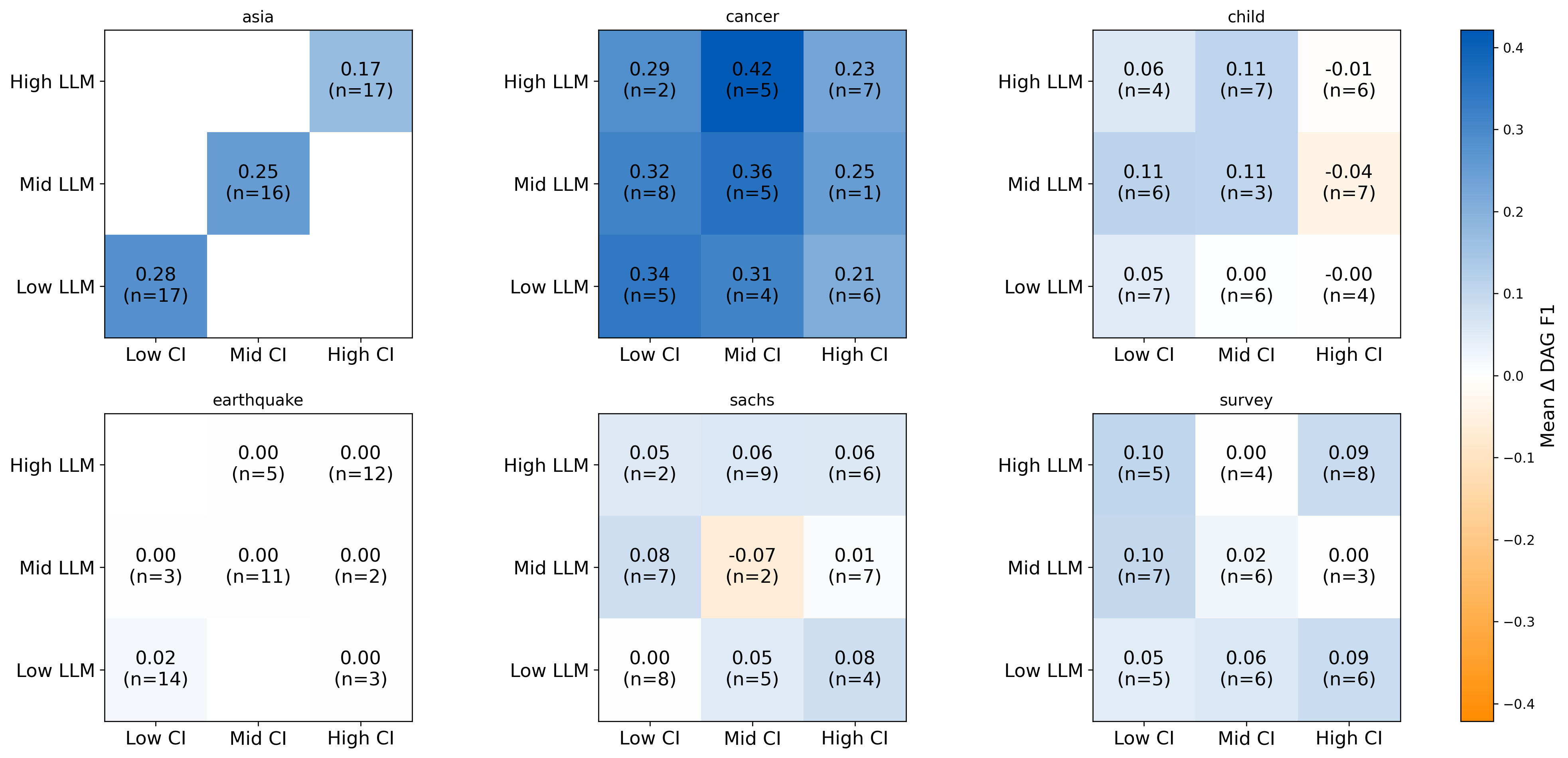}
    \caption{\texttt{bnlearn}: per-dataset interaction heatmaps using within-dataset quantile bins, highlighting relative variation inside each benchmark.}
    \label{fig:heatmap_per_bndataset}
\end{figure*}

Figures~\ref{fig:heatmap_per_bndataset_globcut} and \ref{fig:heatmap_per_bndataset} split the \texttt{bnlearn} heatmap by dataset. The global-bin version is useful for comparing absolute LLM-constraint quality across datasets; the within-dataset quantile version is useful for seeing whether better LLM-derived constraints help inside each benchmark even when a dataset occupies only a narrow global quality range. The public benchmarks concentrate in fewer bins than \texttt{CauseNet}, again consistent with smaller graphs and possible benchmark familiarity.

\clearpage
\subsubsection{Statistical Test Results}
\label{sec:appendix_statistical_tests}
The following tables report Welch tests used by the result tables throughout the paper. There are two kinds of test summaries. Indicator-test tables compare a bold top entry with the next non-bold method in the same metric/group ranking; these rows explain the significance symbols shown in the result tables. Tied-bolding tables list the non-significant best-vs-comparator tests that justify bolding methods other than the numeric best. In all cases, $p$-values are Benjamini--Hochberg~\citep{benjamini1995controlling} corrected within the corresponding metric/group family.

For \texttt{CauseNet}, Table~\ref{tab:causenet_main_indicator_tests} gives the indicator tests for the main node-count table, while Table~\ref{tab:structural_tied_bolding_tests} collects the tied-bolding tests for the structural \texttt{CauseNet} tables and the structural ablations. These two tables should be read together: the first explains the significance symbols against the next non-bold method, and the second explains why multiple top methods can be bolded when they are not significantly different.
\begin{table*}[t]
    \caption{Welch tests for Table~\ref{tab:causenet_dag_full_metrics}. Each row compares a bolded top entry with the next non-bold method in that column. Significance levels for both reported $p$-values are: \textnormal{***} $p<0.001$, \textnormal{**} $p<0.01$, \textnormal{*} $p<0.05$, \textnormal{.} $p<0.1$.}
    \label{tab:causenet_main_indicator_tests}
    \centering
    \scriptsize
    \setlength{\tabcolsep}{2pt}
    \resizebox{\textwidth}{!}{%
    \begin{tabular}{ll l|c|r|rl|rl}
    \toprule
    \textbf{Group} & \textbf{Metric} & \textbf{Methods} & \textbf{Means$\pm$Std} & \textbf{t} & \multicolumn{2}{c|}{\textbf{p-value}} & \multicolumn{2}{c}{\textbf{$p_{\mathrm{BH}}$}} \\
    \midrule
    $|\nodeSet|=5$ & SHD & ABAPC-LLM vs MPC-LLM & \num{1.230 +- 0.369} vs \num{1.500 +- 0.318} & $-5.60$ & $2.19\text{e-}8$ & \!\!\!*** & $2.19\text{e-}8$ & \!\!\!*** \\
    $|\nodeSet|=5$ & Precision & MPC-LLM vs ABAPC-LLM & \num{0.958 +- 0.031} vs \num{0.882 +- 0.046} & $12.94$ & $2.57\text{e-}38$ & \!\!\!*** & $2.57\text{e-}38$ & \!\!\!*** \\
    $|\nodeSet|=5$ & Recall & ABAPC-LLM vs MPC-LLM & \num{0.763 +- 0.062} vs \num{0.710 +- 0.050} & $5.12$ & $3.13\text{e-}7$ & \!\!\!*** & $3.13\text{e-}7$ & \!\!\!*** \\
    $|\nodeSet|=5$ & F1 & ABAPC-LLM vs MPC-LLM & \num{0.812 +- 0.049} vs \num{0.795 +- 0.037} & $1.98$ & $0.048$ & \!\!\!* & $0.048$ & \!\!\!* \\
    \midrule
    $|\nodeSet|=10$ & SHD & ABAPC-LLM vs ABAPC & \num{5.626 +- 1.176} vs \num{6.166 +- 1.442} & $-5.03$ & $4.89\text{e-}7$ & \!\!\!*** & $4.89\text{e-}7$ & \!\!\!*** \\
    $|\nodeSet|=10$ & Precision & BOSS vs MPC-LLM & \num{0.880 +- 0.197} vs \num{0.856 +- 0.088} & $4.34$ & $1.40\text{e-}5$ & \!\!\!*** & $1.40\text{e-}5$ & \!\!\!*** \\
    $|\nodeSet|=10$ & Recall & ABAPC-LLM vs ABAPC & \num{0.565 +- 0.086} vs \num{0.535 +- 0.105} & $5.40$ & $6.63\text{e-}8$ & \!\!\!*** & $6.63\text{e-}8$ & \!\!\!*** \\
    $|\nodeSet|=10$ & F1 & ABAPC-LLM vs MPC-LLM & \num{0.653 +- 0.084} vs \num{0.627 +- 0.077} & $4.38$ & $1.17\text{e-}5$ & \!\!\!*** & $1.17\text{e-}5$ & \!\!\!*** \\
    \midrule
    $|\nodeSet|=15$ & SHD & ABAPC-LLM vs MPC-LLM & \num{7.550 +- 1.281} vs \num{7.874 +- 1.038} & $-2.94$ & $0.003$ & \!\!\!** & $0.003$ & \!\!\!** \\
    $|\nodeSet|=15$ & Precision & MPC vs BOSS & \num{0.936 +- 0.071} vs \num{0.880 +- 0.191} & $18.74$ & $2.28\text{e-}78$ & \!\!\!*** & $4.55\text{e-}78$ & \!\!\!*** \\
    $|\nodeSet|=15$ & Precision & MPC-LLM vs BOSS & \num{0.929 +- 0.061} vs \num{0.880 +- 0.191} & $16.57$ & $1.14\text{e-}61$ & \!\!\!*** & $1.14\text{e-}61$ & \!\!\!*** \\
    $|\nodeSet|=15$ & Recall & ABAPC-LLM vs ABAPC & \num{0.565 +- 0.074} vs \num{0.546 +- 0.087} & $4.42$ & $9.95\text{e-}6$ & \!\!\!*** & $9.95\text{e-}6$ & \!\!\!*** \\
    $|\nodeSet|=15$ & F1 & MPC-LLM vs ABAPC-LLM & \num{0.667 +- 0.058} vs \num{0.654 +- 0.072} & $2.52$ & $0.012$ & \!\!\!* & $0.012$ & \!\!\!* \\
    \bottomrule
    \end{tabular}
    }
\end{table*}
\begin{table*}[t]
    \caption{Non-significant best-vs-comparator tests that justify additional bolded methods in Tables~\ref{tab:causenet_dag_full_metrics}, \ref{tab:causenet_dag_v_graph_type}, \ref{tab:alpha_ablation_dag}, and \ref{tab:gpt_model_ablation_dag}. Significance levels for both reported $p$-values are: \textnormal{***} $p<0.001$, \textnormal{**} $p<0.01$, \textnormal{*} $p<0.05$, \textnormal{.} $p<0.1$.}
    \label{tab:structural_tied_bolding_tests}
    \centering
    \scriptsize
    \setlength{\tabcolsep}{2pt}
    \resizebox{\textwidth}{!}{%
    \begin{tabular}{cll l|c|r|rl|rl}
    \toprule
    \textbf{Table} & \textbf{Group} & \textbf{Metric} & \textbf{Methods} & \textbf{Means$\pm$Std} & \textbf{t} & \multicolumn{2}{c|}{\textbf{p-value}} & \multicolumn{2}{c}{\textbf{$p_{\mathrm{BH}}$}} \\
    \midrule
    \ref{tab:causenet_dag_full_metrics} & $|\nodeSet|=15$ & Precision & MPC vs MPC-LLM & \num{0.936 +- 0.071} vs \num{0.929 +- 0.061} & $1.91$ & $0.056$ & \!\!\!. & $0.056$ & \!\!\!. \\
    \midrule
    \ref{tab:causenet_dag_v_graph_type} & \shortstack[l]{SF\\$|\nodeSet|=5$} & F1 & ABAPC-LLM vs MPC-LLM & \num{0.734 +- 0.046} vs \num{0.697 +- 0.055} & $1.81$ & $0.071$ & \!\!\!. & $0.071$ & \!\!\!. \\
    \midrule
    \ref{tab:causenet_dag_v_graph_type} & \shortstack[l]{LT\\$|\nodeSet|=5$} & SHD & MPC-LLM vs ABAPC-LLM & \num{1.217 +- 0.345} vs \num{1.335 +- 0.457} & $-1.20$ & $0.231$ & \!\!\!ns & $0.231$ & \!\!\!ns \\
    \midrule
    \ref{tab:causenet_dag_v_graph_type} & \shortstack[l]{ER\\$|\nodeSet|=10$} & F1 & MPC-LLM vs ABAPC-LLM & \num{0.663 +- 0.057} vs \num{0.656 +- 0.057} & $0.57$ & $0.570$ & \!\!\!ns & $0.570$ & \!\!\!ns \\
    \midrule
    \ref{tab:causenet_dag_v_graph_type} & \shortstack[l]{SF\\$|\nodeSet|=10$} & SHD & ABAPC-LLM vs MPC-LLM & \num{5.711 +- 1.302} vs \num{6.112 +- 1.198} & $-1.95$ & $0.051$ & \!\!\!. & $0.051$ & \!\!\!. \\
    \midrule
    \ref{tab:causenet_dag_v_graph_type} & \shortstack[l]{LT\\$|\nodeSet|=10$} & F1 & BOSS vs ABAPC-LLM & \num{0.638 +- 0.192} vs \num{0.623 +- 0.111} & $1.67$ & $0.095$ & \!\!\!. & $0.095$ & \!\!\!. \\
    \midrule
    \ref{tab:causenet_dag_v_graph_type} & \shortstack[l]{SF\\$|\nodeSet|=15$} & SHD & ABAPC vs ABAPC-LLM & \num{7.103 +- 1.144} vs \num{7.226 +- 1.151} & $-1.71$ & $0.087$ & \!\!\!. & $0.087$ & \!\!\!. \\
    \ref{tab:causenet_dag_v_graph_type} & \shortstack[l]{SF\\$|\nodeSet|=15$} & F1 & ABAPC vs ABAPC-LLM & \num{0.601 +- 0.080} vs \num{0.590 +- 0.077} & $1.84$ & $0.065$ & \!\!\!. & $0.065$ & \!\!\!. \\
    \midrule
    \ref{tab:causenet_dag_v_graph_type} & \shortstack[l]{ER\\Avg.} & F1 & ABAPC-LLM vs MPC-LLM & \num{0.720 +- 0.056} vs \num{0.715 +- 0.048} & $0.70$ & $0.481$ & \!\!\!ns & $0.481$ & \!\!\!ns \\
    \midrule
    \ref{tab:causenet_dag_v_graph_type} & \shortstack[l]{LT\\Avg.} & SHD & ABAPC-LLM vs MPC-LLM & \num{4.868 +- 1.114} vs \num{4.912 +- 0.816} & $-0.27$ & $0.784$ & \!\!\!ns & $0.784$ & \!\!\!ns \\
    \ref{tab:causenet_dag_v_graph_type} & \shortstack[l]{LT\\Avg.} & F1 & MPC-LLM vs ABAPC-LLM & \num{0.739 +- 0.052} vs \num{0.731 +- 0.079} & $1.14$ & $0.253$ & \!\!\!ns & $0.253$ & \!\!\!ns \\
    \midrule
    \ref{tab:alpha_ablation_dag} & \shortstack[l]{$\alpha=0.05$\\$|\nodeSet|=5$} & SHD & MPC-LLM vs ABAPC-LLM & \num{1.170 +- 0.583} vs \num{1.233 +- 0.514} & $-1.45$ & $0.146$ & \!\!\!ns & $0.146$ & \!\!\!ns \\
    \midrule
    \ref{tab:gpt_model_ablation_dag} & $|\nodeSet|=5$ & F1 & ABAPC-LLM (Gemini) vs MPC-LLM (Gemini) & \num{0.828 +- 0.048} vs \num{0.822 +- 0.041} & $0.67$ & $0.506$ & \!\!\!ns & $0.506$ & \!\!\!ns \\
    \midrule
    \ref{tab:gpt_model_ablation_dag} & $|\nodeSet|=10$ & SHD & ABAPC-LLM (GPT) vs ABAPC-LLM (Gemini) & \num{5.755 +- 1.417} vs \num{5.984 +- 1.257} & $-1.95$ & $0.051$ & \!\!\!. & $0.051$ & \!\!\!. \\
    \midrule
    \ref{tab:gpt_model_ablation_dag} & $|\nodeSet|=15$ & SHD & ABAPC-LLM (Gemini) vs MPC-LLM (Gemini) & \num{7.508 +- 1.292} vs \num{7.616 +- 1.052} & $-0.84$ & $0.400$ & \!\!\!ns & $0.444$ & \!\!\!ns \\
    \ref{tab:gpt_model_ablation_dag} & $|\nodeSet|=15$ & SHD & ABAPC-LLM (Gemini) vs ABAPC-LLM (GPT) & \num{7.508 +- 1.292} vs \num{7.575 +- 1.513} & $-0.52$ & $0.604$ & \!\!\!ns & $0.604$ & \!\!\!ns \\
    \ref{tab:gpt_model_ablation_dag} & $|\nodeSet|=15$ & Precision & MPC-LLM (GPT) vs MPC & \num{0.941 +- 0.064} vs \num{0.934 +- 0.068} & $1.67$ & $0.094$ & \!\!\!. & $0.094$ & \!\!\!. \\
    \ref{tab:gpt_model_ablation_dag} & $|\nodeSet|=15$ & Precision & MPC-LLM (GPT) vs MPC-LLM (Gemini) & \num{0.941 +- 0.064} vs \num{0.934 +- 0.049} & $1.69$ & $0.092$ & \!\!\!. & $0.094$ & \!\!\!. \\
    \ref{tab:gpt_model_ablation_dag} & $|\nodeSet|=15$ & Recall & ABAPC-LLM (Gemini) vs ABAPC-LLM (GPT) & \num{0.573 +- 0.073} vs \num{0.570 +- 0.086} & $0.78$ & $0.435$ & \!\!\!ns & $0.435$ & \!\!\!ns \\
    \bottomrule
    \end{tabular}
    }
\end{table*}
Table~\ref{tab:causenet_type_size_indicator_tests} reports the corresponding indicator tests for the \texttt{CauseNet} graph-type/node-count breakdown, including its marginal averages. Table~\ref{tab:alpha_indicator_tests} gives the indicator tests for the \texttt{CauseNet} $\alpha$ threshold ablation.
\begin{table*}[t]
    \caption{Welch tests for Table~\ref{tab:causenet_dag_v_graph_type}, including marginal averages. Significance levels for both reported $p$-values are: \textnormal{***} $p<0.001$, \textnormal{**} $p<0.01$, \textnormal{*} $p<0.05$, \textnormal{.} $p<0.1$.}
    \label{tab:causenet_type_size_indicator_tests}
    \centering
    \scriptsize
    \setlength{\tabcolsep}{2pt}
    \resizebox{\textwidth}{!}{%
    \begin{tabular}{ll l|c|r|rl|rl}
    \toprule
    \textbf{Group} & \textbf{Metric} & \textbf{Methods} & \textbf{Means$\pm$Std} & \textbf{t} & \multicolumn{2}{c|}{\textbf{p-value}} & \multicolumn{2}{c}{\textbf{$p_{\mathrm{BH}}$}} \\
    \midrule
    ER, $|\nodeSet|=5$ & SHD & ABAPC-LLM vs ABAPC & \num{1.054 +- 0.372} vs \num{1.515 +- 0.542} & $-7.91$ & $2.63\text{e-}15$ & \!\!\!*** & $2.63\text{e-}15$ & \!\!\!*** \\
    ER, $|\nodeSet|=5$ & F1 & ABAPC-LLM vs MPC-LLM & \num{0.843 +- 0.044} vs \num{0.805 +- 0.025} & $3.94$ & $8.25\text{e-}5$ & \!\!\!*** & $8.25\text{e-}5$ & \!\!\!*** \\
    \midrule
    SF, $|\nodeSet|=5$ & SHD & ABAPC-LLM vs MPC-LLM & \num{1.302 +- 0.278} vs \num{1.603 +- 0.305} & $-3.06$ & $0.002$ & \!\!\!** & $0.002$ & \!\!\!** \\
    SF, $|\nodeSet|=5$ & F1 & ABAPC-LLM vs ABAPC & \num{0.734 +- 0.046} vs \num{0.632 +- 0.075} & $7.05$ & $1.79\text{e-}12$ & \!\!\!*** & $3.59\text{e-}12$ & \!\!\!*** \\
    SF, $|\nodeSet|=5$ & F1 & MPC-LLM vs ABAPC & \num{0.697 +- 0.055} vs \num{0.632 +- 0.075} & $3.91$ & $9.07\text{e-}5$ & \!\!\!*** & $9.07\text{e-}5$ & \!\!\!*** \\
    \midrule
    LT, $|\nodeSet|=5$ & SHD & MPC-LLM vs ABAPC & \num{1.217 +- 0.345} vs \num{2.709 +- 0.722} & $-13.57$ & $5.71\text{e-}42$ & \!\!\!*** & $1.14\text{e-}41$ & \!\!\!*** \\
    LT, $|\nodeSet|=5$ & SHD & ABAPC-LLM vs ABAPC & \num{1.335 +- 0.457} vs \num{2.709 +- 0.722} & $-13.03$ & $8.06\text{e-}39$ & \!\!\!*** & $8.06\text{e-}39$ & \!\!\!*** \\
    LT, $|\nodeSet|=5$ & F1 & MPC-LLM vs ABAPC-LLM & \num{0.883 +- 0.030} vs \num{0.858 +- 0.057} & $2.65$ & $0.008$ & \!\!\!** & $0.008$ & \!\!\!** \\
    \midrule
    Avg., $|\nodeSet|=5$ & SHD & ABAPC-LLM vs MPC-LLM & \num{1.230 +- 0.369} vs \num{1.500 +- 0.318} & $-5.60$ & $2.19\text{e-}8$ & \!\!\!*** & $2.19\text{e-}8$ & \!\!\!*** \\
    Avg., $|\nodeSet|=5$ & F1 & ABAPC-LLM vs MPC-LLM & \num{0.812 +- 0.049} vs \num{0.795 +- 0.037} & $1.98$ & $0.048$ & \!\!\!* & $0.048$ & \!\!\!* \\
    \midrule
    ER, $|\nodeSet|=10$ & SHD & ABAPC-LLM vs MPC-LLM & \num{5.061 +- 0.791} vs \num{5.488 +- 0.772} & $-3.28$ & $0.001$ & \!\!\!** & $0.001$ & \!\!\!** \\
    ER, $|\nodeSet|=10$ & F1 & MPC-LLM vs ABAPC & \num{0.663 +- 0.057} vs \num{0.616 +- 0.094} & $4.20$ & $2.63\text{e-}5$ & \!\!\!*** & $5.25\text{e-}5$ & \!\!\!*** \\
    ER, $|\nodeSet|=10$ & F1 & ABAPC-LLM vs ABAPC & \num{0.656 +- 0.057} vs \num{0.616 +- 0.094} & $3.77$ & $1.65\text{e-}4$ & \!\!\!*** & $1.65\text{e-}4$ & \!\!\!*** \\
    \midrule
    SF, $|\nodeSet|=10$ & SHD & ABAPC-LLM vs ABAPC & \num{5.711 +- 1.302} vs \num{6.221 +- 1.493} & $-2.43$ & $0.015$ & \!\!\!* & $0.030$ & \!\!\!* \\
    SF, $|\nodeSet|=10$ & SHD & MPC-LLM vs ABAPC & \num{6.112 +- 1.198} vs \num{6.221 +- 1.493} & $-0.50$ & $0.618$ & \!\!\!ns & $0.618$ & \!\!\!ns \\
    SF, $|\nodeSet|=10$ & F1 & ABAPC-LLM vs MPC-LLM & \num{0.681 +- 0.086} vs \num{0.657 +- 0.092} & $2.69$ & $0.007$ & \!\!\!** & $0.007$ & \!\!\!** \\
    \midrule
    LT, $|\nodeSet|=10$ & SHD & ABAPC-LLM vs BOSS & \num{6.105 +- 1.437} vs \num{6.500 +- 2.993} & $-2.34$ & $0.019$ & \!\!\!* & $0.019$ & \!\!\!* \\
    LT, $|\nodeSet|=10$ & F1 & BOSS vs GRaSP & \num{0.638 +- 0.192} vs \num{0.602 +- 0.205} & $5.88$ & $4.15\text{e-}9$ & \!\!\!*** & $8.30\text{e-}9$ & \!\!\!*** \\
    LT, $|\nodeSet|=10$ & F1 & ABAPC-LLM vs GRaSP & \num{0.623 +- 0.111} vs \num{0.602 +- 0.205} & $2.33$ & $0.020$ & \!\!\!* & $0.020$ & \!\!\!* \\
    \midrule
    Avg., $|\nodeSet|=10$ & SHD & ABAPC-LLM vs ABAPC & \num{5.626 +- 1.176} vs \num{6.166 +- 1.442} & $-5.03$ & $4.89\text{e-}7$ & \!\!\!*** & $4.89\text{e-}7$ & \!\!\!*** \\
    Avg., $|\nodeSet|=10$ & F1 & ABAPC-LLM vs MPC-LLM & \num{0.653 +- 0.084} vs \num{0.627 +- 0.077} & $4.38$ & $1.17\text{e-}5$ & \!\!\!*** & $1.17\text{e-}5$ & \!\!\!*** \\
    \midrule
    ER, $|\nodeSet|=15$ & SHD & ABAPC-LLM vs MPC-LLM & \num{8.259 +- 1.245} vs \num{8.815 +- 1.132} & $-2.53$ & $0.011$ & \!\!\!* & $0.011$ & \!\!\!* \\
    ER, $|\nodeSet|=15$ & F1 & MPC-LLM vs ABAPC-LLM & \num{0.677 +- 0.062} vs \num{0.660 +- 0.068} & $3.26$ & $0.001$ & \!\!\!** & $0.001$ & \!\!\!** \\
    \midrule
    SF, $|\nodeSet|=15$ & SHD & ABAPC vs MPC-LLM & \num{7.103 +- 1.144} vs \num{8.307 +- 0.978} & $-12.22$ & $2.55\text{e-}34$ & \!\!\!*** & $5.11\text{e-}34$ & \!\!\!*** \\
    SF, $|\nodeSet|=15$ & SHD & ABAPC-LLM vs MPC-LLM & \num{7.226 +- 1.151} vs \num{8.307 +- 0.978} & $-11.20$ & $4.12\text{e-}29$ & \!\!\!*** & $4.12\text{e-}29$ & \!\!\!*** \\
    SF, $|\nodeSet|=15$ & F1 & ABAPC vs MPC & \num{0.601 +- 0.080} vs \num{0.557 +- 0.073} & $6.04$ & $1.59\text{e-}9$ & \!\!\!*** & $3.17\text{e-}9$ & \!\!\!*** \\
    SF, $|\nodeSet|=15$ & F1 & ABAPC-LLM vs MPC & \num{0.590 +- 0.077} vs \num{0.557 +- 0.073} & $4.60$ & $4.25\text{e-}6$ & \!\!\!*** & $4.25\text{e-}6$ & \!\!\!*** \\
    \midrule
    LT, $|\nodeSet|=15$ & SHD & MPC-LLM vs ABAPC-LLM & \num{6.502 +- 1.003} vs \num{7.165 +- 1.447} & $-2.90$ & $0.004$ & \!\!\!** & $0.004$ & \!\!\!** \\
    LT, $|\nodeSet|=15$ & F1 & MPC-LLM vs ABAPC-LLM & \num{0.775 +- 0.042} vs \num{0.712 +- 0.070} & $12.24$ & $1.93\text{e-}34$ & \!\!\!*** & $1.93\text{e-}34$ & \!\!\!*** \\
    \midrule
    Avg., $|\nodeSet|=15$ & SHD & ABAPC-LLM vs MPC-LLM & \num{7.550 +- 1.281} vs \num{7.874 +- 1.038} & $-2.94$ & $0.003$ & \!\!\!** & $0.003$ & \!\!\!** \\
    Avg., $|\nodeSet|=15$ & F1 & MPC-LLM vs ABAPC-LLM & \num{0.667 +- 0.058} vs \num{0.654 +- 0.072} & $2.52$ & $0.012$ & \!\!\!* & $0.012$ & \!\!\!* \\
    \midrule
    ER, Avg. & SHD & ABAPC-LLM vs MPC-LLM & \num{4.791 +- 0.803} vs \num{5.328 +- 0.736} & $-3.27$ & $0.001$ & \!\!\!** & $0.001$ & \!\!\!** \\
    ER, Avg. & F1 & ABAPC-LLM vs ABAPC & \num{0.720 +- 0.056} vs \num{0.667 +- 0.090} & $8.73$ & $2.50\text{e-}18$ & \!\!\!*** & $5.01\text{e-}18$ & \!\!\!*** \\
    ER, Avg. & F1 & MPC-LLM vs ABAPC & \num{0.715 +- 0.048} vs \num{0.667 +- 0.090} & $7.96$ & $1.72\text{e-}15$ & \!\!\!*** & $1.72\text{e-}15$ & \!\!\!*** \\
    \midrule
    SF, Avg. & SHD & ABAPC-LLM vs ABAPC & \num{4.747 +- 0.910} vs \num{5.058 +- 1.007} & $-2.26$ & $0.024$ & \!\!\!* & $0.024$ & \!\!\!* \\
    SF, Avg. & F1 & ABAPC-LLM vs MPC-LLM & \num{0.668 +- 0.070} vs \num{0.634 +- 0.073} & $4.31$ & $1.62\text{e-}5$ & \!\!\!*** & $1.62\text{e-}5$ & \!\!\!*** \\
    \midrule
    LT, Avg. & SHD & ABAPC-LLM vs ABAPC & \num{4.868 +- 1.114} vs \num{5.676 +- 1.387} & $-5.05$ & $4.31\text{e-}7$ & \!\!\!*** & $8.62\text{e-}7$ & \!\!\!*** \\
    LT, Avg. & SHD & MPC-LLM vs ABAPC & \num{4.912 +- 0.816} vs \num{5.676 +- 1.387} & $-4.88$ & $1.06\text{e-}6$ & \!\!\!*** & $1.06\text{e-}6$ & \!\!\!*** \\
    LT, Avg. & F1 & MPC-LLM vs BOSS & \num{0.739 +- 0.052} vs \num{0.626 +- 0.166} & $17.32$ & $3.32\text{e-}67$ & \!\!\!*** & $3.32\text{e-}67$ & \!\!\!*** \\
    LT, Avg. & F1 & ABAPC-LLM vs BOSS & \num{0.731 +- 0.079} vs \num{0.626 +- 0.166} & $17.83$ & $4.11\text{e-}71$ & \!\!\!*** & $8.21\text{e-}71$ & \!\!\!*** \\
    \midrule
    Avg. & SHD & ABAPC-LLM vs MPC-LLM & \num{4.802 +- 0.942} vs \num{5.193 +- 0.793} & $-4.38$ & $1.18\text{e-}5$ & \!\!\!*** & $1.18\text{e-}5$ & \!\!\!*** \\
    Avg. & F1 & ABAPC-LLM vs MPC-LLM & \num{0.706 +- 0.069} vs \num{0.696 +- 0.057} & $2.42$ & $0.015$ & \!\!\!* & $0.015$ & \!\!\!* \\
    \bottomrule
    \end{tabular}
    }
\end{table*}
\begin{table*}[t]
    \caption{Welch tests for Table~\ref{tab:alpha_ablation_dag}. Significance levels for both reported $p$-values are: \textnormal{***} $p<0.001$, \textnormal{**} $p<0.01$, \textnormal{*} $p<0.05$, \textnormal{.} $p<0.1$.}
    \label{tab:alpha_indicator_tests}
    \centering
    \scriptsize
    \setlength{\tabcolsep}{2pt}
    \resizebox{\textwidth}{!}{%
    \begin{tabular}{ll l|c|r|rl|rl}
    \toprule
    \textbf{Group} & \textbf{Metric} & \textbf{Methods} & \textbf{Means$\pm$Std} & \textbf{t} & \multicolumn{2}{c|}{\textbf{p-value}} & \multicolumn{2}{c}{\textbf{$p_{\mathrm{BH}}$}} \\
    \midrule
    $\alpha=0.01$, $|\nodeSet|=5$ & SHD & ABAPC-LLM vs MPC-LLM & \num{1.230 +- 0.369} vs \num{1.500 +- 0.318} & $-5.60$ & $2.19\text{e-}8$ & \!\!\!*** & $2.19\text{e-}8$ & \!\!\!*** \\
    $\alpha=0.01$, $|\nodeSet|=5$ & F1 & ABAPC-LLM vs MPC-LLM & \num{0.812 +- 0.049} vs \num{0.795 +- 0.037} & $1.98$ & $0.048$ & \!\!\!* & $0.048$ & \!\!\!* \\
    \midrule
    $\alpha=0.01$, $|\nodeSet|=10$ & SHD & ABAPC-LLM vs ABAPC & \num{5.626 +- 1.176} vs \num{6.166 +- 1.442} & $-5.03$ & $4.89\text{e-}7$ & \!\!\!*** & $4.89\text{e-}7$ & \!\!\!*** \\
    $\alpha=0.01$, $|\nodeSet|=10$ & F1 & ABAPC-LLM vs MPC-LLM & \num{0.653 +- 0.084} vs \num{0.627 +- 0.077} & $4.38$ & $1.17\text{e-}5$ & \!\!\!*** & $1.17\text{e-}5$ & \!\!\!*** \\
    \midrule
    $\alpha=0.01$, $|\nodeSet|=15$ & SHD & ABAPC-LLM vs MPC-LLM & \num{7.550 +- 1.281} vs \num{7.874 +- 1.038} & $-2.94$ & $0.003$ & \!\!\!** & $0.003$ & \!\!\!** \\
    $\alpha=0.01$, $|\nodeSet|=15$ & F1 & MPC-LLM vs ABAPC-LLM & \num{0.667 +- 0.058} vs \num{0.654 +- 0.072} & $2.52$ & $0.012$ & \!\!\!* & $0.012$ & \!\!\!* \\
    \midrule
    $\alpha=0.05$, $|\nodeSet|=5$ & SHD & MPC-LLM vs ABAPC & \num{1.170 +- 0.583} vs \num{1.894 +- 0.695} & $-16.90$ & $4.28\text{e-}64$ & \!\!\!*** & $8.56\text{e-}64$ & \!\!\!*** \\
    $\alpha=0.05$, $|\nodeSet|=5$ & SHD & ABAPC-LLM vs ABAPC & \num{1.233 +- 0.514} vs \num{1.894 +- 0.695} & $-14.12$ & $2.85\text{e-}45$ & \!\!\!*** & $2.85\text{e-}45$ & \!\!\!*** \\
    $\alpha=0.05$, $|\nodeSet|=5$ & F1 & MPC-LLM vs ABAPC-LLM & \num{0.850 +- 0.073} vs \num{0.817 +- 0.064} & $4.79$ & $1.70\text{e-}6$ & \!\!\!*** & $1.70\text{e-}6$ & \!\!\!*** \\
    \midrule
    $\alpha=0.05$, $|\nodeSet|=10$ & SHD & MPC-LLM vs MPC & \num{4.531 +- 1.516} vs \num{5.267 +- 1.791} & $-8.48$ & $2.23\text{e-}17$ & \!\!\!*** & $2.23\text{e-}17$ & \!\!\!*** \\
    $\alpha=0.05$, $|\nodeSet|=10$ & F1 & MPC-LLM vs MPC & \num{0.734 +- 0.103} vs \num{0.691 +- 0.131} & $9.40$ & $5.46\text{e-}21$ & \!\!\!*** & $5.46\text{e-}21$ & \!\!\!*** \\
    \midrule
    $\alpha=0.05$, $|\nodeSet|=15$ & SHD & MPC-LLM vs MPC & \num{6.439 +- 1.769} vs \num{7.289 +- 1.971} & $-7.84$ & $4.63\text{e-}15$ & \!\!\!*** & $4.63\text{e-}15$ & \!\!\!*** \\
    $\alpha=0.05$, $|\nodeSet|=15$ & F1 & MPC-LLM vs MPC & \num{0.721 +- 0.092} vs \num{0.684 +- 0.106} & $8.67$ & $4.19\text{e-}18$ & \!\!\!*** & $4.19\text{e-}18$ & \!\!\!*** \\
    \bottomrule
    \end{tabular}
    }
\end{table*}
For \texttt{bnlearn}, Tables~\ref{tab:bnlearn_earthquake_indicator_tests}--\ref{tab:bnlearn_child_indicator_tests} split the indicator tests by dataset blocks to keep the tables readable, and Table~\ref{tab:bnlearn_tied_bolding_tests} gives the tied-bolding tests for the same benchmark table.
\begin{table*}[t]
    \caption{Welch tests for the \texttt{earthquake} rows in Table~\ref{tab:bnlearn_dag_all_metrics}. Significance levels for both reported $p$-values are: \textnormal{***} $p<0.001$, \textnormal{**} $p<0.01$, \textnormal{*} $p<0.05$, \textnormal{.} $p<0.1$.}
    \label{tab:bnlearn_earthquake_indicator_tests}
    \centering
    \scriptsize
    \setlength{\tabcolsep}{2pt}
    \resizebox{\textwidth}{!}{%
    \begin{tabular}{ll l|c|r|rl|rl}
    \toprule
    \textbf{Group} & \textbf{Metric} & \textbf{Methods} & \textbf{Means$\pm$Std} & \textbf{t} & \multicolumn{2}{c|}{\textbf{p-value}} & \multicolumn{2}{c}{\textbf{$p_{\mathrm{BH}}$}} \\
    \midrule
    earthquake & SHD & MPC vs NOTEARS-MLP & \num{0.040 +- 0.200} vs \num{2.120 +- 1.170} & $-12.39$ & $2.92\text{e-}35$ & \!\!\!*** & $8.77\text{e-}35$ & \!\!\!*** \\
    earthquake & SHD & MPC-LLM vs NOTEARS-MLP & \num{0.040 +- 0.200} vs \num{2.120 +- 1.170} & $-12.39$ & $2.92\text{e-}35$ & \!\!\!*** & $8.77\text{e-}35$ & \!\!\!*** \\
    earthquake & SHD & ABAPC-LLM vs NOTEARS-MLP & \num{0.080 +- 0.340} vs \num{2.120 +- 1.170} & $-11.84$ & $2.48\text{e-}32$ & \!\!\!*** & $4.97\text{e-}32$ & \!\!\!*** \\
    earthquake & SHD & ABAPC vs NOTEARS-MLP & \num{0.100 +- 0.463} vs \num{2.120 +- 1.170} & $-11.35$ & $7.25\text{e-}30$ & \!\!\!*** & $1.09\text{e-}29$ & \!\!\!*** \\
    earthquake & SHD & GRaSP vs NOTEARS-MLP & \num{0.160 +- 0.790} vs \num{2.120 +- 1.170} & $-9.82$ & $9.49\text{e-}23$ & \!\!\!*** & $1.14\text{e-}22$ & \!\!\!*** \\
    earthquake & SHD & BOSS vs NOTEARS-MLP & \num{0.400 +- 1.210} vs \num{2.120 +- 1.170} & $-7.23$ & $4.98\text{e-}13$ & \!\!\!*** & $4.98\text{e-}13$ & \!\!\!*** \\
    earthquake & Precision & GRaSP vs NOTEARS-MLP & \num{1.000 +- 0.000} vs \num{0.500 +- 0.270} & $13.09$ & $3.54\text{e-}39$ & \!\!\!*** & $1.06\text{e-}38$ & \!\!\!*** \\
    earthquake & Precision & BOSS vs NOTEARS-MLP & \num{1.000 +- 0.000} vs \num{0.500 +- 0.270} & $13.09$ & $3.54\text{e-}39$ & \!\!\!*** & $1.06\text{e-}38$ & \!\!\!*** \\
    earthquake & Precision & MPC vs NOTEARS-MLP & \num{0.990 +- 0.040} vs \num{0.500 +- 0.270} & $12.69$ & $6.37\text{e-}37$ & \!\!\!*** & $9.56\text{e-}37$ & \!\!\!*** \\
    earthquake & Precision & MPC-LLM vs NOTEARS-MLP & \num{0.990 +- 0.040} vs \num{0.500 +- 0.270} & $12.69$ & $6.37\text{e-}37$ & \!\!\!*** & $9.56\text{e-}37$ & \!\!\!*** \\
    earthquake & Precision & ABAPC-LLM vs NOTEARS-MLP & \num{0.987 +- 0.052} vs \num{0.500 +- 0.270} & $12.52$ & $5.71\text{e-}36$ & \!\!\!*** & $6.85\text{e-}36$ & \!\!\!*** \\
    earthquake & Precision & ABAPC vs NOTEARS-MLP & \num{0.982 +- 0.080} vs \num{0.500 +- 0.270} & $12.10$ & $1.02\text{e-}33$ & \!\!\!*** & $1.02\text{e-}33$ & \!\!\!*** \\
    earthquake & Recall & MPC vs BOSS & \num{1.000 +- 0.000} vs \num{0.900 +- 0.300} & $2.36$ & $0.018$ & \!\!\!* & $0.044$ & \!\!\!* \\
    earthquake & Recall & MPC-LLM vs BOSS & \num{1.000 +- 0.000} vs \num{0.900 +- 0.300} & $2.36$ & $0.018$ & \!\!\!* & $0.044$ & \!\!\!* \\
    earthquake & Recall & ABAPC-LLM vs BOSS & \num{0.995 +- 0.035} vs \num{0.900 +- 0.300} & $2.22$ & $0.026$ & \!\!\!* & $0.044$ & \!\!\!* \\
    earthquake & Recall & ABAPC vs BOSS & \num{0.990 +- 0.071} vs \num{0.900 +- 0.300} & $2.06$ & $0.039$ & \!\!\!* & $0.049$ & \!\!\!* \\
    earthquake & Recall & GRaSP vs BOSS & \num{0.960 +- 0.200} vs \num{0.900 +- 0.300} & $1.18$ & $0.239$ & \!\!\!ns & $0.239$ & \!\!\!ns \\
    earthquake & F1 & GRaSP vs NOTEARS-MLP & \num{1.000 +- 0.000} vs \num{0.500 +- 0.270} & $13.09$ & $3.54\text{e-}39$ & \!\!\!*** & $8.49\text{e-}39$ & \!\!\!*** \\
    earthquake & F1 & BOSS vs NOTEARS-MLP & \num{1.000 +- 0.000} vs \num{0.500 +- 0.270} & $13.09$ & $3.54\text{e-}39$ & \!\!\!*** & $8.49\text{e-}39$ & \!\!\!*** \\
    earthquake & F1 & MPC vs NOTEARS-MLP & \num{1.000 +- 0.020} vs \num{0.500 +- 0.270} & $13.06$ & $5.66\text{e-}39$ & \!\!\!*** & $8.49\text{e-}39$ & \!\!\!*** \\
    earthquake & F1 & MPC-LLM vs NOTEARS-MLP & \num{1.000 +- 0.020} vs \num{0.500 +- 0.270} & $13.06$ & $5.66\text{e-}39$ & \!\!\!*** & $8.49\text{e-}39$ & \!\!\!*** \\
    earthquake & F1 & ABAPC-LLM vs NOTEARS-MLP & \num{0.991 +- 0.041} vs \num{0.500 +- 0.270} & $12.70$ & $5.84\text{e-}37$ & \!\!\!*** & $7.00\text{e-}37$ & \!\!\!*** \\
    earthquake & F1 & ABAPC vs NOTEARS-MLP & \num{0.986 +- 0.073} vs \num{0.500 +- 0.270} & $12.27$ & $1.30\text{e-}34$ & \!\!\!*** & $1.30\text{e-}34$ & \!\!\!*** \\
    earthquake & SID & MPC vs BOSS & \num{0.000 +- 0.000} vs \num{1.000 +- 3.030} & $-2.33$ & $0.020$ & \!\!\!* & $0.049$ & \!\!\!* \\
    earthquake & SID & MPC-LLM vs BOSS & \num{0.000 +- 0.000} vs \num{1.000 +- 3.030} & $-2.33$ & $0.020$ & \!\!\!* & $0.049$ & \!\!\!* \\
    earthquake & SID & ABAPC-LLM vs BOSS & \num{0.080 +- 0.566} vs \num{1.000 +- 3.030} & $-2.11$ & $0.035$ & \!\!\!* & $0.058$ & \!\!\!. \\
    earthquake & SID & ABAPC vs BOSS & \num{0.160 +- 1.131} vs \num{1.000 +- 3.030} & $-1.84$ & $0.066$ & \!\!\!. & $0.083$ & \!\!\!. \\
    earthquake & SID & GRaSP vs BOSS & \num{0.400 +- 1.980} vs \num{1.000 +- 3.030} & $-1.17$ & $0.241$ & \!\!\!ns & $0.241$ & \!\!\!ns \\
    \bottomrule
    \end{tabular}
    }
\end{table*}
\begin{table*}[t]
    \caption{Welch tests for Table~\ref{tab:bnlearn_dag_all_metrics}, excluding the \texttt{earthquake} and \texttt{child} rows. Significance levels for both reported $p$-values are: \textnormal{***} $p<0.001$, \textnormal{**} $p<0.01$, \textnormal{*} $p<0.05$, \textnormal{.} $p<0.1$.}
    \label{tab:bnlearn_other_indicator_tests}
    \centering
    \scriptsize
    \setlength{\tabcolsep}{2pt}
    \resizebox{\textwidth}{!}{%
    \begin{tabular}{ll l|c|r|rl|rl}
    \toprule
    \textbf{Group} & \textbf{Metric} & \textbf{Methods} & \textbf{Means$\pm$Std} & \textbf{t} & \multicolumn{2}{c|}{\textbf{p-value}} & \multicolumn{2}{c}{\textbf{$p_{\mathrm{BH}}$}} \\
    \midrule
    asia & SHD & MPC-LLM vs ABAPC-LLM & \num{2.840 +- 0.470} vs \num{3.580 +- 0.710} & $-6.15$ & $7.89\text{e-}10$ & \!\!\!*** & $7.89\text{e-}10$ & \!\!\!*** \\
    asia & Precision & ABAPC-LLM vs ABAPC & \num{0.992 +- 0.040} vs \num{0.684 +- 0.207} & $10.31$ & $6.65\text{e-}25$ & \!\!\!*** & $9.97\text{e-}25$ & \!\!\!*** \\
    asia & Precision & MPC-LLM vs ABAPC & \num{0.990 +- 0.030} vs \num{0.684 +- 0.207} & $10.32$ & $5.96\text{e-}25$ & \!\!\!*** & $9.97\text{e-}25$ & \!\!\!*** \\
    asia & Precision & MPC vs ABAPC & \num{0.980 +- 0.080} vs \num{0.684 +- 0.207} & $9.41$ & $5.11\text{e-}21$ & \!\!\!*** & $5.11\text{e-}21$ & \!\!\!*** \\
    asia & Recall & MPC-LLM vs ABAPC-LLM & \num{0.650 +- 0.050} vs \num{0.557 +- 0.082} & $6.83$ & $8.49\text{e-}12$ & \!\!\!*** & $8.49\text{e-}12$ & \!\!\!*** \\
    asia & F1 & MPC-LLM vs ABAPC-LLM & \num{0.780 +- 0.040} vs \num{0.710 +- 0.068} & $6.22$ & $5.00\text{e-}10$ & \!\!\!*** & $5.00\text{e-}10$ & \!\!\!*** \\
    asia & SID & MPC-LLM vs ABAPC-LLM & \num{13.720 +- 0.540} vs \num{15.500 +- 2.308} & $-5.31$ & $1.10\text{e-}7$ & \!\!\!*** & $1.10\text{e-}7$ & \!\!\!*** \\
    \midrule
    cancer & SHD & MPC-LLM vs ABAPC & \num{0.680 +- 0.590} vs \num{2.106 +- 1.148} & $-7.81$ & $5.65\text{e-}15$ & \!\!\!*** & $5.65\text{e-}15$ & \!\!\!*** \\
    cancer & SHD & ABAPC-LLM vs ABAPC & \num{0.680 +- 0.587} vs \num{2.106 +- 1.148} & $-7.82$ & $5.28\text{e-}15$ & \!\!\!*** & $5.65\text{e-}15$ & \!\!\!*** \\
    cancer & Precision & ABAPC-LLM vs MPC & \num{0.996 +- 0.028} vs \num{0.940 +- 0.180} & $2.17$ & $0.030$ & \!\!\!* & $0.058$ & \!\!\!. \\
    cancer & Precision & MPC-LLM vs MPC & \num{0.990 +- 0.050} vs \num{0.940 +- 0.180} & $1.89$ & $0.058$ & \!\!\!. & $0.058$ & \!\!\!. \\
    cancer & Recall & MPC-LLM vs ABAPC & \num{0.840 +- 0.130} vs \num{0.478 +- 0.276} & $8.38$ & $5.51\text{e-}17$ & \!\!\!*** & $1.10\text{e-}16$ & \!\!\!*** \\
    cancer & Recall & ABAPC-LLM vs ABAPC & \num{0.835 +- 0.148} vs \num{0.478 +- 0.276} & $8.04$ & $8.72\text{e-}16$ & \!\!\!*** & $8.72\text{e-}16$ & \!\!\!*** \\
    cancer & F1 & MPC vs BOSS & \num{0.940 +- 0.180} vs \num{0.770 +- 0.300} & $3.44$ & $5.91\text{e-}4$ & \!\!\!*** & $0.002$ & \!\!\!** \\
    cancer & F1 & ABAPC-LLM vs BOSS & \num{0.901 +- 0.091} vs \num{0.770 +- 0.300} & $2.95$ & $0.003$ & \!\!\!** & $0.003$ & \!\!\!** \\
    cancer & F1 & MPC-LLM vs BOSS & \num{0.900 +- 0.080} vs \num{0.770 +- 0.300} & $2.96$ & $0.003$ & \!\!\!** & $0.003$ & \!\!\!** \\
    cancer & SID & MPC-LLM vs ABAPC & \num{0.780 +- 1.000} vs \num{8.070 +- 4.491} & $-11.20$ & $3.96\text{e-}29$ & \!\!\!*** & $7.93\text{e-}29$ & \!\!\!*** \\
    cancer & SID & ABAPC-LLM vs ABAPC & \num{0.960 +- 1.324} vs \num{8.070 +- 4.491} & $-10.74$ & $6.86\text{e-}27$ & \!\!\!*** & $6.86\text{e-}27$ & \!\!\!*** \\
    \midrule
    survey & SHD & MPC-LLM vs ABAPC-LLM & \num{2.060 +- 1.040} vs \num{3.297 +- 0.981} & $-6.12$ & $9.51\text{e-}10$ & \!\!\!*** & $9.51\text{e-}10$ & \!\!\!*** \\
    survey & Precision & GRaSP vs MPC-LLM & \num{1.000 +- 0.000} vs \num{0.890 +- 0.120} & $6.48$ & $9.06\text{e-}11$ & \!\!\!*** & $9.06\text{e-}11$ & \!\!\!*** \\
    survey & Precision & BOSS vs MPC-LLM & \num{1.000 +- 0.000} vs \num{0.890 +- 0.120} & $6.48$ & $9.06\text{e-}11$ & \!\!\!*** & $9.06\text{e-}11$ & \!\!\!*** \\
    survey & Recall & MPC-LLM vs ABAPC-LLM & \num{0.670 +- 0.160} vs \num{0.457 +- 0.155} & $6.76$ & $1.38\text{e-}11$ & \!\!\!*** & $1.38\text{e-}11$ & \!\!\!*** \\
    survey & F1 & MPC-LLM vs MPC & \num{0.750 +- 0.140} vs \num{0.680 +- 0.140} & $2.50$ & $0.012$ & \!\!\!* & $0.012$ & \!\!\!* \\
    survey & SID & MPC-LLM vs ABAPC-LLM & \num{6.320 +- 4.180} vs \num{13.477 +- 3.868} & $-8.89$ & $6.36\text{e-}19$ & \!\!\!*** & $6.36\text{e-}19$ & \!\!\!*** \\
    \midrule
    sachs & SHD & ABAPC-LLM vs ABAPC & \num{4.460 +- 1.264} vs \num{5.244 +- 1.684} & $-2.63$ & $0.008$ & \!\!\!** & $0.008$ & \!\!\!** \\
    sachs & Precision & ABAPC-LLM vs MPC-LLM & \num{0.759 +- 0.063} vs \num{0.740 +- 0.020} & $2.01$ & $0.044$ & \!\!\!* & $0.044$ & \!\!\!* \\
    sachs & Recall & ABAPC-LLM vs ABAPC & \num{0.739 +- 0.075} vs \num{0.693 +- 0.100} & $2.61$ & $0.009$ & \!\!\!** & $0.009$ & \!\!\!** \\
    sachs & F1 & ABAPC-LLM vs ABAPC & \num{0.748 +- 0.069} vs \num{0.706 +- 0.100} & $2.47$ & $0.014$ & \!\!\!* & $0.014$ & \!\!\!* \\
    sachs & SID & ABAPC-LLM vs ABAPC & \num{21.056 +- 6.983} vs \num{26.212 +- 9.472} & $-3.10$ & $0.002$ & \!\!\!** & $0.002$ & \!\!\!** \\
    \bottomrule
    \end{tabular}
    }
\end{table*}
\begin{table*}[t]
    \caption{Welch tests for the \texttt{child} rows in Table~\ref{tab:bnlearn_dag_all_metrics}. Significance levels for both reported $p$-values are: \textnormal{***} $p<0.001$, \textnormal{**} $p<0.01$, \textnormal{*} $p<0.05$, \textnormal{.} $p<0.1$.}
    \label{tab:bnlearn_child_indicator_tests}
    \centering
    \scriptsize
    \setlength{\tabcolsep}{2pt}
    \resizebox{\textwidth}{!}{%
    \begin{tabular}{ll l|c|r|rl|rl}
    \toprule
    \textbf{Group} & \textbf{Metric} & \textbf{Methods} & \textbf{Means$\pm$Std} & \textbf{t} & \multicolumn{2}{c|}{\textbf{p-value}} & \multicolumn{2}{c}{\textbf{$p_{\mathrm{BH}}$}} \\
    \midrule
    child & SHD & ABAPC-LLM vs MPC & \num{4.417 +- 2.009} vs \num{9.160 +- 4.130} & $-7.30$ & $2.82\text{e-}13$ & \!\!\!*** & $8.45\text{e-}13$ & \!\!\!*** \\
    child & SHD & MPC-LLM vs MPC & \num{5.060 +- 1.800} vs \num{9.160 +- 4.130} & $-6.44$ & $1.23\text{e-}10$ & \!\!\!*** & $1.85\text{e-}10$ & \!\!\!*** \\
    child & SHD & ABAPC vs MPC & \num{5.372 +- 3.250} vs \num{9.160 +- 4.130} & $-5.10$ & $3.46\text{e-}7$ & \!\!\!*** & $3.46\text{e-}7$ & \!\!\!*** \\
    child & Precision & BOSS vs ABAPC-LLM & \num{0.960 +- 0.080} vs \num{0.894 +- 0.050} & $4.94$ & $7.67\text{e-}7$ & \!\!\!*** & $7.67\text{e-}7$ & \!\!\!*** \\
    child & Recall & ABAPC-LLM vs MPC & \num{0.824 +- 0.080} vs \num{0.630 +- 0.170} & $7.31$ & $2.59\text{e-}13$ & \!\!\!*** & $7.76\text{e-}13$ & \!\!\!*** \\
    child & Recall & MPC-LLM vs MPC & \num{0.800 +- 0.070} vs \num{0.630 +- 0.170} & $6.54$ & $6.22\text{e-}11$ & \!\!\!*** & $9.32\text{e-}11$ & \!\!\!*** \\
    child & Recall & ABAPC vs MPC & \num{0.786 +- 0.129} vs \num{0.630 +- 0.170} & $5.17$ & $2.38\text{e-}7$ & \!\!\!*** & $2.38\text{e-}7$ & \!\!\!*** \\
    child & F1 & ABAPC-LLM vs MPC-LLM & \num{0.855 +- 0.052} vs \num{0.830 +- 0.050} & $2.48$ & $0.013$ & \!\!\!* & $0.013$ & \!\!\!* \\
    child & SID & ABAPC-LLM vs MPC-LLM & \num{72.653 +- 24.836} vs \num{87.820 +- 26.340} & $-2.96$ & $0.003$ & \!\!\!** & $0.006$ & \!\!\!** \\
    child & SID & ABAPC vs MPC-LLM & \num{84.833 +- 45.226} vs \num{87.820 +- 26.340} & $-0.40$ & $0.687$ & \!\!\!ns & $0.687$ & \!\!\!ns \\
    \bottomrule
    \end{tabular}
    }
\end{table*}
\begin{table*}[t]
    \caption{Non-significant best-vs-comparator tests that justify additional bolded methods in Table~\ref{tab:bnlearn_dag_all_metrics}. Significance levels for both reported $p$-values are: \textnormal{***} $p<0.001$, \textnormal{**} $p<0.01$, \textnormal{*} $p<0.05$, \textnormal{.} $p<0.1$.}
    \label{tab:bnlearn_tied_bolding_tests}
    \centering
    \scriptsize
    \setlength{\tabcolsep}{2pt}
    \resizebox{\textwidth}{!}{%
    \begin{tabular}{ll l|c|r|rl|rl}
    \toprule
    \textbf{Group} & \textbf{Metric} & \textbf{Methods} & \textbf{Means$\pm$Std} & \textbf{t} & \multicolumn{2}{c|}{\textbf{p-value}} & \multicolumn{2}{c}{\textbf{$p_{\mathrm{BH}}$}} \\
    \midrule
    asia & Precision & ABAPC-LLM vs MPC & \num{0.992 +- 0.040} vs \num{0.980 +- 0.080} & $0.95$ & $0.342$ & \!\!\!ns & $0.391$ & \!\!\!ns \\
    asia & Precision & ABAPC-LLM vs MPC-LLM & \num{0.992 +- 0.040} vs \num{0.990 +- 0.030} & $0.28$ & $0.776$ & \!\!\!ns & $0.776$ & \!\!\!ns \\
    \midrule
    cancer & SHD & MPC-LLM vs ABAPC-LLM & \num{0.680 +- 0.590} vs \num{0.680 +- 0.587} & $0.00$ & $1.000$ & \!\!\!ns & $1.000$ & \!\!\!ns \\
    cancer & Precision & ABAPC-LLM vs MPC-LLM & \num{0.996 +- 0.028} vs \num{0.990 +- 0.050} & $0.74$ & $0.460$ & \!\!\!ns & $0.460$ & \!\!\!ns \\
    cancer & Recall & MPC-LLM vs ABAPC-LLM & \num{0.840 +- 0.130} vs \num{0.835 +- 0.148} & $0.18$ & $0.858$ & \!\!\!ns & $0.858$ & \!\!\!ns \\
    cancer & F1 & MPC vs MPC-LLM & \num{0.940 +- 0.180} vs \num{0.900 +- 0.080} & $1.44$ & $0.151$ & \!\!\!ns & $0.167$ & \!\!\!ns \\
    cancer & F1 & MPC vs ABAPC-LLM & \num{0.940 +- 0.180} vs \num{0.901 +- 0.091} & $1.38$ & $0.167$ & \!\!\!ns & $0.167$ & \!\!\!ns \\
    cancer & SID & MPC-LLM vs ABAPC-LLM & \num{0.780 +- 1.000} vs \num{0.960 +- 1.324} & $-0.77$ & $0.443$ & \!\!\!ns & $0.443$ & \!\!\!ns \\
    \midrule
    earthquake & SHD & MPC vs GRaSP & \num{0.040 +- 0.200} vs \num{0.160 +- 0.790} & $-1.04$ & $0.298$ & \!\!\!ns & $0.476$ & \!\!\!ns \\
    earthquake & SHD & MPC vs BOSS & \num{0.040 +- 0.200} vs \num{0.400 +- 1.210} & $-2.08$ & $0.038$ & \!\!\!* & $0.076$ & \!\!\!. \\
    earthquake & SHD & MPC vs MPC-LLM & \num{0.040 +- 0.200} vs \num{0.040 +- 0.200} & $0.00$ & $1.000$ & \!\!\!ns & $1.000$ & \!\!\!ns \\
    earthquake & SHD & MPC vs ABAPC & \num{0.040 +- 0.200} vs \num{0.100 +- 0.463} & $-0.84$ & $0.400$ & \!\!\!ns & $0.534$ & \!\!\!ns \\
    earthquake & SHD & MPC vs ABAPC-LLM & \num{0.040 +- 0.200} vs \num{0.080 +- 0.340} & $-0.72$ & $0.474$ & \!\!\!ns & $0.541$ & \!\!\!ns \\
    earthquake & Precision & GRaSP vs BOSS & \num{1.000 +- 0.000} vs \num{1.000 +- 0.000} & $0.00$ & $1.000$ & \!\!\!ns & $1.000$ & \!\!\!ns \\
    earthquake & Precision & GRaSP vs MPC & \num{1.000 +- 0.000} vs \num{0.990 +- 0.040} & $1.77$ & $0.077$ & \!\!\!. & $0.105$ & \!\!\!ns \\
    earthquake & Precision & GRaSP vs MPC-LLM & \num{1.000 +- 0.000} vs \num{0.990 +- 0.040} & $1.77$ & $0.077$ & \!\!\!. & $0.105$ & \!\!\!ns \\
    earthquake & Precision & GRaSP vs ABAPC & \num{1.000 +- 0.000} vs \num{0.982 +- 0.080} & $1.59$ & $0.112$ & \!\!\!ns & $0.128$ & \!\!\!ns \\
    earthquake & Precision & GRaSP vs ABAPC-LLM & \num{1.000 +- 0.000} vs \num{0.987 +- 0.052} & $1.76$ & $0.079$ & \!\!\!. & $0.105$ & \!\!\!ns \\
    earthquake & Recall & MPC vs GRaSP & \num{1.000 +- 0.000} vs \num{0.960 +- 0.200} & $1.41$ & $0.157$ & \!\!\!ns & $0.252$ & \!\!\!ns \\
    earthquake & Recall & MPC vs MPC-LLM & \num{1.000 +- 0.000} vs \num{1.000 +- 0.000} & $0.00$ & $1.000$ & \!\!\!ns & $1.000$ & \!\!\!ns \\
    earthquake & Recall & MPC vs ABAPC & \num{1.000 +- 0.000} vs \num{0.990 +- 0.071} & $1.00$ & $0.317$ & \!\!\!ns & $0.363$ & \!\!\!ns \\
    earthquake & Recall & MPC vs ABAPC-LLM & \num{1.000 +- 0.000} vs \num{0.995 +- 0.035} & $1.00$ & $0.317$ & \!\!\!ns & $0.363$ & \!\!\!ns \\
    earthquake & F1 & GRaSP vs BOSS & \num{1.000 +- 0.000} vs \num{1.000 +- 0.000} & $0.00$ & $1.000$ & \!\!\!ns & $1.000$ & \!\!\!ns \\
    earthquake & F1 & GRaSP vs MPC & \num{1.000 +- 0.000} vs \num{1.000 +- 0.020} & $0.00$ & $1.000$ & \!\!\!ns & $1.000$ & \!\!\!ns \\
    earthquake & F1 & GRaSP vs MPC-LLM & \num{1.000 +- 0.000} vs \num{1.000 +- 0.020} & $0.00$ & $1.000$ & \!\!\!ns & $1.000$ & \!\!\!ns \\
    earthquake & F1 & GRaSP vs ABAPC & \num{1.000 +- 0.000} vs \num{0.986 +- 0.073} & $1.39$ & $0.164$ & \!\!\!ns & $0.263$ & \!\!\!ns \\
    earthquake & F1 & GRaSP vs ABAPC-LLM & \num{1.000 +- 0.000} vs \num{0.991 +- 0.041} & $1.63$ & $0.104$ & \!\!\!ns & $0.208$ & \!\!\!ns \\
    earthquake & SID & MPC vs GRaSP & \num{0.000 +- 0.000} vs \num{0.400 +- 1.980} & $-1.43$ & $0.153$ & \!\!\!ns & $0.245$ & \!\!\!ns \\
    earthquake & SID & MPC vs MPC-LLM & \num{0.000 +- 0.000} vs \num{0.000 +- 0.000} & $0.00$ & $1.000$ & \!\!\!ns & $1.000$ & \!\!\!ns \\
    earthquake & SID & MPC vs ABAPC & \num{0.000 +- 0.000} vs \num{0.160 +- 1.131} & $-1.00$ & $0.317$ & \!\!\!ns & $0.363$ & \!\!\!ns \\
    earthquake & SID & MPC vs ABAPC-LLM & \num{0.000 +- 0.000} vs \num{0.080 +- 0.566} & $-1.00$ & $0.317$ & \!\!\!ns & $0.363$ & \!\!\!ns \\
    \midrule
    survey & Precision & GRaSP vs BOSS & \num{1.000 +- 0.000} vs \num{1.000 +- 0.000} & $0.00$ & $1.000$ & \!\!\!ns & $1.000$ & \!\!\!ns \\
    \midrule
    child & SHD & ABAPC-LLM vs MPC-LLM & \num{4.417 +- 2.009} vs \num{5.060 +- 1.800} & $-1.69$ & $0.092$ & \!\!\!. & $0.092$ & \!\!\!. \\
    child & SHD & ABAPC-LLM vs ABAPC & \num{4.417 +- 2.009} vs \num{5.372 +- 3.250} & $-1.77$ & $0.077$ & \!\!\!. & $0.088$ & \!\!\!. \\
    child & Recall & ABAPC-LLM vs MPC-LLM & \num{0.824 +- 0.080} vs \num{0.800 +- 0.070} & $1.61$ & $0.107$ & \!\!\!ns & $0.107$ & \!\!\!ns \\
    child & Recall & ABAPC-LLM vs ABAPC & \num{0.824 +- 0.080} vs \num{0.786 +- 0.129} & $1.78$ & $0.075$ & \!\!\!. & $0.085$ & \!\!\!. \\
    child & SID & ABAPC-LLM vs ABAPC & \num{72.653 +- 24.836} vs \num{84.833 +- 45.226} & $-1.67$ & $0.095$ & \!\!\!. & $0.095$ & \!\!\!. \\
    \bottomrule
    \end{tabular}
    }
\end{table*}
Finally, Table~\ref{tab:matched_llm_indicator_tests} gives the indicator tests for the LLM-source ablation.
\begin{table*}[t]
    \caption{Welch tests for Table~\ref{tab:gpt_model_ablation_dag}. Significance levels for both reported $p$-values are: \textnormal{***} $p<0.001$, \textnormal{**} $p<0.01$, \textnormal{*} $p<0.05$, \textnormal{.} $p<0.1$.}
    \label{tab:matched_llm_indicator_tests}
    \centering
    \scriptsize
    \setlength{\tabcolsep}{2pt}
    \resizebox{\textwidth}{!}{%
    \begin{tabular}{ll l|c|r|rl|rl}
    \toprule
    \textbf{Group} & \textbf{Metric} & \textbf{Methods} & \textbf{Means$\pm$Std} & \textbf{t} & \multicolumn{2}{c|}{\textbf{p-value}} & \multicolumn{2}{c}{\textbf{$p_{\mathrm{BH}}$}} \\
    \midrule
    $|\nodeSet|=5$ & SHD & ABAPC-LLM (Gemini) vs MPC-LLM (Gemini) & \num{1.186 +- 0.375} vs \num{1.328 +- 0.331} & $-2.57$ & $0.010$ & \!\!\!* & $0.010$ & \!\!\!* \\
    $|\nodeSet|=5$ & Precision & MPC-LLM (Gemini) vs ABAPC-LLM (Gemini) & \num{0.952 +- 0.027} vs \num{0.902 +- 0.043} & $7.39$ & $1.50\text{e-}13$ & \!\!\!*** & $1.50\text{e-}13$ & \!\!\!*** \\
    $|\nodeSet|=5$ & Recall & ABAPC-LLM (Gemini) vs MPC-LLM (Gemini) & \num{0.778 +- 0.063} vs \num{0.753 +- 0.053} & $2.20$ & $0.028$ & \!\!\!* & $0.028$ & \!\!\!* \\
    $|\nodeSet|=5$ & F1 & ABAPC-LLM (Gemini) vs ABAPC-LLM (GPT) & \num{0.828 +- 0.048} vs \num{0.755 +- 0.084} & $7.82$ & $5.42\text{e-}15$ & \!\!\!*** & $1.08\text{e-}14$ & \!\!\!*** \\
    $|\nodeSet|=5$ & F1 & MPC-LLM (Gemini) vs ABAPC-LLM (GPT) & \num{0.822 +- 0.041} vs \num{0.755 +- 0.084} & $6.56$ & $5.37\text{e-}11$ & \!\!\!*** & $5.37\text{e-}11$ & \!\!\!*** \\
    \midrule
    $|\nodeSet|=10$ & SHD & ABAPC-LLM (GPT) vs MPC-LLM (Gemini) & \num{5.755 +- 1.417} vs \num{6.197 +- 1.095} & $-3.63$ & $2.81\text{e-}4$ & \!\!\!*** & $5.62\text{e-}4$ & \!\!\!*** \\
    $|\nodeSet|=10$ & SHD & ABAPC-LLM (Gemini) vs MPC-LLM (Gemini) & \num{5.984 +- 1.257} vs \num{6.197 +- 1.095} & $-1.86$ & $0.063$ & \!\!\!. & $0.063$ & \!\!\!. \\
    $|\nodeSet|=10$ & Precision & BOSS vs MPC-LLM (GPT) & \num{0.884 +- 0.201} vs \num{0.860 +- 0.095} & $3.80$ & $1.43\text{e-}4$ & \!\!\!*** & $1.43\text{e-}4$ & \!\!\!*** \\
    $|\nodeSet|=10$ & Recall & ABAPC-LLM (GPT) vs ABAPC-LLM (Gemini) & \num{0.569 +- 0.101} vs \num{0.542 +- 0.090} & $4.71$ & $2.50\text{e-}6$ & \!\!\!*** & $2.50\text{e-}6$ & \!\!\!*** \\
    $|\nodeSet|=10$ & F1 & ABAPC-LLM (GPT) vs MPC-LLM (Gemini) & \num{0.655 +- 0.102} vs \num{0.634 +- 0.084} & $3.59$ & $3.27\text{e-}4$ & \!\!\!*** & $3.27\text{e-}4$ & \!\!\!*** \\
    \midrule
    $|\nodeSet|=15$ & SHD & ABAPC-LLM (Gemini) vs ABAPC & \num{7.508 +- 1.292} vs \num{7.863 +- 1.602} & $-2.67$ & $0.007$ & \!\!\!** & $0.022$ & \!\!\!* \\
    $|\nodeSet|=15$ & SHD & ABAPC-LLM (GPT) vs ABAPC & \num{7.575 +- 1.513} vs \num{7.863 +- 1.602} & $-2.14$ & $0.033$ & \!\!\!* & $0.049$ & \!\!\!* \\
    $|\nodeSet|=15$ & SHD & MPC-LLM (Gemini) vs ABAPC & \num{7.616 +- 1.052} vs \num{7.863 +- 1.602} & $-1.85$ & $0.064$ & \!\!\!. & $0.064$ & \!\!\!. \\
    $|\nodeSet|=15$ & Precision & MPC-LLM (GPT) vs BOSS & \num{0.941 +- 0.064} vs \num{0.892 +- 0.176} & $14.36$ & $9.39\text{e-}47$ & \!\!\!*** & $2.82\text{e-}46$ & \!\!\!*** \\
    $|\nodeSet|=15$ & Precision & MPC-LLM (Gemini) vs BOSS & \num{0.934 +- 0.049} vs \num{0.892 +- 0.176} & $12.27$ & $1.33\text{e-}34$ & \!\!\!*** & $1.99\text{e-}34$ & \!\!\!*** \\
    $|\nodeSet|=15$ & Precision & MPC vs BOSS & \num{0.934 +- 0.068} vs \num{0.892 +- 0.176} & $12.14$ & $6.40\text{e-}34$ & \!\!\!*** & $6.40\text{e-}34$ & \!\!\!*** \\
    $|\nodeSet|=15$ & Recall & ABAPC-LLM (Gemini) vs MPC-LLM (Gemini) & \num{0.573 +- 0.073} vs \num{0.557 +- 0.060} & $2.96$ & $0.003$ & \!\!\!** & $0.006$ & \!\!\!** \\
    $|\nodeSet|=15$ & Recall & ABAPC-LLM (GPT) vs MPC-LLM (Gemini) & \num{0.570 +- 0.086} vs \num{0.557 +- 0.060} & $2.35$ & $0.019$ & \!\!\!* & $0.019$ & \!\!\!* \\
    $|\nodeSet|=15$ & F1 & MPC-LLM (Gemini) vs MPC-LLM (GPT) & \num{0.691 +- 0.055} vs \num{0.671 +- 0.068} & $3.71$ & $2.09\text{e-}4$ & \!\!\!*** & $2.09\text{e-}4$ & \!\!\!*** \\
    \bottomrule
    \end{tabular}
    }
\end{table*}
\end{document}